\documentclass{article} 
\usepackage{iclr2026_conference,times}

\usepackage{amsmath,amsfonts,bm}

\def\eqref#1{equation~\ref{#1}}

\def\1{\bm{1}}

\DeclareMathAlphabet{\mathsfit}{\encodingdefault}{\sfdefault}{m}{sl}
\SetMathAlphabet{\mathsfit}{bold}{\encodingdefault}{\sfdefault}{bx}{n}

\usepackage{rotating}
\usepackage{tablefootnote}
\usepackage{adjustbox}
\usepackage{hyperref}
\usepackage{url}
\usepackage{algorithm}
\usepackage{algorithmic}
\usepackage{booktabs}
\usepackage{subcaption}
\usepackage{array}
\usepackage{multirow}
\newcolumntype{C}[1]{>{\centering\arraybackslash}m{#1}}
\newcolumntype{L}[1]{>{\raggedright\arraybackslash}m{#1}}

\usepackage{xcolor,tikz}
\usetikzlibrary{arrows.meta,positioning,calc}
\definecolor{legAnti}{HTML}{9BB7D7}  
\definecolor{legBalanced}{HTML}{F4A35D}      
\definecolor{legInduction}{HTML}{F28E8E}  
\definecolor{legBaseline}{HTML}{9FD6C5} 
\usepackage[table]{xcolor}

\title{Induction Signatures Are Not Enough: A Matched-Compute Study of Load-Bearing Structure in In-Context Learning}

\author{%
  Mohammed Sabry \\
  ADAPT Centre, Dublin City University, Ireland\\
  \texttt{mohammed.sabry@adaptcentre.ie} \\
  \And
  Anya Belz \\
   ADAPT Centre, Dublin City University, Ireland\\
  \texttt{anya.belz@dcu.ie} \\
}

\iclrfinalcopy 
\begin{document}

\maketitle

\begin{abstract}
    Mechanism-targeted synthetic data is increasingly proposed as a way to steer pretraining toward desirable capabilities, but it remains unclear how such interventions should be evaluated. We study this question for in-context learning (ICL) under matched compute (iso-FLOPs) using \textbf{Bi-Induct}, a lightweight data rewrite that interleaves short directional copy snippets into a natural pretraining stream: forward-copy (induction), backward-copy (anti-induction, as a directional control), or a balanced mix. Across 0.13B–1B decoder-only models, we evaluate (i) few-shot performance on standard LM benchmarks and function-style ICL probes, (ii) head-level copy telemetry, and (iii) held-out perplexity as a guardrail. Bi-Induct reliably increases induction-head activity, but this does not translate into consistent improvements in few-shot generalization: on standard LM benchmarks, Bi-Induct is largely performance-neutral relative to natural-only training, while on function-style probes the 1B natural-only model performs best. Despite explicit backward-copy cues, anti-induction scores remain near zero across scales, revealing a strong forward/backward asymmetry. Targeted ablations show a sharper distinction: removing the top 2\% induction heads per layer harms ICL more than matched random ablations, with the largest relative drop occurring in the natural-only models. This indicates that natural-only training produces more centralized, load-bearing induction circuitry, whereas Bi-Induct tends to create more distributed and redundant induction activity. Our main conclusion is that \textbf{eliciting a mechanism is not the same as making it load-bearing}. For data-centric foundation model design, this suggests that synthetic data interventions should be evaluated not only by signature amplification, but by whether they create causally necessary computation while preserving natural-data modeling quality.
\end{abstract}

\section{Introduction}

Transformer language models learn a simple copy pattern early in training: when a token \(A\) reappears in context, the model increases the probability of the token that followed the previous \(A\). Prior work identified a two-head motif  implementing this behavior and linked it to in-context learning on pattern-matching tasks \citep{olsson2022context}. Despite its simplicity, this motif typically emerges only after many billions of tokens, well after the first training-loss plateau. Theoretical and empirical studies frame the delay as a phase transition \citep{chen2024unveilinginductionheadsprovable, edelman2024the}. In principle, if one could make induction-like computation become useful earlier in training, this could reduce the compute needed to reach ICL-relevant behaviors and would expose the responsible circuitry sooner for analysis.

A practical way to approach this is to intervene on the \emph{data} rather than the \emph{architecture} or \emph{objective}. 
In contrast to synthetic-task-only plateau-shortening studies \citep{kim2025taskdiversityshortensicl} and objective-level interventions such as multi-token prediction \citep{gloeckle2024betterfasterlarge}, we adopt a data-rewrite perspective that is easy to deploy at scale: inject a small fraction of targeted inputs into the pretraining stream that selectively excite the putative induction mechanism while keeping the architecture and objective fixed. Concretely, we replace a small slice of natural tokens with synthetic copy snippets that cleanly exercise the copier-selector behavior of induction (forward copy) and anti-induction (backward copy). Copy-style cues are the canonical probe for the induction circuit and are widely used to measure it \citep{olsson2022context, nanda2022transformerlens}. While other synthetic families have been studied (for example n-gram statistics \citep{edelman2024the}, p-hop tasks \citep{sanford2024transformersparallelcomputationlogarithmic}, and intrinsic tasks \citep{gu2023pretraininglearncontext}), copy snippets align most directly with the hypothesized mechanism and with distributional properties such as burstiness that correlate with the rise of in-context learning \citep{chan2022datadistributionalpropertiesdrive}. 

These considerations lead to a single testable question: \textbf{under iso-FLOPs, is it more effective for ICL to pretrain purely on natural text, or to allocate a small early-training budget to synthetic directional copy snippets that directly exercise the induction circuit?} Here, “earlier” refers to earlier-peaking induction signatures across layers at a fixed checkpoint under matched compute, not fewer optimization steps to reach a fixed capability.

To answer this, we introduce \textbf{Bi-Induct}, a lightweight curriculum that interleaves short duplicate-span snippets with natural text during early training. We evaluate three variants under matched compute: forward induction, backward anti-induction, and a balanced mix where the direction is chosen independently at each injection. We assess each pretraining strategy on three axes: (i) downstream ICL performance on standard few-shot benchmarks and function-style probes; (ii) mechanistic telemetry targeting induction and anti-induction heads; and (iii) held-out perplexity as a quality guardrail. We study these effects across 0.13B, 0.5B, and 1B decoder-only models, using the 0.13B setting as a design lab for span length and mix-ratio selection before scaling the chosen operating point.

Our objective is \textbf{not} to present Bi-Induct as a generally superior pretraining recipe. Rather, we use it as a controlled case study for a broader data-centric question: when a synthetic data intervention successfully amplifies a target mechanism, does that mechanism become \textbf{causally load-bearing} for downstream behavior, or does it remain a redundant by-product of training? This distinction is especially important for data-centric foundation model design, where synthetic rewrites are attractive because they are easy to deploy, but their practical value depends on whether they improve useful computation rather than merely producing stronger internal signatures.

\textbf{Contributions:}
\vspace{-0.1cm}
\begin{itemize}\setlength{\itemsep}{-1pt}
\vspace{-0.1cm}

\item \textbf{A mechanism-aware evaluation criterion for synthetic data interventions.} We distinguish between \textbf{circuit emergence} and \textbf{circuit load-bearing}: a targeted mechanism may become visible in telemetry without becoming necessary for task performance.

\item \textbf{A matched-compute case study of this distinction.} Under iso-FLOPs across 0.13B, 0.5B, and 1B models, we show that Bi-Induct reliably increases induction-head activity but does not reliably improve few-shot ICL, and at 1B the natural-only baseline performs best on function-style probes.

\item \textbf{Causal evidence via targeted ablation.} Using head-level copy telemetry together with top-2\% induction-head ablations, we show that the largest ICL drops occur in the natural-only condition, indicating more centralized, load-bearing induction circuitry there than under Bi-Induct.

\item \textbf{Directional controls for interpreting copy signals.} By comparing forward induction, backward anti-induction, and a balanced mixture, we find a strong forward/backward asymmetry: even explicit anti-induction training does not materially increase anti-induction scores.

\item \textbf{Practical lessons for the data-centric design of foundation models (FMs).} Our results suggest that synthetic data interventions should be judged not only by whether they amplify a target signature, but by whether they create behaviorally useful and causally necessary computation without unduly sacrificing natural-data modeling quality.

\end{itemize}

For a concise glossary of terms and symbols, see Appendix~\ref{app:glossary}.

\section{Related Work}
\label{sec:related_work}

\textbf{Induction heads and the mechanics of ICL:}
The induction-head motif-a two-head circuit that matches a repeated cue and copies its following token-was introduced by \citet{olsson2022context}, who provided multiple converging tests linking it to the rise of in-context learning (ICL). Follow-up theory and controlled synthetic-task studies characterize the behavior as a phase transition: on Markov-chain data, models move from uniform predictions to unigram heuristics and then abruptly to bigram induction \citep{edelman2024the}. Provable analyses show that even shallow transformers implement generalized induction via a copier-selector-classifier pipeline, tightening the link between optimization dynamics and the circuit \citep{chen2024unveilinginductionheadsprovable}. At scale, targeted head ablations support causality: removing a small fraction of high-score induction heads reduces few-shot gains by up to $\sim$32\% on abstract pattern tasks and weakens benefits on NLP tasks \citep{crosbie-shutova-2025-induction}. Open suites and tools (e.g., Pythia and TransformerLens) have made these effects reproducible across model sizes \citep{biderman2023pythiasuiteanalyzinglarge, nanda2022transformerlens}.

\textbf{Anti-induction and copy-suppression circuits:}
Beyond forward copying, models host heads that suppress copying or implement the backward, “anti-induction” direction. Work on negative heads explains copy suppression as a coherent mechanism that interacts with induction patterns \citep{mcdougall2023copysuppressioncomprehensivelyunderstanding}. Large-scale empirical reports find a pretrained asymmetry-transformers are stronger at forward induction than backward copy-an imbalance that targeted fine-tuning can reduce while isolating distinct head families \citep{veitsman2025borntransformertransformer}. In parallel, \citet{wang2025inductionheadtoxicitymechanistically} mechanistically link the `repetition curse' to over-dominant induction heads and propose head-level regularization to restore output diversity.

\textbf{Curricula that accelerate circuit emergence:}
A growing line of work seeks to shorten the ICL plateau. Training on diverse ICL tasks in parallel reduces plateau length and eases optimization relative to single-task settings \citep{kim2025taskdiversityshortensicl}. Orthogonal to data choice, multi-token prediction modifies the objective to encourage longer-range patterns and shows favorable development of induction-like behaviors together with efficiency gains \citep{gloeckle2024betterfasterlarge}. However, these studies concentrate on \emph{forward} induction and are typically evaluated in \emph{synthetic-task-only} training regimes rather than natural-language pretraining, or they rely on objective/architectural changes rather than data-only interventions in end-to-end runs. To our knowledge, they also do not systematically induce or measure \emph{anti}-induction emergence.

\textbf{Embedding induction heads in downstream systems:}
Architectural and application work has begun to `install' $n$-gram induction heads to stabilize in-context RL and reduce data needs, demonstrating practical leverage of the circuit in agents \citep{zisman2025ngraminductionheadsincontext}.

\textbf{Data rewriting:}
Beyond filtering, recent work rewrites pretraining text to shift style and structure before learning. Rephrasing the Web (WRAP) uses instruction-tuned models to paraphrase web pages into target styles, yielding $\sim$3$\times$ faster pretraining on noisy corpora, lower perplexity, and modest zero-shot gains under matched compute budgets \citep{maini2024rephrasingwebrecipecompute}. \citet{nguyen2025recyclingwebmethodenhance} pursue a related transform-and-retain strategy focused on discarded low-quality documents. \citet{fujii2025rewritingpretrainingdataboosts} expand the rewrite paradigm beyond style, reporting boosts on math and code. In parallel, large open datasets such as RefinedWeb show that aggressive deduplication and domain organization improve pretraining without synthetic rewrites, positioning data rewriting as complementary to quality and mixture knobs \citep{penedo2023refinedwebdatasetfalconllm}.

\textbf{Circuit discovery and emergence shaping:}
Mechanistic interpretability maps internal circuits via activation interventions \citep{zhang2024bestpracticesactivationpatching}, probing \citep{gurnee2023findingneuronshaystackcase}, and increasingly, sparse-autoencoder features \citep{cunningham2023sparseautoencodershighlyinterpretable}. Our focus is earlier in the lifecycle: shaping the data distribution so that desirable circuits appear sooner and more predictably, then verifying the link with head-level telemetry.

\textit{\textbf{Summary:}}
Existing work 
investigates \emph{circuit discovery/emergence shaping} and, separately, \emph{data rewriting}, but rarely bridges the two-i.e., using mechanistic insight to design pretraining data and evaluating the outcome under matched-compute conditions. We make that link explicit: we target a canonical ICL circuit (forward and backward induction) with the minimal inputs that excite it (directional copy snippets), and compare \emph{pure natural pretraining} to \emph{Bi-Induct}, a small, linearly annealed replacement policy, under iso-FLOPs conditions on the same corpus. We 
measure effects behaviorally (few-shot ICL benchmarks) and mechanistically (top 2\% head concentration by layer), 
alongside a perplexity guardrail. Unlike prior curricula that emphasize forward copy alone, we study a symmetric forward/backward curriculum side by side to ask whether targeted copy signals are more valuable than additional natural tokens at the same compute.

\section{Bi-Induct}
\label{sec:bi_induct}
We investigate the effect of \textit{data rewrites} on circuit emergence: we \emph{interleave} synthetic copy snippets into the pretraining stream to explicitly exercise a canonical copy pattern associated with induction. Bi-Induct has two primary variants that differ only in the direction of the copy cue (forward vs.\ backward). A third variant, \textit{balanced}, {flips a coin between forward and backward injections} to provide a mixed signal.

\subsection{Synthetic snippet construction}
Let $\mathcal{V}$ be the tokenizer vocabulary and let $\mathrm{BPE}(\cdot)$ be the tokenizer. For a span length $L$, we first sample a token span
\[
S=(s_1,\ldots,s_L),\qquad
s_i \sim \mathrm{Uniform}\!\left(\{ \lfloor 0.05|\mathcal{V}|\rfloor, \ldots, \lfloor 0.95|\mathcal{V}|\rfloor \}\right),
\]
which avoids special/rare IDs. We use a single space as a neutral separator, $\mathrm{SEP}=\mathrm{BPE}(\texttt{" "})$.

\begin{figure}[t]
\centering
\begin{tikzpicture}[x=0.7cm,y=0.9cm,font=\small]
\tikzset{
  tok/.style={draw,rounded corners=1pt,minimum width=0.6cm,minimum height=0.5cm,inner sep=1pt},
  tokA/.style={tok,fill=blue!20,   draw=blue!60!black},
  tokB/.style={tok,fill=orange!25, draw=orange!70!black},
  tokC/.style={tok,fill=violet!20, draw=violet!60!black},
  tokD/.style={tok,fill=green!20,  draw=green!60!black},
  tokE/.style={tok,fill=red!15,    draw=red!60!black},
  sepbox/.style={tok,fill=gray!10, draw=gray!50}
}

\node[align=left] at (-1.6,2.0) {\textbf{Induction}};
\foreach \i/\t in {1/A,2/B,3/C,4/D,5/E}{
  \node[tok\t] at (\i,2) {\t};
}
\node[sepbox] at (6,2) {\textvisiblespace};
\foreach \i/\t in {7/A,8/B,9/C,10/D,11/E}{
  \node[tok\t] at (\i,2) {\t};
}

\node[align=left] at (-1.6,1.0) {\textbf{Anti-induction}};
\foreach \i/\t in {1/A,2/B,3/C,4/D,5/E}{
  \node[tok\t] at (\i,1) {\t};
}
\node[sepbox] at (6,1) {\textvisiblespace};
\foreach \i/\t in {7/E,8/D,9/C,10/B,11/A}{
  \node[tok\t] at (\i,1) {\t};
}
\end{tikzpicture}
\caption{Examples of copy-style snippets injected into the pretraining stream. Each snippet is a span of \(L\) random non-special tokens, followed by a separator, then either the same span (induction) or the reversed span (anti-induction). Colors align repeated tokens across the two halves. The illustration uses \(L{=}5\) for clarity.}
\label{fig:biinduct-snippets}
\end{figure}
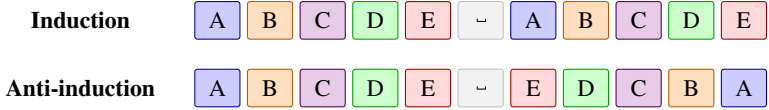

\paragraph{Forward/induction (Figure~\ref{fig:biinduct-snippets}, top):}
\[
\mathrm{Induction}(S)=[S \;\Vert\; \mathrm{SEP} \;\Vert\; S].
\]

\paragraph{Backward/anti-induction (Figure~\ref{fig:biinduct-snippets}, bottom):}
\[
\mathrm{Anti}(S)=[S \;\Vert\; \mathrm{SEP} \;\Vert\; \mathrm{reverse}(S)].
\]

\paragraph{Balanced (a mix of forward and backward injections):}
On each injection, flip a fair coin to choose between forward or backward. 

\noindent Each snippet has length $\ell_{\text{snip}} = 2L + |\mathrm{SEP}|$ (e.g., $2L{+}1$ when $\mathrm{SEP}$ is a single space).

\subsection{Curriculum schedule and injection rule}
We interleave snippets 
on the fly during streaming pretraining. Let $m_0$ be the initial mix ratio\footnote{
Or mix ratio, for short.} and $T_a$ an anneal budget (in natural tokens). After $t$ natural tokens have been seen, the instantaneous injection probability is
\[
m(t)=\max\!\bigl\{\,m_0\!\cdot\!(1{-}t/T_a),\,0\,\bigr\}.
\]

On each natural example, draw $u \sim \mathrm{Uniform}(0,1)$. If $u<m(t)$, we first yield one synthetic snippet (
depending on the current $\texttt{synthetic\_task}\in\{\texttt{induction}, \texttt{anti}, \texttt{balanced}\}$), then yield the natural tokenized sequence. \emph{Else} ($u\ge m(t)$), we emit only the natural tokenized sequence. This implements a light interleave rather than full replacement and keeps the natural distribution dominant.

\paragraph{Expected injection budget ($\bar m$):}
With a linear anneal $m(t)=m_0(1-t/T_a)$ for $t\!\in[0,T_a]$ and $m(t){=}0$ afterwards, the \emph{average} injection rate over the anneal is $m_0/2$. Let $T_{\text{base}}$ be the natural-token budget of the run. The fraction of injected snippets over the \emph{whole} run is therefore:
\[
\bar m \;=\; \frac{1}{T_{\text{base}}}\!\int_0^{T_{\text{base}}} m(t)\,dt
\;=\;
\begin{cases}
m_0/2, & T_a \ge T_{\text{base}},\\[4pt]
m_0\,\dfrac{T_a}{2T_{\text{base}}}, & T_a < T_{\text{base}}.
\end{cases}
\]

\textbf{Why this schedule?}
(i) \emph{Front-loading the signal:} Induction circuits typically emerge after the first loss plateau; concentrating copy cues early helps trigger the phase transition without interfering with late-stage calibration; 
(ii) \emph{Stability:} A linear anneal avoids abrupt distribution shifts and exposes a single knob ($m_0$) for clean sweeps; and
(iii) \emph{Compute considerations:} Under standard packing, snippets can share sequences with natural text so the incremental token cost scales with $\ell_{\text{snip}}$ rather than a full segment. We \emph{enforce iso-FLOPs across conditions} (fixed sequence length and optimizer steps), so any potential savings from aggressive packing are intentionally not exploited.\footnote{We fix compute to avoid conflating efficiency optimizations with capability changes; our focus is whether targeted directional copy improves ICL \emph{at the same compute}.}

\begin{table}[h!]
\centering
\small
\setlength{\tabcolsep}{4pt}
\renewcommand{\arraystretch}{1.0}
\caption{Model presets used in experiments. Attention uses head dimension 64; \#heads $=\texttt{hidden}/64$ and \#KV heads $=\max(1,\lfloor\text{\#heads}/4\rfloor)$.}
\label{tab:model-presets}
\begin{tabular}{lcccccc}
\toprule
\textbf{Model} & \textbf{Layers} & \textbf{Hidden} & \textbf{MLP (intermediate)} & \textbf{Head dim} & \textbf{\#Attn heads} & \textbf{\#KV heads} \\
\midrule
0.13B & 12 & \;\;768 & \;\;3{,}072  & 64 & 12 & 3 \\
0.5B  & 30 & 1{,}024 & \;\;4{,}096  & 64 & 16 & 4 \\
1B    & 30 & 1{,}536 & \;\;6{,}144  & 64 & 24 & 6 \\
\bottomrule
\end{tabular}
\end{table}

\section{Experimental Setup}\label{sec:experiments_setup}
\textbf{Model:}
We use a causal decoder-only Transformer with rotary position embeddings (\textsc{RoPE}, $\theta{=}10{,}000$), pre-norm residual blocks, and a gated MLP with SiLU activation (SwiGLU). Self-attention uses \emph{grouped key-value attention} (GQA): for head dimension 64, we set number of attention heads $=\texttt{hidden}/64$ and number of KV heads $=\max(1,\lfloor\text{\#heads}/4\rfloor)$. We train in \textbf{bfloat16} with context length 1{,}024 and \emph{untied} input/output embeddings.\footnote{Our architecture largely follows the Mistral-7B design (decoder-only, pre-norm, RoPE, SwiGLU, GQA) \citep{jiang2023mistral7b}.} Hidden sizes, heads, KV heads, layer counts, and proportional MLP widths are shown in Table~\ref{tab:model-presets}.

\textbf{Pretraining data:}
We pretrain on the deduplicated \textsc{The Pile} dataset \citep{gao2020pile800gbdatasetdiverse} in streaming/shuffled mode. A stable MD5-based hash assigns a fixed held-out evaluation slice so train/eval partitions remain identical across runs; we set this slice to \textbf{0.2\%} of the corpus which corresponds to roughly \textbf{0.4B} tokens. Tokenization truncates to 1{,}024 tokens per sequence. Synthetic snippets are interleaved by the Bi-Induct iterator as described in Section~\ref{sec:bi_induct}. 

\textbf{Training recipe:}
We train all model presets with peak learning rate 1e-3 with linear warmup of 3\% of the token budget then cosine decay for the rest. We optimize using AdamW ($\beta_1 = 0.9,\; \beta_2 = 0.999,$ weight‑decay $0.1$), with each update consuming $2^{16}$ tokens. Following the Chinchilla compute-optimal rule \citep{hoffmann2022trainingcomputeoptimallargelanguage}, we set the total token budget to $T_{base} \approx 20N$ tokens, where $N$ is the number of model non-embedding parameters. We compute the baseline update count as $U=\lceil T_{base}/2^{16}\rceil$ and keep $U$ identical across all \emph{Bi\mbox{-}Induct} curricula at a given scale to enforce iso-FLOPs. We monitor training loss and evaluate perplexity at the final checkpoint on a held-out split of the natural corpus (without synthetic snippets).

\textbf{Variants:}
We compare four variants, namely
\textsc{Baseline} (no snippets),
\textsc{Induction} (forward copy),
\textsc{Anti} (backward copy), and
\textsc{Balanced} (coin flip per injection).

\textbf{Metrics and guardrails:}
We assess Bi\mbox{-}Induct along three complementary axes: (i) \emph{downstream ICL performance} on standard few\mbox{-}shot benchmarks; (ii) \emph{mechanistic telemetry} that targets the intended circuit (induction and anti\mbox{-}induction heads); and (iii) \emph{quality} guardrails. Benchmarks are run few\mbox{-}shot (3\mbox{-}shot by default); we also include function\mbox{-}style tasks from the \citet{todd2024functionvectorslargelanguage} suite at 10\mbox{-}shot, evaluated with HITS@1 accuracy to stress simple copy and selection behaviors. For mechanism evidence, we compute per\mbox{-}head copy scores and, at the final checkpoint, report the \textbf{top 2\%} of heads per layer (and their concentration) for both induction and anti\mbox{-}induction, contrasting Bi\mbox{-}Induct curricula with the baseline. As a guardrail, we report held\mbox{-}out language modeling perplexity (PPL). Table~\ref{tab:metrics-summary} summarizes each metric and its preferred direction; for detailed definitions see Appendix~\ref{app:metrics-details}.

\begin{table}[t]
\centering
\small
\setlength{\tabcolsep}{4pt}
\renewcommand{\arraystretch}{1.0}
\caption{Summary of outcome metrics and guardrails. Full definitions in Appendix~\ref{app:metrics-details}.}
\label{tab:metrics-summary}
\resizebox{\textwidth}{!}{
\begin{tabular}{llp{7.4cm}c}
\toprule
\textbf{Family} & \textbf{Metric} & \textbf{What it measures / protocol} & \textbf{Better} \\
\midrule
Standard LM benchmarks &
ICL composite (macro) &
Unweighted mean across 3-shot tasks (MMLU, ARC-C, BoolQ, LAMBADA, PIQA; plus others where used). Accuracy or exact match per task; averages over demo seeds. &
$\uparrow$ \\
&
Per-task scores &
Per-benchmark few-shot evaluation (3-shot by default). Report mean over seeds with the benchmark’s standard metric (Acc or EM). &
$\uparrow$ \\
Function probes &
ICL composite (macro) &
Unweighted mean across ICL tasks probing string manipulation/selection (\texttt{capitalize\_*}, \texttt{next\_item}, \texttt{word\_length}, \texttt{alphabetically\_*}, \texttt{choose\_*}). Default 10-shot; metric is HITS@1 accuracy; seeds averaged. &
$\uparrow$ \\
&
Per-task scores &
Per-probe 10-shot (unless stated) with HITS@1 accuracy; seeds averaged. &
$\uparrow$ \\
Mechanistic telemetry &
Head copy score (top 2\% per layer) &
Per-head induction and anti-induction copy scores at the final checkpoint; report, for each layer, the top 2\% heads and their concentration to reveal circuit strength and specialization vs. baseline. &
$\uparrow$ \\
Quality &
Perplexity (held-out) &
PPL on a fixed 0.2\% \textsc{The Pile} validation slice (stable hash), same tokenizer and context across runs; mean over seeds at iso-FLOPs. &
$\downarrow$ \\
\bottomrule
\end{tabular}}
\end{table}

\textbf{Design lab at 0.13B:}
We use the \textbf{0.13B} model as a design lab to select the operating point for larger\mbox{-}scale runs. Unless noted otherwise, all ablations use a \textbf{2.6B} token budget, a \textbf{1024} context length, and a \textbf{linear anneal over the full budget}.

\begin{itemize}\itemsep=0pt
  \item \textbf{Span length ($L$):} We sweep $L\!\in\!\{5,20,500\}$ under Bi\mbox{-}Induct and find that $L{=}20$ offers the best trade\mbox{-}off between few\mbox{-}shot ICL performance and held\mbox{-}out perplexity. For detailed analysis see Section~\ref{app:span_len_sweep}, Appendix~\ref{app:ablation_study}.
  \item \textbf{Mix ratio ($m_0$):} With span fixed at $L{=}20$, we sweep the initial mix ratio $m_0\!\in\!\{25\%,50\%,100\%\}$ (linearly annealed to zero over the full budget). We select \textbf{50\%} because it yields stronger and more concentrated induction\mbox{-}head activity (top\mbox{-}2\% concentration by layer) while maintaining competitive ICL and PPL; see also Section~\ref{app:mix_ratio_sweep}, Appendix~\ref{app:ablation_study}.
\end{itemize}

\textbf{Summary and choice for scaling:}
For the main experiments across \textbf{0.13B}, \textbf{0.5B}, and \textbf{1B}, we adopt \textbf{$L{=}20$} and \textbf{$m_0{=}50\%$}, which we linearly anneal to $0$ over each model’s full training token budget (anneal horizon $T_a = T_{\text{base}}$)\footnote{Annealing over the full budget avoids introducing a second scheduling timescale, reduces re-tuning, and keeps the recipe portable across scales.}.

\section{Natural-Only vs. Directional Copy-Snippet Mix (Iso-FLOPs)}\label{sec:results}

\subsection{Downstream ICL Performance}\label{sec:icl_performance}
We evaluate two groups of tasks under iso\mbox{-}FLOPs and average over three seeds: (i) \emph{standard LM benchmarks} (14 tasks; e.g., MMLU, BBH, GSM8K, ARC\mbox{-}C, HellaSwag) and (ii) \emph{function\mbox{-}style probes} from the \citet{todd2024functionvectorslargelanguage} suite (19 tasks). For the full inventory of \emph{both} groups see Table~\ref{tab:benchmarks-detail}, Appendix~\ref{app:metrics-details}. We report macro\mbox{-}averages per group here and provide per\mbox{-}task scores in Table~\ref{tab:main_icl_performance}, Appendix~\ref{app:icl_performance}.

\begin{figure*}[t]
  \centering

  \begin{subfigure}[t]{\textwidth}
    \centering
    \includegraphics[width=0.7\textwidth]{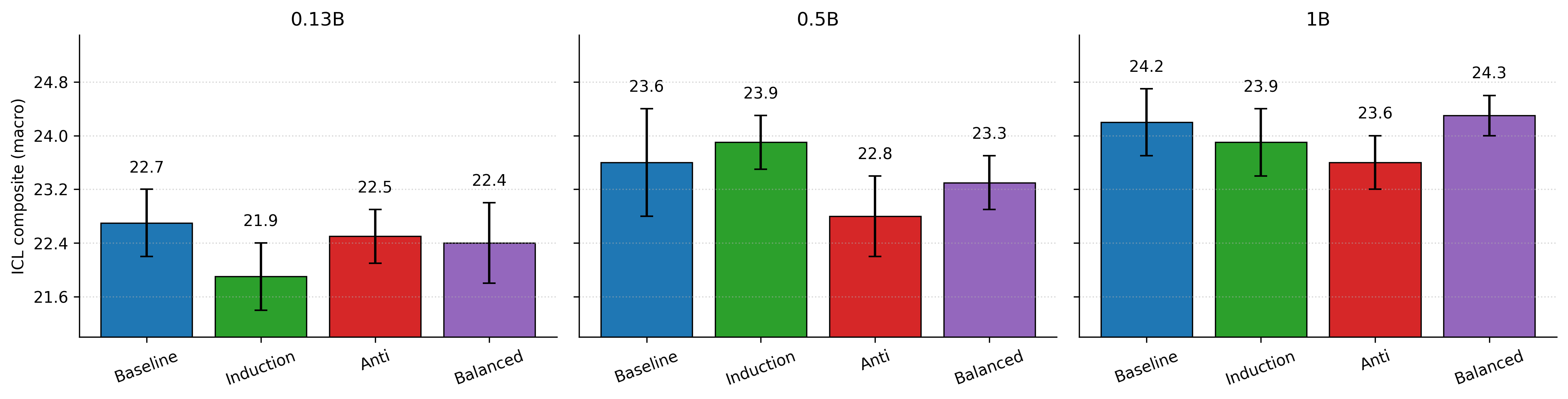}
    \caption{}
    \label{fig:icl-lm-eval}
  \end{subfigure}

  \vspace{0.6em}

  \begin{subfigure}[t]{\textwidth}
    \centering
    \includegraphics[width=0.7\textwidth]{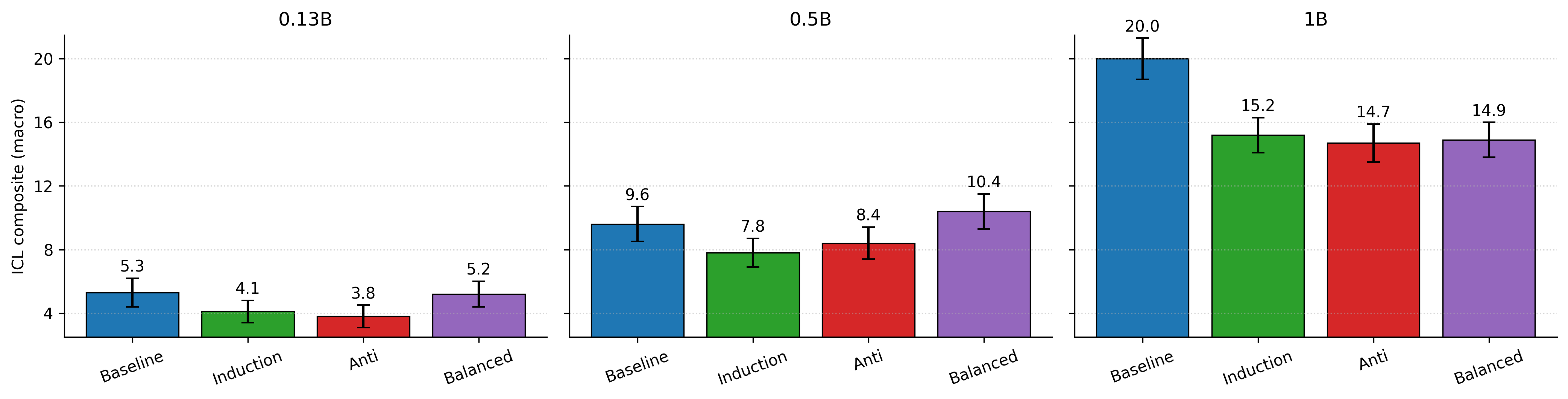}
    \caption{}
    \label{fig:icl-todd-suit}
  \end{subfigure}

  \caption{ICL Composite (macro) across two evaluation families: (a) Standard LM benchmarks; (b) \citet{todd2024functionvectorslargelanguage}'s function-probe suite. Each panel groups by model size (0.13B, 0.5B, 1B), bar colors by training regime (Baseline, Induction, Anti, Balanced); error bars show $\pm$1~s.d. For per-task results see Appendix~\ref{app:icl_performance}, Table~\ref{tab:main_icl_performance}.}
  \label{fig:icl-combined}
\end{figure*}

\textbf{Standard LM benchmarks:}
Figure~\ref{fig:icl-lm-eval} reports 5\mbox{-}shot macro\mbox{-}averages across 14 benchmarks. At each scale, at least one \emph{Bi\mbox{-}Induct} variant matches or very slightly exceeds the natural\mbox{-}only baseline: at 0.13B, \emph{Anti} is closest (22.5 vs.\ 22.7); at 0.5B, \emph{Induction} leads (23.9 vs.\ 23.6); at 1B, \emph{Balanced} is on par or marginally higher (24.3 vs.\ 24.2). Within measurement noise, copy-snippet curricula are \emph{largely performance-neutral} on standard LM benchmarks, but this neutrality is misleading in isolation: function-style probes and targeted ablations reveal important differences in whether induction becomes behaviorally load-bearing.

\textbf{Function probes \citep{todd2024functionvectorslargelanguage}:}
Figure~\ref{fig:icl-todd-suit} shows 10\mbox{-}shot macro\mbox{-}averages over 19 probes. At 0.13B and 0.5B, \emph{Bi\mbox{-}Induct} variants are comparable to baseline; at 1B, the natural\mbox{-}only baseline is clearly stronger across the suite. 

Robustness checks (1\mbox{-}shot evaluation, label permutation, and shifting the decision rule from \textsc{HITS@}1 to \textsc{HITS@}3) shift absolute scores, but preserve the cross\mbox{-}regime ordering; for details see Appendix~\ref{app:icl_robustness}.

\subsection{Mechanistic Telemetry}
Figure~\ref{fig:1b_500m_125m_induction_anti_scores} visualizes layerwise copy scores for the top 2\% attention heads per layer (with a floor of one head per layer to avoid sampling artifacts). Three 
clear patterns emerge.

\emph{(i) Layerwise localization:}
At 0.13B and 0.5B, Bi-Induct variants show earlier induction-head emergence than the baseline (by roughly 3 and 2 layers, respectively). At 1B, the trend reverses: the baseline is the first to form clear induction peaks (around layers 10-11) and its early peaks are higher than any Bi-Induct variant. For anti-induction, absolute scores are small at all scales; the largest peak we observe is $\approx 0.04$ at 0.5B (Induction curriculum), followed by $\approx 0.02$ at 1B (Baseline). In keeping with \citet{veitsman2025borntransformertransformer}, forward-induction heads dominate in pretrained LMs; in our runs, even the \emph{Anti} curriculum did not materially increase anti-induction copy scores. \emph{Notably, the strongest induction activity is concentrated in mid layers, in keeping with prior observations of where induction heads typically emerge} \citep{olsson2022context}.

\emph{(ii) Peak strength:}
The maximum normalized induction score reaches values close to 1.0 at 0.13B and 0.5B, but stays well below 0.5 at 1B. Thus, even when induction emerges early at 1B (Baseline), its strongest heads are less polarized than at smaller scales.

\emph{(iii) Spread:}
We count heads with a positive copy score among those selected by our per-layer top-2\% criterion (Section~\ref{sec:copy_scores}). Using the best-performing \emph{Bi\mbox{-}Induct} variant at each scale for a like-for-like comparison (Balanced at 0.13B/0.5B; Induction at 1B) we observe: for 0.13B, Baseline 3 vs.\ Balanced 5; for 0.5B, Baseline 7 vs.\ Balanced 8; for 1B, Baseline 12 vs.\ Induction 6. In short, \emph{Bi\mbox{-}Induct} tends to yield earlier and slightly broader induction activity at 0.13B/0.5B, whereas at 1B the natural-only Baseline shows the broader spread.

\paragraph{Link to ICL performance (telemetry vs.\ causal reliance):}
The layerwise copy telemetry above is correlational: it localizes where induction-like signatures concentrate, but it does not by itself establish that those heads are \emph{necessary} for ICL.
We therefore make a \emph{falsifiable} distinction between \emph{emergence} (telemetry) and \emph{load-bearing} reliance (ablation sensitivity): if stronger/earlier signatures implied necessity, then removing the top induction heads would hurt most precisely in the conditions with the strongest telemetry peaks.
Instead, we interpret telemetry jointly with behavioral results and the targeted ablations below.

On standard LM benchmarks (Figure~\ref{fig:icl-lm-eval}), macro ICL composites are similar across curricula at each scale. In particular, even when Bi\mbox{-}Induct shifts induction signatures toward earlier layers at 0.13B/0.5B, endpoints remain comparable to Baseline; and at 1B, despite the Baseline exhibiting an earlier depth-peak and a broader set of weakly-positive induction heads, Bi\mbox{-}Induct remains broadly competitive on these benchmarks. One plausible explanation is that many of these tasks are knowledge- and calibration-heavy, and larger models can route a substantial fraction of prediction mass through FFN/residual pathways rather than a small set of copy heads (consistent with evidence that FFNs act as key-value memories) \citep{geva2021transformerfeedforwardlayerskeyvalue}. We treat this as a hypothesis rather than a mechanistic claim.

In contrast, on \citet{todd2024functionvectorslargelanguage}'s function-style suite (Figure~\ref{fig:icl-todd-suit}), the 1B Baseline shows a clear performance advantage, consistent with these probes being more sensitive to explicit copy-head computation. Notably, this advantage does \emph{not} require that the Baseline have more uniformly high-scoring heads: visually (Figure~\ref{fig:1b_500m_125m_induction_anti_scores}), the Baseline is more \emph{concentrated} (dominated by a small subset, including a single prominent head), whereas Bi\mbox{-}Induct variants can exhibit multiple prominent heads yet still fall short on these probes.

Causally, ablating the top-2\% induction heads per layer decreases ICL composites more than ablating an equal number of random heads (Table~\ref{tab:percent_head_ablate_drop}). (Occasional small gains from random ablations are consistent with noise/regularization effects.)
The relative drop is largest for the natural-only Baseline, while Bi\mbox{-}Induct variants degrade less, consistent with more redundancy or a more distributed implementation of induction-like behavior.
A plausible explanation is \emph{front-loaded (in-training) recruitment}: because Bi-Induct injects copy cues early and then anneals them away, it may recruit multiple induction-capable heads that persist as redundant ``backup'' pathways rather than becoming a single load-bearing bottleneck. Distinguishing early recruitment from late diffusion would require tracking copy telemetry across checkpoints.
Detailed per-task comparisons for clean runs vs.\ induction-head and random-head ablations appear in Table~\ref{tab:head-ablate-all}, Appendix~\ref{app:head_ablation}.

\begin{table}[t]
\centering
\small
\setlength{\tabcolsep}{7pt}
\renewcommand{\arraystretch}{1.0}
\caption{Percent change in the ICL composite on the function-probe suite of \cite{todd2024functionvectorslargelanguage} when ablating either the top-2\% highest-scoring induction heads per layer(\(\Delta_{\text{induct}}\)) or an equal number of random heads (\(\Delta_{\text{rand}}\)), each measured relative to the model’s clean run. Negative values indicate a drop in accuracy; positive values indicate an improvement.}

\label{tab:percent_head_ablate_drop}
\resizebox{\textwidth}{!}{\begin{tabular}{l *{4}{cc}}
\toprule
& \multicolumn{2}{c}{\textbf{Baseline}}
& \multicolumn{2}{c}{\textbf{Induction}}
& \multicolumn{2}{c}{\textbf{Anti-induction}}
& \multicolumn{2}{c}{\textbf{Balanced}} \\
\cmidrule(lr){2-3}\cmidrule(lr){4-5}\cmidrule(lr){6-7}\cmidrule(lr){8-9}
\textbf{Model} & $\Delta_{\text{induct}}$ & $\Delta_{\text{rand}}$
               & $\Delta_{\text{induct}}$ & $\Delta_{\text{rand}}$
               & $\Delta_{\text{induct}} $& $\Delta_{\text{rand}}$
               & $\Delta_{\text{induct}}$ & $\Delta_{\text{rand}}$ \\
\midrule
0.13B & \(-22.6\%\) & \(+17.0\%\) & \(-4.9\%\) & \(+7.3\%\) & \(-5.3\%\) & \(+5.3\%\) & \(-19.2\%\) & \(0.0\%\) \\
0.5B  & \(-14.6\%\) & \(-4.2\%\)  & \(-10.3\%\) & \(+3.8\%\) & \(-8.3\%\)  & \(-1.2\%\) & \(-12.5\%\) & \(0.0\%\) \\
1B    & \(-19.5\%\) & \(-4.0\%\)  & \(-14.5\%\) & \(-2.6\%\) & \(-12.9\%\) & \(+0.7\%\) & \(-8.7\%\)  & \(-4.0\%\) \\
\bottomrule
\end{tabular}}
\end{table}

\paragraph{What might drive the 1B-scale behavior?}
Two non-exclusive factors may contribute, both consistent with prior literature: 
(1) \emph{Width-dilution}: 
the 1B model has the same depth as the 0.5B model but a larger hidden size and more attention heads (24) per layer. As a result, copy behavior may be spread across more heads, reducing the peak score of any single head even if the behavior is present.\footnote{Prior work finds substantial head redundancy and small subsets of specialized/important heads \citep{voita-etal-2019-analyzing, michel2019sixteenheadsreallybetter, olah2020zoom}. 
In wider models, this redundancy may spread copy behavior across more heads, lowering any single head’s score (a speculative `Width-dilution' effect).}
(2) 
\emph{Pathway shift}: larger models may increasingly leverage FFN and residual pathways, reducing reliance on localized, 
high-scoring induction heads \citep{geva2021transformerfeedforwardlayerskeyvalue}.

Overall, the mechanistic readout suggests that \textit{Bi-Induct consistently accelerates and broadens induction activity at smaller scales}, but \textit{at 1B the natural-only baseline exhibits earlier and broader consolidation of induction}, 
aligning with its stronger performance on the \cite{todd2024functionvectorslargelanguage} suite. In contrast, standard LM few-shot appear largely insensitive to these differences, likely due to the availability of alternative computation pathways.

\begin{figure*}[t]
  \centering
  \begin{subfigure}[t]{0.32\linewidth}
    \centering
    \includegraphics[width=\linewidth]{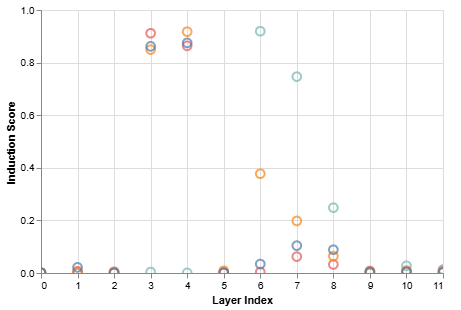}
    \caption{}
  \end{subfigure}\hfill
  \begin{subfigure}[t]{0.32\linewidth}
    \centering
    \includegraphics[width=\linewidth]{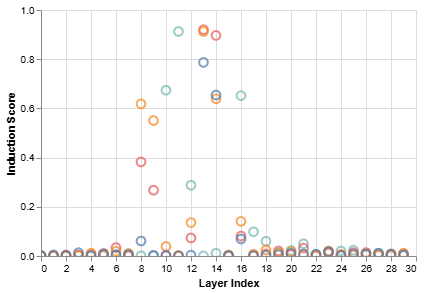}
    \caption{}
  \end{subfigure}\hfill
  \begin{subfigure}[t]{0.32\linewidth}
    \centering
    \includegraphics[width=\linewidth]{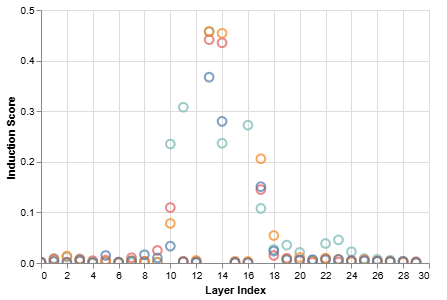}
    \caption{}
  \end{subfigure}

  \vspace{0.5em}

  \begin{subfigure}[t]{0.32\linewidth}
    \centering
    \includegraphics[width=\linewidth]{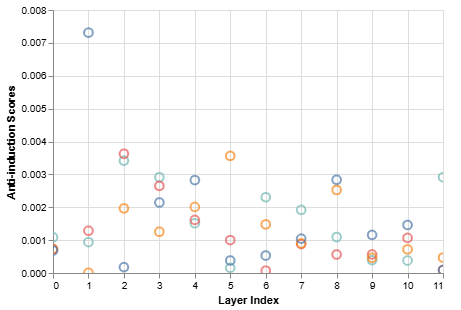}
    \caption{}
  \end{subfigure}\hfill
  \begin{subfigure}[t]{0.32\linewidth}
    \centering
    \includegraphics[width=\linewidth]{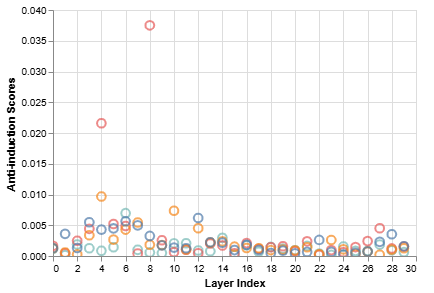}
    \caption{}
  \end{subfigure}\hfill
  \begin{subfigure}[t]{0.32\linewidth}
    \centering
    \includegraphics[width=\linewidth]{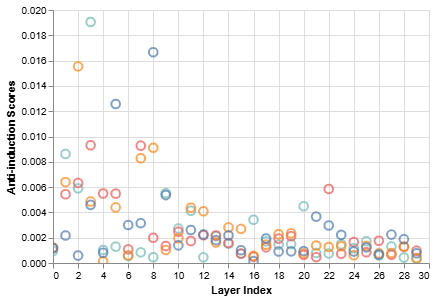} 
    \caption{}
  \end{subfigure}

  \par\vspace{0.25em}
  \begin{minipage}{0.98\linewidth}
    \centering
    \begin{tabular}{p{0.32\linewidth} p{0.32\linewidth} p{0.32\linewidth}}
      \centering 0.13B & \centering 0.5B & \centering 1B \\
    \end{tabular}
  \end{minipage}

   \newcommand{\legendcircsize}{2.5pt} 
\vspace{0.6em}
\begin{minipage}{0.95\linewidth}
  \centering
  \setlength{\tabcolsep}{10pt}
  \begin{tabular}{cccc}
    
    \tikz\draw[line width=0.9pt, legBaseline] (0,0) circle (\legendcircsize);~ Baseline &
    \tikz\draw[line width=0.9pt, legInduction]  (0,0) circle (\legendcircsize);~ Induction &
    \tikz\draw[line width=0.9pt, legAnti]  (0,0) circle (\legendcircsize);~ Anti &
    \tikz\draw[line width=0.9pt, legBalanced]      (0,0) circle (\legendcircsize);~ Balanced
  \end{tabular}
\end{minipage}
  
  \caption{Layer-wise copy-head telemetry. Top row: induction scores; bottom row: anti-induction scores. For each layer we plot the best-scoring head (top 2\% by score with a floor of one head per layer), averaged over three seeds, for the 0.13B, 0.5B, and 1B models. Head counts for each model are given in Table~\ref{tab:model-presets}.}
  \label{fig:1b_500m_125m_induction_anti_scores}
\end{figure*}

\subsection{Guardrail: Language Modeling Perplexity}
Table~\ref{tab:ppl-by-model-and-curriculum} reports held-out perplexity (mean over three seeds) under iso-FLOPs for each model size. We observe a consistent pattern: the \emph{perplexity gap} between copy-snippet curricula and the baseline \emph{shrinks with scale}, suggesting that larger models can absorb a small synthetic perturbation of the training stream without lasting calibration cost. Qualitatively, this trend is consistent with benign overfitting/double-descent intuitions: larger models can accommodate mild training perturbations while continuing to improve test loss~\citep{nakkiran2019deepdoubledescentbigger}. 

These observations show that a light, annealed Bi-Induct schedule keeps the perplexity penalty bounded and shrinking with scale, but the natural-only baseline remains consistently better on held-out perplexity at every scale.
\begin{table}[t]
\centering
\small
\setlength{\tabcolsep}{5pt}
\renewcommand{\arraystretch}{1.0}
\caption{Held-out perplexity (PPL $\downarrow$) on the fixed \textsc{The Pile} eval split at iso-FLOPs.
Values are averaged over three seeds. For each model size, curricula use a \textbf{50\%} mix ratio linearly annealed over the full training budget.}
\label{tab:ppl-by-model-and-curriculum}
\begin{tabular}{lccc}
\toprule
\textbf{Curriculum} & \textbf{0.13B} & \textbf{0.5B} & \textbf{1B} \\
\midrule
Induction       & \textit{25.8} & \textit{17.9} & \textit{14.9} \\
Anti\mbox{-}induction  & \textit{26.2} & \textit{18.2} & \textit{14.9} \\
Balanced & \textit{26.2} & \textit{18.2} & \textit{14.9} \\
\midrule
Baseline & 21.8 & 16.0 & 14.1 \\
\bottomrule
\end{tabular}
\end{table}

\subsection{Implications for Data-Centric Foundation Model Design} \label{sec:data_fm_implications}
Taken together, these results suggest a practical evaluation principle for synthetic data interventions. Under matched compute, it is not sufficient to show that a data rewrite amplifies a target internal signature. A useful intervention should also improve or at least preserve downstream behavior, avoid unnecessary degradation in natural-data modeling quality, and ideally make the targeted computation more causally necessary rather than merely more visible. In our case, Bi-Induct succeeds at signature amplification but not at consistently creating load-bearing ICL circuitry, especially relative to natural-only training at 1B.

\section{Conclusion and Future Work}\label{sec:conclusion_future_work}
We asked a single matched-compute question: for in-context learning, is it more effective to pretrain purely on natural text, or to allocate a small early-training budget to synthetic directional copy snippets that explicitly exercise the induction circuit? Using Bi-Induct (forward, backward, or balanced injections), we evaluated 0.13B–1B models with three complementary readouts: few-shot ICL performance, head-level copy telemetry, and held-out perplexity.

Bi-Induct reliably amplifies induction signatures, but it does not consistently improve few-shot ICL, and at 1B the natural-only baseline remains more load-bearing. For data-centric foundation model design, the broader lesson is methodological. Synthetic data interventions should not be evaluated only by whether they amplify a desired mechanistic signature. They should also be tested for whether that mechanism becomes \textbf{causally necessary} for the downstream behaviors of interest, and whether the intervention preserves natural-language modeling quality. In our study, Bi-Induct is best understood not as a generally superior curriculum, but as a controlled example showing that \textbf{signature amplification alone is too weak a success criterion}.

Several directions could extend this study. First, richer synthetic signals that incorporate semantic or linguistic structure may better align with the mechanisms underlying real-world induction behavior than the minimal token-level snippets used here. Second, scaling the analysis to larger models and longer context windows would clarify whether the emergence–vs.–load-bearing distinction persists in regimes where long-context retrieval becomes more critical. Finally, extending this methodology to alternative architectures, such as linear-attention models, may provide insight into whether induction-like behaviors in those systems become structurally necessary for long-context reasoning or remain redundant artifacts of training.

\section*{Limitations}
Our conclusions are specific to a lightweight copy-based intervention under the conditions studied here. First, the injected snippets are token-level and intentionally minimal; they are not tied to richer linguistic or semantic structure, so our findings should not be read as ruling out more expressive mechanism-aware data rewrites. Second, our study uses final-checkpoint mechanistic analysis and models up to 1B parameters; broader generalization to larger scales remains open. Third, our runs use a context length of 1,024, so we do not claim that the same trade-offs must hold in substantially longer-context settings, where induction-like retrieval may matter more. Finally, because Bi-Induct replaces a fraction of natural data under iso-FLOPs, some of the observed trade-off may reflect natural-text displacement in addition to mechanistic redundancy.

These limitations sharpen, rather than weaken, the main takeaway: mechanism-aware data design should be assessed against a stricter standard than whether it makes a target circuit easier to measure. The key question is whether it makes that circuit matter.

\section*{Use of Large Language Models (LLMs)}
We used general-purpose large language models as assistive tools for \emph{writing} and \emph{typesetting}. Concretely: (i) LLMs helped draft and polish prose across multiple sections (e.g., Introduction, Related Work, and Conclusion), including line-level rewrites for clarity, grammar, and flow; and (ii) LLMs assisted with LaTeX boilerplate and table scaffolding (e.g., column definitions, \texttt{\textbackslash resizebox}, and \texttt{booktabs} structure) but did not determine the content of any table.

LLMs \textbf{were not} used to design experiments, analyze data, run code, generate results, or make scientific claims. All technical decisions, datasets, models, and analyses originated from the authors. Every LLM suggestion was reviewed, edited, and verified by the authors; all references and factual statements were cross-checked against primary sources.

\bibliography{iclr2026_conference}

@article{olsson2022context,
   title={In-context Learning and Induction Heads},
   author={Olsson, Catherine and Elhage, Nelson and Nanda, Neel and Joseph, Nicholas and DasSarma, Nova and Henighan, Tom and Mann, Ben and Askell, Amanda and Bai, Yuntao and Chen, Anna and Conerly, Tom and Drain, Dawn and Ganguli, Deep and Hatfield-Dodds, Zac and Hernandez, Danny and Johnston, Scott and Jones, Andy and Kernion, Jackson and Lovitt, Liane and Ndousse, Kamal and Amodei, Dario and Brown, Tom and Clark, Jack and Kaplan, Jared and McCandlish, Sam and Olah, Chris},
   year={2022},
   journal={Transformer Circuits Thread},
   note={https://transformer-circuits.pub/2022/in-context-learning-and-induction-heads/index.html}
}

@misc{chen2024unveilinginductionheadsprovable,
      title={Unveiling Induction Heads: Provable Training Dynamics and Feature Learning in Transformers}, 
      author={Siyu Chen and Heejune Sheen and Tianhao Wang and Zhuoran Yang},
      year={2024},
      eprint={2409.10559},
      archivePrefix={arXiv},
      primaryClass={cs.LG},
      url={https://arxiv.org/abs/2409.10559}, 
}

@inproceedings{
edelman2024the,
title={The Evolution of Statistical Induction Heads: In-Context Learning Markov Chains},
author={Ezra Edelman and Nikolaos Tsilivis and Benjamin L. Edelman and eran malach and Surbhi Goel},
booktitle={The Thirty-eighth Annual Conference on Neural Information Processing Systems},
year={2024},
url={https://openreview.net/forum?id=qaRT6QTIqJ}
}

@misc{zisman2025ngraminductionheadsincontext,
      title={N-Gram Induction Heads for In-Context RL: Improving Stability and Reducing Data Needs}, 
      author={Ilya Zisman and Alexander Nikulin and Viacheslav Sinii and Denis Tarasov and Nikita Lyubaykin and Andrei Polubarov and Igor Kiselev and Vladislav Kurenkov},
      year={2025},
      eprint={2411.01958},
      archivePrefix={arXiv},
      primaryClass={cs.LG},
      url={https://arxiv.org/abs/2411.01958}, 
}

@inproceedings{crosbie-shutova-2025-induction,
    title = "Induction Heads as an Essential Mechanism for Pattern Matching in In-context Learning",
    author = "Crosbie, Joy  and
      Shutova, Ekaterina",
    editor = "Chiruzzo, Luis  and
      Ritter, Alan  and
      Wang, Lu",
    booktitle = "Findings of the Association for Computational Linguistics: NAACL 2025",
    month = apr,
    year = "2025",
    address = "Albuquerque, New Mexico",
    publisher = "Association for Computational Linguistics",
    url = "https://aclanthology.org/2025.findings-naacl.283/",
    doi = "10.18653/v1/2025.findings-naacl.283",
    pages = "5034--5096",
    ISBN = "979-8-89176-195-7"
}

@misc{nanda2022transformerlens,
    title = {TransformerLens},
    author = {Neel Nanda and Joseph Bloom},
    year = {2022},
    howpublished = {\url{https://github.com/TransformerLensOrg/TransformerLens}},
}

@misc{biderman2023pythiasuiteanalyzinglarge,
      title={Pythia: A Suite for Analyzing Large Language Models Across Training and Scaling}, 
      author={Stella Biderman and Hailey Schoelkopf and Quentin Anthony and Herbie Bradley and Kyle O'Brien and Eric Hallahan and Mohammad Aflah Khan and Shivanshu Purohit and USVSN Sai Prashanth and Edward Raff and Aviya Skowron and Lintang Sutawika and Oskar van der Wal},
      year={2023},
      eprint={2304.01373},
      archivePrefix={arXiv},
      primaryClass={cs.CL},
      url={https://arxiv.org/abs/2304.01373}, 
}

@misc{mcdougall2023copysuppressioncomprehensivelyunderstanding,
      title={Copy Suppression: Comprehensively Understanding an Attention Head}, 
      author={Callum McDougall and Arthur Conmy and Cody Rushing and Thomas McGrath and Neel Nanda},
      year={2023},
      eprint={2310.04625},
      archivePrefix={arXiv},
      primaryClass={cs.LG},
      url={https://arxiv.org/abs/2310.04625}, 
}

@misc{wang2025inductionheadtoxicitymechanistically,
      title={Induction Head Toxicity Mechanistically Explains Repetition Curse in Large Language Models}, 
      author={Shuxun Wang and Qingyu Yin and Chak Tou Leong and Qiang Zhang and Linyi Yang},
      year={2025},
      eprint={2505.13514},
      archivePrefix={arXiv},
      primaryClass={cs.CL},
      url={https://arxiv.org/abs/2505.13514}, 
}

@misc{gurnee2023findingneuronshaystackcase,
      title={Finding Neurons in a Haystack: Case Studies with Sparse Probing}, 
      author={Wes Gurnee and Neel Nanda and Matthew Pauly and Katherine Harvey and Dmitrii Troitskii and Dimitris Bertsimas},
      year={2023},
      eprint={2305.01610},
      archivePrefix={arXiv},
      primaryClass={cs.LG},
      url={https://arxiv.org/abs/2305.01610}, 
}

@misc{zhang2024bestpracticesactivationpatching,
      title={Towards Best Practices of Activation Patching in Language Models: Metrics and Methods}, 
      author={Fred Zhang and Neel Nanda},
      year={2024},
      eprint={2309.16042},
      archivePrefix={arXiv},
      primaryClass={cs.LG},
      url={https://arxiv.org/abs/2309.16042}, 
}

@misc{cunningham2023sparseautoencodershighlyinterpretable,
      title={Sparse Autoencoders Find Highly Interpretable Features in Language Models}, 
      author={Hoagy Cunningham and Aidan Ewart and Logan Riggs and Robert Huben and Lee Sharkey},
      year={2023},
      eprint={2309.08600},
      archivePrefix={arXiv},
      primaryClass={cs.LG},
      url={https://arxiv.org/abs/2309.08600}, 
}

@misc{veitsman2025borntransformertransformer,
      title={Born a Transformer -- Always a Transformer?}, 
      author={Yana Veitsman and Mayank Jobanputra and Yash Sarrof and Aleksandra Bakalova and Vera Demberg and Ellie Pavlick and Michael Hahn},
      year={2025},
      eprint={2505.21785},
      archivePrefix={arXiv},
      primaryClass={cs.LG},
      url={https://arxiv.org/abs/2505.21785}, 
}

@misc{kim2025taskdiversityshortensicl,
      title={Task Diversity Shortens the ICL Plateau}, 
      author={Jaeyeon Kim and Sehyun Kwon and Joo Young Choi and Jongho Park and Jaewoong Cho and Jason D. Lee and Ernest K. Ryu},
      year={2025},
      eprint={2410.05448},
      archivePrefix={arXiv},
      primaryClass={cs.LG},
      url={https://arxiv.org/abs/2410.05448}, 
}

@misc{gloeckle2024betterfasterlarge,
      title={Better \& Faster Large Language Models via Multi-token Prediction}, 
      author={Fabian Gloeckle and Badr Youbi Idrissi and Baptiste Rozière and David Lopez-Paz and Gabriel Synnaeve},
      year={2024},
      eprint={2404.19737},
      archivePrefix={arXiv},
      primaryClass={cs.CL},
      url={https://arxiv.org/abs/2404.19737}, 
}

@misc{maini2024rephrasingwebrecipecompute,
      title={Rephrasing the Web: A Recipe for Compute and Data-Efficient Language Modeling}, 
      author={Pratyush Maini and Skyler Seto and He Bai and David Grangier and Yizhe Zhang and Navdeep Jaitly},
      year={2024},
      eprint={2401.16380},
      archivePrefix={arXiv},
      primaryClass={cs.CL},
      url={https://arxiv.org/abs/2401.16380}, 
}

@misc{nguyen2025recyclingwebmethodenhance,
      title={Recycling the Web: A Method to Enhance Pre-training Data Quality and Quantity for Language Models}, 
      author={Thao Nguyen and Yang Li and Olga Golovneva and Luke Zettlemoyer and Sewoong Oh and Ludwig Schmidt and Xian Li},
      year={2025},
      eprint={2506.04689},
      archivePrefix={arXiv},
      primaryClass={cs.CL},
      url={https://arxiv.org/abs/2506.04689}, 
}

@misc{fujii2025rewritingpretrainingdataboosts,
      title={Rewriting Pre-Training Data Boosts LLM Performance in Math and Code}, 
      author={Kazuki Fujii and Yukito Tajima and Sakae Mizuki and Hinari Shimada and Taihei Shiotani and Koshiro Saito and Masanari Ohi and Masaki Kawamura and Taishi Nakamura and Takumi Okamoto and Shigeki Ishida and Kakeru Hattori and Youmi Ma and Hiroya Takamura and Rio Yokota and Naoaki Okazaki},
      year={2025},
      eprint={2505.02881},
      archivePrefix={arXiv},
      primaryClass={cs.LG},
      url={https://arxiv.org/abs/2505.02881}, 
}

@misc{penedo2023refinedwebdatasetfalconllm,
      title={The RefinedWeb Dataset for Falcon LLM: Outperforming Curated Corpora with Web Data, and Web Data Only}, 
      author={Guilherme Penedo and Quentin Malartic and Daniel Hesslow and Ruxandra Cojocaru and Alessandro Cappelli and Hamza Alobeidli and Baptiste Pannier and Ebtesam Almazrouei and Julien Launay},
      year={2023},
      eprint={2306.01116},
      archivePrefix={arXiv},
      primaryClass={cs.CL},
      url={https://arxiv.org/abs/2306.01116}, 
}

@misc{gao2020pile800gbdatasetdiverse,
      title={The Pile: An 800GB Dataset of Diverse Text for Language Modeling}, 
      author={Leo Gao and Stella Biderman and Sid Black and Laurence Golding and Travis Hoppe and Charles Foster and Jason Phang and Horace He and Anish Thite and Noa Nabeshima and Shawn Presser and Connor Leahy},
      year={2020},
      eprint={2101.00027},
      archivePrefix={arXiv},
      primaryClass={cs.CL},
      url={https://arxiv.org/abs/2101.00027}, 
}

@misc{hoffmann2022trainingcomputeoptimallargelanguage,
      title={Training Compute-Optimal Large Language Models}, 
      author={Jordan Hoffmann and Sebastian Borgeaud and Arthur Mensch and Elena Buchatskaya and Trevor Cai and Eliza Rutherford and Diego de Las Casas and Lisa Anne Hendricks and Johannes Welbl and Aidan Clark and Tom Hennigan and Eric Noland and Katie Millican and George van den Driessche and Bogdan Damoc and Aurelia Guy and Simon Osindero and Karen Simonyan and Erich Elsen and Jack W. Rae and Oriol Vinyals and Laurent Sifre},
      year={2022},
      eprint={2203.15556},
      archivePrefix={arXiv},
      primaryClass={cs.CL},
      url={https://arxiv.org/abs/2203.15556}, 
}

@misc{nakkiran2019deepdoubledescentbigger,
      title={Deep Double Descent: Where Bigger Models and More Data Hurt}, 
      author={Preetum Nakkiran and Gal Kaplun and Yamini Bansal and Tristan Yang and Boaz Barak and Ilya Sutskever},
      year={2019},
      eprint={1912.02292},
      archivePrefix={arXiv},
      primaryClass={cs.LG},
      url={https://arxiv.org/abs/1912.02292}, 
}

@misc{todd2024functionvectorslargelanguage,
      title={Function Vectors in Large Language Models}, 
      author={Eric Todd and Millicent L. Li and Arnab Sen Sharma and Aaron Mueller and Byron C. Wallace and David Bau},
      year={2024},
      eprint={2310.15213},
      archivePrefix={arXiv},
      primaryClass={cs.CL},
      url={https://arxiv.org/abs/2310.15213}, 
}

@misc{yin2025attentionheadsmatterincontext,
      title={Which Attention Heads Matter for In-Context Learning?}, 
      author={Kayo Yin and Jacob Steinhardt},
      year={2025},
      eprint={2502.14010},
      archivePrefix={arXiv},
      primaryClass={cs.LG},
      url={https://arxiv.org/abs/2502.14010}, 
}

@misc{chan2022datadistributionalpropertiesdrive,
      title={Data Distributional Properties Drive Emergent In-Context Learning in Transformers}, 
      author={Stephanie C. Y. Chan and Adam Santoro and Andrew K. Lampinen and Jane X. Wang and Aaditya Singh and Pierre H. Richemond and Jay McClelland and Felix Hill},
      year={2022},
      eprint={2205.05055},
      archivePrefix={arXiv},
      primaryClass={cs.LG},
      url={https://arxiv.org/abs/2205.05055}, 
}

@misc{gu2023pretraininglearncontext,
      title={Pre-Training to Learn in Context}, 
      author={Yuxian Gu and Li Dong and Furu Wei and Minlie Huang},
      year={2023},
      eprint={2305.09137},
      archivePrefix={arXiv},
      primaryClass={cs.CL},
      url={https://arxiv.org/abs/2305.09137}, 
}

@misc{sanford2024transformersparallelcomputationlogarithmic,
      title={Transformers, parallel computation, and logarithmic depth}, 
      author={Clayton Sanford and Daniel Hsu and Matus Telgarsky},
      year={2024},
      eprint={2402.09268},
      archivePrefix={arXiv},
      primaryClass={cs.LG},
      url={https://arxiv.org/abs/2402.09268}, 
}

@article{hendryckstest2021,
  title={Measuring Massive Multitask Language Understanding},
  author={Dan Hendrycks and Collin Burns and Steven Basart and Andy Zou and Mantas Mazeika and Dawn Song and Jacob Steinhardt},
  journal={Proceedings of the International Conference on Learning Representations (ICLR)},
  year={2021}
}

@article{sakaguchi2019winogrande,
    title={WinoGrande: An Adversarial Winograd Schema Challenge at Scale},
    author={Sakaguchi, Keisuke and Bras, Ronan Le and Bhagavatula, Chandra and Choi, Yejin},
    journal={arXiv preprint arXiv:1907.10641},
    year={2019}
}

@article{suzgun2022challenging,
  title={Challenging BIG-Bench Tasks and Whether Chain-of-Thought Can Solve Them},
  author={Suzgun, Mirac and Scales, Nathan and Sch{\"a}rli, Nathanael and Gehrmann, Sebastian and Tay, Yi and Chung, Hyung Won and Chowdhery, Aakanksha and Le, Quoc V and Chi, Ed H and Zhou, Denny and and Wei, Jason},
  journal={arXiv preprint arXiv:2210.09261},
  year={2022}
}

@inproceedings{talmor-etal-2019-commonsenseqa,
    title = "{C}ommonsense{QA}: A Question Answering Challenge Targeting Commonsense Knowledge",
    author = "Talmor, Alon  and
      Herzig, Jonathan  and
      Lourie, Nicholas  and
      Berant, Jonathan",
    booktitle = "Proceedings of the 2019 Conference of the North {A}merican Chapter of the Association for Computational Linguistics: Human Language Technologies, Volume 1 (Long and Short Papers)",
    month = jun,
    year = "2019",
    address = "Minneapolis, Minnesota",
    publisher = "Association for Computational Linguistics",
    url = "https://aclanthology.org/N19-1421",
    doi = "10.18653/v1/N19-1421",
    pages = "4149--4158",
    archivePrefix = "arXiv",
    eprint        = "1811.00937",
    primaryClass  = "cs",
}

@inproceedings{Bisk2020,
    author = {Yonatan Bisk and Rowan Zellers and
            Ronan Le Bras and Jianfeng Gao
            and Yejin Choi},
    title = {PIQA: Reasoning about Physical Commonsense in
           Natural Language},
    booktitle = {Thirty-Fourth AAAI Conference on
               Artificial Intelligence},
    year = {2020},
}

@inproceedings{zellers2019hellaswag,
    title={HellaSwag: Can a Machine Really Finish Your Sentence?},
    author={Zellers, Rowan and Holtzman, Ari and Bisk, Yonatan and Farhadi, Ali and Choi, Yejin},
    booktitle ={Proceedings of the 57th Annual Meeting of the Association for Computational Linguistics},
    year={2019}
}

@InProceedings{JoshiTriviaQA2017,
    author = {Joshi, Mandar and Choi, Eunsol and Weld, Daniel S. and Zettlemoyer, Luke},
    title = {TriviaQA: A Large Scale Distantly Supervised Challenge Dataset for Reading Comprehension},
    booktitle = {Proceedings of the 55th Annual Meeting of the Association for Computational Linguistics},
    month = {July},
    year = {2017},
    address = {Vancouver, Canada},
    publisher = {Association for Computational Linguistics},
}

@inproceedings{OpenBookQA2018,
    title={Can a Suit of Armor Conduct Electricity? A New Dataset for Open Book Question Answering},
    author={Todor Mihaylov and Peter Clark and Tushar Khot and Ashish Sabharwal},
    booktitle={EMNLP},
    year={2018}
}

@article{Clark2018ThinkYH,
  title={Think you have Solved Question Answering? Try ARC, the AI2 Reasoning Challenge},
  author={Peter Clark and Isaac Cowhey and Oren Etzioni and Tushar Khot and Ashish Sabharwal and Carissa Schoenick and Oyvind Tafjord},
  journal={ArXiv},
  year={2018},
  volume={abs/1803.05457}
}

@misc{rein2023gpqa,
      title={GPQA: A Graduate-Level Google-Proof Q\&A Benchmark},
      author={David Rein and Betty Li Hou and Asa Cooper Stickland and Jackson Petty and Richard Yuanzhe Pang and Julien Dirani and Julian Michael and Samuel R. Bowman},
      year={2023},
      eprint={2311.12022},
      archivePrefix={arXiv},
      primaryClass={cs.AI}
}

@misc{cobbe2021training,
      title={Training Verifiers to Solve Math Word Problems},
      author={Karl Cobbe and Vineet Kosaraju and Mohammad Bavarian and Jacob Hilton and Reiichiro Nakano and Christopher Hesse and John Schulman},
      year={2021},
      eprint={2110.14168},
      archivePrefix={arXiv},
      primaryClass={cs.LG}
}

@misc{amini2019mathqa,
    title={MathQA: Towards Interpretable Math Word Problem Solving with Operation-Based Formalisms},
    author={Aida Amini and Saadia Gabriel and Peter Lin and Rik Koncel-Kedziorski and Yejin Choi and Hannaneh Hajishirzi},
    year={2019},
    eprint={1905.13319},
    archivePrefix={arXiv},
    primaryClass={cs.CL}
}

@misc{lambada, 
    author={Paperno, Denis and Kruszewski, Germán and Lazaridou, Angeliki and Pham, Quan Ngoc and Bernardi, Raffaella and Pezzelle, Sandro and Baroni, Marco and Boleda, Gemma and Fernández, Raquel}, 
    title={The LAMBADA dataset}, 
    DOI={10.5281/zenodo.2630551}, 
    publisher={Zenodo}, 
    year={2016}, 
    month={Aug} 
}

@inproceedings{clark-etal-2019-boolq,
    title = "{B}ool{Q}: Exploring the Surprising Difficulty of Natural Yes/No Questions",
    author = "Clark, Christopher  and
      Lee, Kenton  and
      Chang, Ming-Wei  and
      Kwiatkowski, Tom  and
      Collins, Michael  and
      Toutanova, Kristina",
    editor = "Burstein, Jill  and
      Doran, Christy  and
      Solorio, Thamar",
    booktitle = "Proceedings of the 2019 Conference of the North {A}merican Chapter of the Association for Computational Linguistics: Human Language Technologies, Volume 1 (Long and Short Papers)",
    month = jun,
    year = "2019",
    address = "Minneapolis, Minnesota",
    publisher = "Association for Computational Linguistics",
    url = "https://aclanthology.org/N19-1300/",
    doi = "10.18653/v1/N19-1300",
    pages = "2924--2936"
}

@misc{geva2021transformerfeedforwardlayerskeyvalue,
      title={Transformer Feed-Forward Layers Are Key-Value Memories}, 
      author={Mor Geva and Roei Schuster and Jonathan Berant and Omer Levy},
      year={2021},
      eprint={2012.14913},
      archivePrefix={arXiv},
      primaryClass={cs.CL},
      url={https://arxiv.org/abs/2012.14913}, 
}

@inproceedings{voita-etal-2019-analyzing,
    title = "Analyzing Multi-Head Self-Attention: Specialized Heads Do the Heavy Lifting, the Rest Can Be Pruned",
    author = "Voita, Elena  and
      Talbot, David  and
      Moiseev, Fedor  and
      Sennrich, Rico  and
      Titov, Ivan",
    editor = "Korhonen, Anna  and
      Traum, David  and
      M{\`a}rquez, Llu{\'i}s",
    booktitle = "Proceedings of the 57th Annual Meeting of the Association for Computational Linguistics",
    month = jul,
    year = "2019",
    address = "Florence, Italy",
    publisher = "Association for Computational Linguistics",
    url = "https://aclanthology.org/P19-1580/",
    doi = "10.18653/v1/P19-1580",
    pages = "5797--5808"
}

@misc{michel2019sixteenheadsreallybetter,
      title={Are Sixteen Heads Really Better than One?}, 
      author={Paul Michel and Omer Levy and Graham Neubig},
      year={2019},
      eprint={1905.10650},
      archivePrefix={arXiv},
      primaryClass={cs.CL},
      url={https://arxiv.org/abs/1905.10650}, 
}

@article{olah2020zoom,
  author    = {Chris Olah and Nick Cammarata and
               Ludwig Schubert and Gabriel Goh and
               Michael Petrov and Shan Carter},
  title     = {Zoom In: An Introduction to Circuits},
  journal   = {Distill},
  year      = {2020},
  volume    = {5},
  number    = {3},
  pages     = {e00024.001},
  month     = mar,
  doi       = {10.23915/distill.00024.001},
  url       = {https://distill.pub/2020/circuits/zoom-in/}
}

@misc{min2022rethinkingroledemonstrationsmakes,
      title={Rethinking the Role of Demonstrations: What Makes In-Context Learning Work?}, 
      author={Sewon Min and Xinxi Lyu and Ari Holtzman and Mikel Artetxe and Mike Lewis and Hannaneh Hajishirzi and Luke Zettlemoyer},
      year={2022},
      eprint={2202.12837},
      archivePrefix={arXiv},
      primaryClass={cs.CL},
      url={https://arxiv.org/abs/2202.12837}, 
}

@misc{wei2023largerlanguagemodelsincontext,
      title={Larger language models do in-context learning differently}, 
      author={Jerry Wei and Jason Wei and Yi Tay and Dustin Tran and Albert Webson and Yifeng Lu and Xinyun Chen and Hanxiao Liu and Da Huang and Denny Zhou and Tengyu Ma},
      year={2023},
      eprint={2303.03846},
      archivePrefix={arXiv},
      primaryClass={cs.CL},
      url={https://arxiv.org/abs/2303.03846}, 
}

@misc{zhao2021calibrateuseimprovingfewshot,
      title={Calibrate Before Use: Improving Few-Shot Performance of Language Models}, 
      author={Tony Z. Zhao and Eric Wallace and Shi Feng and Dan Klein and Sameer Singh},
      year={2021},
      eprint={2102.09690},
      archivePrefix={arXiv},
      primaryClass={cs.CL},
      url={https://arxiv.org/abs/2102.09690}, 
}

@misc{jiang2023mistral7b,
      title={Mistral 7B}, 
      author={Albert Q. Jiang and Alexandre Sablayrolles and Arthur Mensch and Chris Bamford and Devendra Singh Chaplot and Diego de las Casas and Florian Bressand and Gianna Lengyel and Guillaume Lample and Lucile Saulnier and Lélio Renard Lavaud and Marie-Anne Lachaux and Pierre Stock and Teven Le Scao and Thibaut Lavril and Thomas Wang and Timothée Lacroix and William El Sayed},
      year={2023},
      eprint={2310.06825},
      archivePrefix={arXiv},
      primaryClass={cs.CL},
      url={https://arxiv.org/abs/2310.06825}, 
}
\bibliographystyle{iclr2026_conference}

\appendix
\section{Glossary and Terminology}
\label{app:glossary}

This section defines the terms and metrics used throughout the paper. We group entries by theme for quick reference.

\subsection*{Copy-style circuits and interpretability}

\begin{description}
\item[Mechanistic interpretability]
  The study of internal circuits and features that give rise to behavior in neural networks. Typical tools include ablations/masking, activation patching, causal tracing, and sparse autoencoders.

  \item[Interpretability challenges]
  Practical difficulties include superposition (features sharing parameters), circuit non-uniqueness (multiple decompositions fit the data), intervention fragility (ablations can misattribute causality), scale transfer (circuits shift across sizes), and dataset confounds (spurious correlations masquerading as mechanisms).
  \item[Induction head / induction circuit]
  A two-head attention motif that implements forward copy: when a cue token reappears in the context, attention retrieves what followed the \emph{previous} occurrence and predicts it again. Empirically linked to few-shot pattern matching.

  \item[Anti-induction]
  The mirror of induction: backward copy. Given a repeated cue, the model predicts the \emph{preceding} token from an earlier occurrence (useful for reversal-style tasks and some code transforms).

  \item[Copy-suppression (negative) heads]
  Attention heads whose contribution reduces copying (e.g., down-weights repeated spans), often interacting with induction heads to prevent degenerate repetition.
\end{description}

\subsection*{Curriculum and data-rewrite terms}

\begin{description}
  \item[Data rewrite]
  Deliberate modification of a small fraction of pretraining tokens to teach a target algorithm (here, copy patterns) without changing the model architecture.

  \item[Bi-Induct]
  Our symmetric copy-style curriculum that injects synthetic snippets during pretraining in one of two directions: \texttt{induction} (forward copy) or \texttt{anti} (backward copy). Injection probability linearly anneals to zero.

  \item[Span length ($L$)]
  Number of random tokens in the snippet’s base span before duplication or reversal (e.g., $L\in\{5,20,100\}$).

  \item[(Initial)  Mix ratio]
  Initial probability of injecting a synthetic snippet before annealing (e.g., $25\%$).

  \item[Anneal tokens]
  The number of natural tokens over which the injection probability decays linearly to zero (e.g., the full $2.5$B-token budget).

  \item[“Balanced” variant]
  A coin-flip per injection between forward and backward copy. Used as an additional control in some ablations.
\end{description}

\subsection*{Compute and efficiency}

\begin{description}
    \item[Chinchilla (compute-optimal) budget]
    The token-parameter trade-off that minimizes validation loss at fixed compute for dense decoder-only LMs. Rule of thumb: a tokens-to-parameters ratio of \(\approx 20{:}1\), i.e., \(T \approx 20N\) (tokens \(T\), parameters \(N\)).


\end{description}

\subsection*{Evaluation endpoints}

\begin{description}
  \item[ICL benchmarks (few-shot endpoints)]
  Standard few-shot (\(k\ge1\)) tasks evaluated at the \emph{final} checkpoint (e.g., 3-shot MMLU, ARC-C, BoolQ, LAMBADA, PIQA). We aggregate with a macro-average as the \textbf{ICL composite}. These are the \emph{main} outcome metrics. \textit{Regarding the standard deviation (s.d.) of the ICL composite:} In Tables~\ref{tab:span_sweep_pile_0.13B} and~\ref{tab:mix_sweep_pile_0.13B}, we compute the s.d. of the per-seed composite across seeds, which is the appropriate uncertainty. Elsewhere, for brevity, we approximate the composite’s uncertainty by averaging per-task s.d.s computed across seeds; this is a readable proxy but not a pooled s.d. and it ignores cross-task covariance.
  \item[Cross-entropy and perplexity]
  Language-model loss on a held-out split of the pretraining corpus. Perplexity $\mathrm{PPL}=\exp(\mathrm{CE})$. Used as a quality and calibration guardrail.
\end{description}

\section{Metrics and guardrails: detailed definitions}
\label{app:metrics-details}

\subsection{Benchmarks and protocols}
\begin{table}[ht]
\centering
\small
\setlength{\tabcolsep}{6pt}
\renewcommand{\arraystretch}{1.12}
\caption{Benchmarks, evaluation metrics, and shot counts used to compute the ICL composite in Section~\ref{sec:icl_performance}.}
\label{tab:benchmarks-detail}
\begin{tabular}{lccp{6.7cm}}
\toprule
\textbf{Benchmark / Tasks} & \textbf{Metric} & \textbf{Shots} & \textbf{Notes} \\
\midrule
MMLU~\citep{hendryckstest2021} & Acc & 3 & 57 subject areas; standard 5\mbox{-}shot setup. \\
Winogrande~\citep{sakaguchi2019winogrande} & Acc & 3 & Commonsense coreference. \\
CommonSenseQA~\citep{talmor-etal-2019-commonsenseqa} & Acc & 3 & Multiple choice commonsense. \\
PIQA~\citep{Bisk2020} & Acc & 3 & Physical commonsense. \\
HellaSwag~\citep{zellers2019hellaswag} & Acc & 3 & Story completion. \\
TriviaQA\mbox{-}Wiki~\citep{JoshiTriviaQA2017} & EM & 3 & Open\mbox{-}domain QA, Wikipedia evidence. \\
BBH (CoT)~\citep{suzgun2022challenging} & EM & 3 & Few hard tasks with chain\mbox{-}of\mbox{-}thought prompts. \\
OpenBookQA~\citep{OpenBookQA2018} & Acc & 3 & Elementary science QA. \\
ARC\mbox{-}Challenge~\citep{Clark2018ThinkYH} & Acc & 3 & Difficult science questions. \\
GPQA~\citep{rein2023gpqa} & Acc & 3 & Graduate\mbox{-}level QA. \\
GSM\mbox{-}8K~\citep{cobbe2021training} & EM & 3 & Math word problems with short CoT. \\
MathQA~\citep{amini2019mathqa} & Acc & 3 & Programmatic math QA. \\
BoolQ~\citep{clark-etal-2019-boolq} & Acc & 3 & Yes/No reading comprehension. \\
LAMBADA (OpenAI)~\citep{lambada} & Acc & 3 & Cloze final\mbox{-}word prediction. \\
\midrule
\multicolumn{4}{l}{\emph{ From \citet{todd2024functionvectorslargelanguage} function-probe suite:}}\\
\addlinespace[2pt]
\texttt{capitalize} & HITS@1 Acc & 10 & Convert the entire input string to uppercase (e.g., ``hello'' $\to$ ``HELLO''). \\
\texttt{capitalize\_first\_letter} & HITS@1 Acc & 10 & Uppercase the first character only; leave the rest unchanged (``alpha'' $\to$ ``Alpha''). \\
\texttt{capitalize\_last\_letter} & HITS@1 Acc & 10 & Uppercase the final character only (``gamma'' $\to$ ``gammA''). \\
\texttt{lowercase\_first\_letter} & HITS@1 Acc & 10 & Lowercase the first character only (``Alpha'' $\to$ ``alpha''). \\
\texttt{lowercase\_last\_letter} & HITS@1 Acc & 10 & Lowercase the final character only (``GammA'' $\to$ ``Gamma''). \\
\texttt{next\_capital\_letter} & HITS@1 Acc & 10 & Map an uppercase letter to its successor in the alphabet (e.g., \texttt{A}$\to$\texttt{B}; wraparound optional). \\
\texttt{next\_item} & HITS@1 Acc & 10 & Given an item from an ordered category (day, month, letter, number word), output the next item (``Monday'' $\to$ ``Tuesday''). \\
\texttt{prev\_item} & HITS@1 Acc & 10 & As above, but return the previous item (``Tuesday'' $\to$ ``Monday''; wraparound for cyclic lists). \\
\texttt{word\_length} & HITS@1 Acc & 10 & Return the number of characters in the input word (``token'' $\to$ \texttt{5}). \\
\texttt{alphabetically\_first\_3} & HITS@1 Acc & 10 & From a list of 3 strings, choose the alphabetically earliest. \\
\texttt{alphabetically\_first\_5} & HITS@1 Acc & 10 & From a list of 5 strings, choose the alphabetically earliest. \\
\texttt{alphabetically\_last\_3} & HITS@1 Acc & 10 & From a list of 3 strings, choose the alphabetically latest. \\
\texttt{alphabetically\_last\_5} & HITS@1 Acc & 10 & From a list of 5 strings, choose the alphabetically latest. \\
\texttt{choose\_first\_of\_3} & HITS@1 Acc & 10 & From a list of 3 items, select the first item by position. \\
\texttt{choose\_first\_of\_5} & HITS@1 Acc & 10 & From a list of 5 items, select the first item by position. \\
\texttt{choose\_last\_of\_3} & HITS@1 Acc & 10 & From a list of 3 items, select the last item by position. \\
\texttt{choose\_last\_of\_5} & HITS@1 Acc & 10 & From a list of 5 items, select the last item by position. \\
\texttt{choose\_middle\_of\_3} & HITS@1 Acc & 10 & From a list of 3 items, select the middle item by position. \\
\texttt{choose\_middle\_of\_5} & HITS@1 Acc & 10 & From a list of 5 items, select the middle item by position. \\
\bottomrule
\end{tabular}
\end{table}

\paragraph{Aggregation:}
We report a \emph{macro} ICL composite (unweighted mean across selected tasks) and per\mbox{-}task scores. All figures and tables show mean over seeds. Full list of benchmarks used is in table~\ref{tab:benchmarks-detail}

\paragraph{Prompting controls:}
For few\mbox{-}shot tasks, we fix a template and average across multiple demonstration seeds. For robustness, we randomize demonstration order and, in \S\ref{app:icl_robustness}, evaluate sensitivity to number of shots and a label\mbox{-}permutation stress test.

\subsection{Mechanistic telemetry}\label{sec:copy_scores}
\paragraph{Targeted circuits:}
We measure two equality-based copy circuits-\emph{induction} and \emph{anti-induction}-
highlighted in prior work(e.g., \mbox{\citep{olsson2022context, veitsman2025borntransformertransformer}}).\footnote{See also \citep{wang2025inductionheadtoxicitymechanistically,yin2025attentionheadsmatterincontext}.}.
In a left-to-right causal decoder, \emph{attention flows from the later span back to the earlier span}.
Consider a repeated sequence $x = s_0\,s_1\,\dots\,s_{L-1}\ \langle\mathrm{sep}\rangle\ s'_0\,s'_1\,\dots\,s'_{L-1}$ with $s'_i=s_i$:

\begin{itemize}\setlength{\itemsep}{2pt}
  \item \emph{Induction (forward copy).} At position $s'_i$ in the second span, the head locates the earlier repeat and
        retrieves payload that helps predict the \emph{next} token $s'_{i+1}$. We operationalize this with a
        \emph{next-token} alignment (defined below).
  \item \emph{Anti-induction (backward copy).} At position $s'_i$, the head again locates the earlier repeat but retrieves
        payload that helps predict the token immediately to the \emph{left}, $s'_{i-1}$. We operationalize this with a
        \emph{same-token} alignment (defined below).
\end{itemize}

\paragraph{Probe sequences:}
We evaluate on $50{,}000$ fresh copy probes disjoint from training, each built as
$x = s\ \langle\mathrm{sep}\rangle\ s$ with a uniformly sampled token span $s$ of length $L{=}500$\footnote{We evaluate with a span length of $L=500$ (rather than $L=20$) to reduce potential confounds from the \emph{Bi\mbox{-}Induct} pretraining curriculum, which used $L=20$.}.

\paragraph{Per\mbox{-}head scores (how we compute them):}
Let $A^{(\ell,h)}\in\mathbb{R}^{T\times T}$ be the attention map (rows = target positions, columns = source positions)
for layer $\ell$, head $h$ on $x$. Index $t_i$ as the row of $s'_i$ (second span) and $m_i$ as the column of $s_i$ (first span).

\emph{Induction (next-token) score:}
Using $\mathcal{D}_{\text{next}}=\{(t_i, m_{i+1})\}_{i=0}^{L-2}$ (later $s'_i$ to earlier $s_{i+1}$),
\[
\mathrm{Score}_I(\ell,h)\;=\;\mathbb{E}_x\!\Bigg[\frac{1}{L-1}\sum_{(t_i,m_{i+1})\in\mathcal{D}_{\text{next}}} A^{(\ell,h)}_{t_i,m_{i+1}}\Bigg].
\]

\emph{Anti-induction (same-token) score:}
Using the same-token diagonal $\mathcal{D}_{\mathrm{same}}=\{(t_i, m_i)\}_{i=0}^{L-1}$ (later $s'_i$ to earlier $s_i$),
\[
\mathrm{Score}_A(\ell,h)\;=\;\mathbb{E}_x\!\Bigg[\frac{1}{L}\sum_{(t_i,m_i)\in\mathcal{D}_{\mathrm{same}}} A^{(\ell,h)}_{t_i,m_i}\Bigg].
\]
\textbf{Higher is better} for both $\mathrm{Score}_I$ and $\mathrm{Score}_A$ (stronger, more localized copy behavior).

\paragraph{Top 2\% concentration by layer:}
Let $H_\ell$ be the heads in layer $\ell$ and $k_\ell=\max\{1,\lceil 0.02\,|H_\ell|\rceil\}$.
For $Score\in\{Score_I,Score_A\}$, let $\mathrm{Top}_\ell(S)$ be the $k_\ell$ heads with largest $S(\ell,h)$.
We report the mass share
\[
\mathrm{MassShare}_\ell^{(Score)}=\frac{\sum_{h\in \mathrm{Top}_\ell(Score)} Score(\ell,h)}{\sum_{h\in H_\ell} Score(\ell,h)},
\]
and the layer mean $\overline Score_\ell=\frac{1}{|H_\ell|}\sum_{h\in H_\ell}Score(\ell,h)$.
Larger values indicate stronger specialization (copy mass concentrated in a few heads).

\subsection{Language modeling quality}
\paragraph{Perplexity:}
We compute cross\mbox{-}entropy and PPL on a fixed 0.2\% \textsc{The Pile} validation slice (stable hash partition), at iso\mbox{-}FLOPs and identical tokenization settings across runs.

\section{Ablation Study} \label{app:ablation_study}

\subsection{Span length}\label{app:span_len_sweep}
We begin by testing how the snippet span \(L\) affects outcomes. At \(\textbf{0.13B}\), with a fixed initial mix of \(25\%\) linearly annealed over the full token budget, we sweep \(L\!\in\!\{5,20,500\}\) and report two endpoints of practical interest-(i) a 5\mbox{-}shot ICL composite over five standard LM benchmarks and (ii) held\mbox{-}out LM perplexity (PPL). We defer the function-probe suite of \citet{todd2024functionvectorslargelanguage} to the cross-scale experiments, where relative differences are more interpretable and the added compute is justified; the span\mbox{-}length results for this subsection are summarized in Table~\ref{tab:span_sweep_pile_0.13B}.

Across curricula, \(L{=}20\) is a stable operating point that balances ICL and calibration: for \emph{Induction}, \(31.9\) ICL / \(23.9\) PPL (vs.\ \(30.7/23.8\) at \(L{=}5\) and \(31.8/24.0\) at \(L{=}500\)); for \emph{Anti}, \(32.1/24.0\) (vs. \(31.6/23.8\) at \(L{=}5\) , \(31.7/24.5\) at \(L{=}500\)) respectively; for \emph{Balanced}, \(31.2/24.0\) (vs.\(31.4/23.8\) at \(L{=}5\), \(32.0/24.0\) at \(L{=}500\)). Very short spans (\(L{=}5\)) underperform on the ICL composite, while very long spans (\(L{=}500\)) offer no consistent ICL gain and tend to slightly worsen PPL. Hence we adopt \(L{=}20\) for the remaining experiments.
Beyond the ICL\;/\;PPL balance, shorter spans are operationally attractive: they minimize snippet length \(2L{+}|\mathrm{SEP}|\), which reduces potential overhead and makes it easier to \emph{pack} snippets alongside natural sequences to exploit variable\mbox{-}length kernels-yielding compute savings when such packing is enabled.\footnote{All reported results are at iso\mbox{-}FLOPs; we do \emph{not} take packing credits in our comparisons. Packing is a deployment optimization, not part of the evaluation protocol.}

\begin{table*}[t]
\centering
\small
\setlength{\tabcolsep}{5pt}
\renewcommand{\arraystretch}{1.15}
\caption{Span\mbox{-}length sweep at \textbf{0.13B} on \textsc{The Pile}. All curricula are linearly annealed over the full training budget with initial mix of 25\%. Results are averaged over six seeds. Evaluation is 5\mbox{-}shot. We report per\mbox{-}task accuracies, the macro ICL composite, and held\mbox{-}out perplexity (PPL).}
\label{tab:span_sweep_pile_0.13B}
\resizebox{\textwidth}{!}{
\begin{tabular}{lcccccccccc}
\toprule
& \textbf{Baseline} & \multicolumn{3}{c}{\textbf{Induction}} & \multicolumn{3}{c}{\textbf{Anti-induction}} & \multicolumn{3}{c}{\textbf{Balanced}} \\
\cmidrule(lr){2-2}\cmidrule(lr){3-5}\cmidrule(lr){6-8}\cmidrule(lr){9-11}
 &
- & 5 & 20  & 500 & 5 & 20  & 500 & 5 & 20  & 500 \\
\midrule
MMLU $\uparrow$    &$25.2\pm0.3$ &$25.3\pm0.5$  & $25.1\pm0.4$ & $25.3\pm0.5$ &$25.3\pm0.4$  &$25.1\pm0.5$  &$24.8\pm0.5$  &$25.2\pm0.4$  &$24.8\pm0.3$  & $25.1\pm0.2$\\
ARC-Challenge $\uparrow$       &$18.1\pm0.6$  &$18.7\pm0.4$  & $18.6\pm1.3$ &$18.1\pm0.6$ &$18.5\pm0.7$  &$18.1\pm0.4$  & $17.6\pm0.6$ & $18.6\pm0.2$ &$18.5\pm0.6$  & $17.6\pm0.4$  \\
BoolQ $\uparrow$    & $52.8\pm4.6$ &$46.2\pm5.2$  & $53.5\pm5.9$ & $53.4\pm3.7$& $50.8\pm6.8$ &$54.5\pm1.4$  & $53.4\pm2.9$ &$51.2\pm2.5$  & $50.1\pm3.5$ & $54.7\pm4.9$ \\
LAMBADA $\uparrow$    &$7.6\pm0.7$  &$6.7\pm0.6$  &$6.4\pm0.6$  &$6.8\pm0.6$ &$6.4\pm0.5$  & $6.7\pm0.5$ & $6.2\pm0.2$ &$7.0\pm0.1$  &$6.9\pm0.4$  &$6.5\pm0.5$         \\
PIQA $\uparrow$     & $56.6\pm0.5$ &$56.5\pm0.6$  &$55.8\pm0.5$  & $55.5\pm0.2$& $55.9\pm0.3$ & $55.9\pm0.6$ & $56.4\pm0.8$ &$56.0\pm0.6$  &$55.8\pm0.5$  & $56.0\pm0.5$        \\
\rowcolor{gray!15}
 ICL composite (macro) $\uparrow$   & $32.1\pm0.9$ &$30.7\pm2.4$  &$31.9\pm1.1$  & $31.8\pm1.7$ &$31.6\pm1.2$  &$32.1\pm0.3$  & $31.7\pm1.4$ & $31.4\pm3.1$ & $31.2\pm0.7$ &$32.0\pm2.2$  \\

\midrule

PPL $\downarrow$   & 21.8 & 23.8 & 23.9 & 24.0 & 23.8  & 24.0  & 24.5 & 23.8 & 24.0  & 24.2  \\
\bottomrule
\end{tabular}}

\end{table*}

\subsection{Mix Ratio}\label{app:mix_ratio_sweep}
We fixed the anneal to the \emph{full} training budget (2.6B token, following the Chinchilla parameter-token rule of thumb \citep{hoffmann2022trainingcomputeoptimallargelanguage}), held the span length at $L{=}20$, and swept the initial mix ratio over \{25\%, 50\%, 100\% \}. 
Table~\ref{tab:mix_sweep_pile_0.13B} reports full per-task results.

\paragraph{Mechanistic readout:}
Figure~\ref{fig:induction_anti_scores} summarizes layerwise copy-head activity across mix ratios. We quantify head quality in two complementary ways: (i) \emph{spread}, the number of heads per model whose induction score is non-zero (and, for comparability, the count above a fixed “specialization” threshold $>0.5$); and (ii) \emph{peak sharpness}, the maximum head score in each condition. We also note \emph{concentration} in depth (whether peaks cluster in the canonical mid-layers).

\textbf{Does synthetic injection improve induction-head quality vs.\ baseline?}
By counts above the $0.5$ threshold, yes up to moderate mixes. The baseline shows 2 specialized heads. With \textbf{Induction} snippets we observe $\{3,3,1\}$ specialized heads at $\{25\%,50\%,100\%\}$ mixes, \textbf{Balanced} yields $\{4,4,1\}$, and \textbf{Anti} yields $\{3,1,2\}$. By peak sharpness, \textbf{Balanced-25\%} attains the highest induction head, with Induction and Anti close behind; at $\geq50\%$ mixes, Balanced and Baseline retain similar peaks while Induction then Anti trail.

\textbf{Does the curriculum create anti-induction heads?}
No. Even under Anti mixes, anti-induction scores remain far below the specialization threshold; the best peaks are $\approx 0.01$ (at 50\% Anti and in Baseline), indicating no robust anti-induction circuit emerges.

\textbf{How does mix ratio affect spread and depth concentration (induction)?}
For \textbf{Induction}, spread rises from 25\% to 50\% (3$\to$5 heads) then contracts at 100\% (2). For \textbf{Balanced}, spread is largest at 25\% (5), then declines (4 at 50\%, 2 at 100\%). For \textbf{Anti}, induction heads peak at 25-50\% (4 each) and drop to 2 at 100\%. Depthwise, specialized heads shift deeper as mix increases: clusters are earlier at 25\% (layers $\sim$2-4), mid-depth at 50\% (layers $\sim$4-6), and later at 100\% (around layer $\sim$8).

\textbf{Takeaways:}
Baseline naturally forms a few strong induction heads. Adding snippets increases the \emph{number} of specialized induction heads up to moderate mixes ($\leq 50\%$); higher mixes reduce spread and push peaks deeper. None of the curricula-especially Anti-produces meaningful anti-induction heads.

\begin{table*}[t]
\centering
\small
\setlength{\tabcolsep}{5pt}
\renewcommand{\arraystretch}{1.15}
\caption{Initial mix\mbox{-}ratio sweep at \textbf{0.13B} on \textsc{The Pile}. All curricula are linearly annealed over the full training budget with span fixed at $L=20$. Results are averaged over six seeds. Evaluation is 5\mbox{-}shot. We report per\mbox{-}task accuracies, the macro ICL composite, and held\mbox{-}out perplexity (PPL).}
\label{tab:mix_sweep_pile_0.13B}
\resizebox{\textwidth}{!}{
\begin{tabular}{lccccccccccc}
\toprule
& \textbf{Baseline} & \multicolumn{3}{c}{\textbf{Induction}} & \multicolumn{3}{c}{\textbf{Anti-induction}} & \multicolumn{3}{c}{\textbf{Balanced}} \\
\cmidrule(lr){2-2}\cmidrule(lr){3-5}\cmidrule(lr){6-8}\cmidrule(lr){9-11}
 &
- & 25\% & 50\% & 100\% & 25\% & 50\% & 100\% & 25\% & 50\% & 100\% \\
\midrule
MMLU $\uparrow$       & $25.2\pm 0.26$ & $25.1\pm 0.38$ & $25.0\pm 0.33$ & $24.9\pm 0.33$ & $25.1\pm 0.48$ & $25.2\pm 0.40$ & $24.9\pm 0.41$ & $24.8\pm 0.29$ & $25.0\pm 0.27$ & $24.6\pm 0.38$ \\
ARC-Challenge $\uparrow$   & $18.1\pm 0.64$ & $18.6\pm 1.28$ & $18.1\pm 0.65$ & $17.9\pm 0.89$ & $18.1\pm 0.42$ & $18.0\pm 0.47$ & $17.5\pm 0.91$ & $18.5\pm 0.58$ & $18.1\pm 0.44$ & $17.8\pm 0.52$ \\
BoolQ $\uparrow$  & $52.8\pm 4.64$ & $53.5\pm 5.92$ & $49.6\pm 6.92$ & $51.8\pm 4.51$ & $54.5\pm 1.36$ & $47.8\pm 6.78$ & $51.1\pm 2.87$ & $50.1\pm 3.52$ & $46.9\pm 4.68$ & $51.8\pm 5.37$ \\
LAMBADA $\uparrow$        &  $7.6\pm 0.69$ &  $6.4\pm 0.60$ &  $5.6\pm 0.50$ &  $4.7\pm 0.50$ &  $6.7\pm 0.45$ &  $5.6\pm 0.51$ &  $4.4\pm 0.45$ &  $6.9\pm 0.43$ &  $5.9\pm 0.66$ &  $4.6\pm 0.38$ \\
PIQA $\uparrow$          & $56.6\pm 0.49$ & $55.8\pm 0.48$ & $55.4\pm 0.27$ & $54.9\pm 0.49$ & $55.9\pm 0.55$ & $55.4\pm 0.37$ & $55.2\pm 0.55$ & $55.8 \pm 0.50$ & $55.8\pm 0.72$ & $54.4\pm 0.45$ \\
\rowcolor{gray!15}
ICL composite (macro) $\uparrow$ & $32.06\pm 0.89$ & $31.88\pm1.08$ & $30.76\pm1.48$ & $30.83\pm0.98$ & $32.06\pm0.25$ & $30.40\pm1.40$ & $30.62\pm0.66$ & $31.22\pm0.69$ & $30.34\pm0.92$ & $30.64\pm1.10$ \\

\midrule

PPL $\downarrow$           & 21.8 & 23.9 & 26.0 & 31.4 & 24.0 & 26.4 & 32.9 & 24.0 & 26.2 & 32.8 \\
\bottomrule
\end{tabular}}
\end{table*}

\begin{figure*}[t]
  \centering
  \begin{subfigure}[t]{0.32\linewidth}
    \centering
    \includegraphics[width=\linewidth]{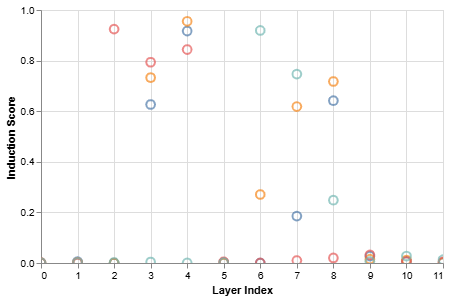}
    \caption{}
  \end{subfigure}\hfill
  \begin{subfigure}[t]{0.32\linewidth}
    \centering
    \includegraphics[width=\linewidth]{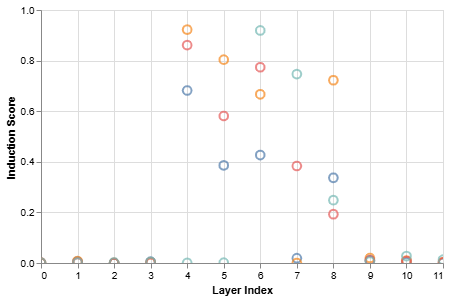}
    \caption{}
  \end{subfigure}\hfill
  \begin{subfigure}[t]{0.32\linewidth}
    \centering
    \includegraphics[width=\linewidth]{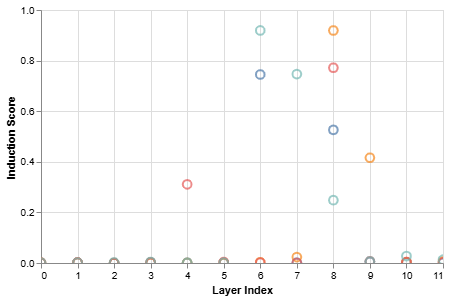}
    \caption{}
  \end{subfigure}

  \vspace{0.5em}

  \begin{subfigure}[t]{0.32\linewidth}
    \centering
    \includegraphics[width=\linewidth]{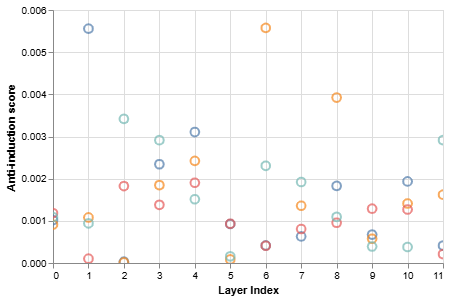}
    \caption{}
  \end{subfigure}\hfill
  \begin{subfigure}[t]{0.32\linewidth}
    \centering
    \includegraphics[width=\linewidth]{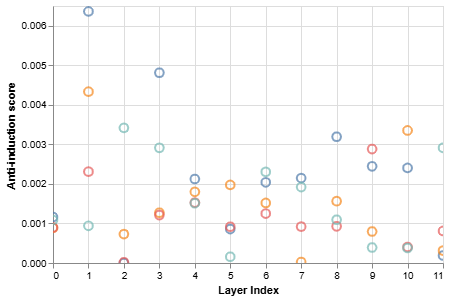}
    \caption{}
  \end{subfigure}\hfill
  \begin{subfigure}[t]{0.32\linewidth}
    \centering
    \includegraphics[width=\linewidth]{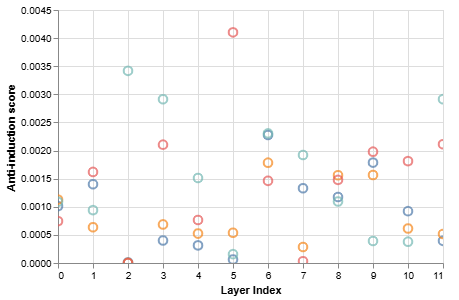} 
    \caption{}
  \end{subfigure}

  \par\vspace{0.25em}
  \begin{minipage}{0.98\linewidth}
    \centering
    \begin{tabular}{p{0.32\linewidth} p{0.32\linewidth} p{0.32\linewidth}}
      \centering 25\% mix ratio & \centering 50\% mix ratio& \centering 100\% mix ratio \\
    \end{tabular}
  \end{minipage}

   \newcommand{\legendcircsize}{3pt} 
\vspace{0.6em}
\begin{minipage}{0.98\linewidth}
  \centering
  \setlength{\tabcolsep}{10pt}
  \begin{tabular}{cccc}
    \tikz\draw[line width=0.9pt, legBaseline]  (0,0) circle (\legendcircsize);~ Baseline &
    \tikz\draw[line width=0.9pt, legInduction] (0,0) circle (\legendcircsize);~ Induction &
    \tikz\draw[line width=0.9pt, legAnti]      (0,0) circle (\legendcircsize);~ Anti &
    \tikz\draw[line width=0.9pt, legBalanced]  (0,0) circle (\legendcircsize);~ Balanced 
    
  \end{tabular}
\end{minipage}
  
  \caption{Layer-wise copy-head telemetry. Top row: induction scores; bottom row: anti-induction scores. For each layer we plot the best-scoring head (top 2\% by score with a floor of one head per layer), averaged over six seeds, for the 0.13B model with initial mix ratios: 25\%, 50\%, and 100\% . Head counts for each model are given in Table~\ref{tab:model-presets}.}
  \label{fig:induction_anti_scores}
\end{figure*}

\section{In-Context Learning Capability}\label{app:icl_capabilities}

\subsection{In-Context Learning Performance}\label{app:icl_performance}
In Table~\ref{tab:main_icl_performance} (summarized in Figure~\ref{fig:icl-lm-eval} and Figure~\ref{fig:icl-todd-suit}; see \S\ref{sec:icl_performance}), we report \emph{per-task} few-shot ICL performance across scales for two families: (i) 14 standard LM benchmarks/tasks and (ii) 19 function-style probes from the \citet{todd2024functionvectorslargelanguage} suite. For the standard LM benchmarks we use \textbf{3-shot} evaluation; for the \cite{todd2024functionvectorslargelanguage} suite we use \textbf{10-shot} evaluation. (Table~\ref{tab:benchmarks-detail} lists all tasks and provides a brief description of each.) Unless otherwise noted, metrics are accuracy (ACC) or exact match (EM) as standard, and all results are averaged over three seeds. 

Because any $>0$ shot setting exercises in-context learning, we also study \textbf{1-shot} sensitivity for the same tasks/benchmarks in Appendix~\ref{app:icl_robustness}.

\begin{table*}[t]
\centering
\small
\setlength{\tabcolsep}{5pt}
\renewcommand{\arraystretch}{1.15}
\caption{Results across model scales (0.13B, 0.5B, 1B) on \textsc{The Pile} at iso\mbox{-}FLOPs. Copy snippets use span $L{=}20$. Evaluation is few\mbox{-}shot: 3\mbox{-}shot for standard LM benchmarks, and 10\mbox{-}shot for function\mbox{-}style probes. We report per\mbox{-}task accuracy (or EM where standard), averaged over three seeds, and the ICL composite (macro\mbox{-}average across tasks). Higher is better.}

\label{tab:main_icl_performance}
\resizebox{\textwidth}{!}{
\begin{tabular}{lcccccccccccc}
\toprule
& \multicolumn{3}{c}{\textbf{Baseline}} & \multicolumn{3}{c}{\textbf{Induction}} & \multicolumn{3}{c}{\textbf{Anti-induction}} & \multicolumn{3}{c}{\textbf{Balanced}} \\
\cmidrule(lr){2-4}\cmidrule(lr){5-7}\cmidrule(lr){8-10}\cmidrule(lr){11-13}
 & 0.13B & 0.5B & 1B & 0.13B & 0.5B & 1B & 0.13B & 0.5B & 1B & 0.13B & 0.5B & 1B \\
\midrule
MMLU        & $26.7\pm0.1$ &$27.5\pm0.4$  &$27.7\pm0.0$  &$25.9\pm0.0$  &$27.4\pm0.0$  &$26.6\pm0.0$  & $27.1\pm0.0$ &$26.8\pm0.0$  &$27.2\pm0.1$  & $26.2\pm0.1$ &$27.1\pm0.2$  & $27.1\pm0.0$ 
\\
Winogrande        & $50.4\pm0.6$ &$50.8\pm1.4$ &$49.8\pm1.1$  &$47.0\pm0.9$  &$50.5\pm1.1$  &$51.1\pm0.8$  &$51.2\pm0.9$  &$50.1\pm1.0$  &$49.8\pm1.3$  & $50.9\pm1.6$ &$50.9\pm0.4$  &$51.0\pm1.2$  
\\
CommonSenseQA       & $20.8\pm1.2$ &$20.0\pm1.1$  &$21.2\pm0.9$  &$20.3\pm0.3$  &$20.5\pm0.9$ &$20.5\pm0.8$  &$20.8\pm0.3$  &$20.0\pm1.2$  &$20.0\pm0.4$  & $20.6\pm0.7$ &$20.5\pm0.3$  & $20.4\pm0.4$ 
\\
PIQA        & $56.6\pm0.3$ &$58.5\pm1.2$  &$59.2\pm0.5$  &$55.0\pm0.2$  &$58.6\pm0.1$  &$58.4\pm0.3$  &$56.1\pm0.2$  &$56.9\pm0.3$  &$58.9\pm0.6$  & $55.5\pm0.7$ &$57.2\pm0.6$  &$58.4\pm0.3$  
\\
HellaSwag        & $26.5\pm0.1$ &$27.1\pm0.4$  &$27.8\pm0.1$  &$26.3\pm0.1$  &$26.5\pm0.1$ &$27.2\pm0.1$  & $26.4\pm0.1$ &$26.8\pm0.1$  &$27.3\pm0.1$  & $26.2\pm0.1$ &$26.5\pm0.1$  &$27.3\pm0.1$  
\\
TriviaQA\mbox{-}Wiki       &$0.1\pm0.0$  &$0.2\pm0.0$  &$0.4\pm0.0$  &$0.1\pm0.0$  &$0.1\pm0.0$  &$0.3\pm0.0$  & $0.1\pm0.0$ &$0.1\pm0.0$  &$0.3\pm0.0$  &$0.1\pm0.0$  &$0.1\pm0.0$  &$0.3\pm0.0$  
\\
BBH (CoT)       &$0.1\pm0.0$  &$1.5\pm0.2$  &$2.8\pm0.0$  &$0.3\pm0.0$  &$1.6\pm0.1$  &$1.2\pm0.1$  &$0.8\pm0.1$  &$0.1\pm0.0$  &$0.6\pm0.0$  & $0.1\pm0.0$ & $2.2\pm0.0$ & $4.2\pm0.0$ 
\\
OpenBookQA        & $14.3\pm0.6$ &$14.0\pm1.6$  &$15.9\pm1.1$  &$13.9\pm0.8$  &$15.3\pm0.1$  &$15.3\pm0.4$  &$14.7\pm0.7$  &$13.9\pm1.3$  & $15.6\pm0.4$ & $13.1\pm0.1$ &$13.6\pm1.0$  & $16.5\pm0.5$ 
\\
ARC\mbox{-}Challenge       &$18.6\pm0.6$  &$18.4\pm1.1$  &$18.2\pm0.6$  &$18.3\pm0.3$ &$18.5\pm0.7$  &$18.1\pm0.4$  & $17.6\pm0.4$ & $17.8\pm0.4$ &$19.0\pm0.4$  & $17.1\pm0.8$ &$17.9\pm0.4$  & $17.9\pm0.3$ 
\\
GPQA       &$25.2\pm2.2$  &$26.1\pm2.1$  &$25.2\pm2.0$  &$23.7\pm2.3$  &$25.2\pm1.7$  &$24.3\pm1.5$  &$23.9\pm2.0$  &$24.9\pm1.4$  & $24.1\pm1.1$ & $24.1\pm2.2$ &$22.8\pm0.8$  &  $24.6\pm0.8$
\\
GSM\mbox{-}8K       &$1.5\pm0.4$  &$1.5\pm0.3$  &$1.7\pm0.2$  &$1.1\pm0.3$  &$1.5\pm0.3$  &$1.5\pm0.2$  &$1.4\pm0.2$  &$1.2\pm0.1$  &$1.6\pm0.1$  &$1.1\pm0.4$  &$1.7\pm0.2$  & $1.4\pm0.2$ 
\\
MathQA       & $20.5\pm0.4$ &$20.9\pm0.7$  &$20.5\pm0.7$  &$19.9\pm0.6$  &$20.3\pm0.2$  &$20.7\pm0.6$  &$20.2\pm0.3$  &$20.4\pm0.4$  & $21.1\pm0.2$ &$21.0\pm0.7$  &$20.8\pm0.4$  & $21.0\pm0.1$ 
\\
BoolQ       & $48.8\pm0.5$ &$53.4\pm0.9$  &$54.7\pm0.2$  &$49.1\pm0.7$  &$60.5\pm0.3$  &$57.0\pm0.9$  & $49.1\pm0.8$ &$52.2\pm2.2$  & $52.7\pm0.3$ &$51.4\pm0.5$  &$57.1\pm1.0$  &$58.1\pm0.4$  
\\
LAMBADA (OpenAI)       & $8.2\pm0.2$ &$11.0\pm0.4$  &$13.2\pm0.1$ &$5.5\pm0.1$  &$8.6\pm0.1$  &$12.2\pm0.2$  &$5.2\pm0.2$  & $8.4\pm0.3$ & $12.2\pm0.2$ & $6.0\pm0.2$ &$8.2\pm0.2$  & $12.0\pm0.3$ 
\\
\rowcolor{gray!15}
ICL composite (macro) $\uparrow$ & $22.7\pm0.5$ & $23.6\pm0.8$ & $24.2\pm0.5$ & $21.9\pm0.5$ & $23.9\pm0.4$ & $23.9\pm0.5$ & $22.5\pm0.4$ & $22.8\pm0.6$ & $23.6\pm0.4$ & $22.4\pm0.6$ & $23.3\pm0.4$ & $24.3\pm0.3$ \\

\midrule
alphabetically\_first\_3 & $4.9\,\pm\,0.6$ & $9.4\,\pm\,0.6$ & $19.7\,\pm\,0.9$ & $4.6\,\pm\,0.6$ & $7.3\,\pm\,0.3$ & $15.1\,\pm\,0.9$ & $3.3\,\pm\,0.5$ & $8.5\,\pm\,0.9$ & $14.4\,\pm\,1.5$ & $4.8\,\pm\,0.6$ & $11.5\,\pm\,0.6$ & $15.3\,\pm\,1.0$ \\
alphabetically\_first\_5 & $4.0\,\pm\,0.6$ & $6.7\,\pm\,0.8$ & $12.4\,\pm\,1.4$ & $4.2\,\pm\,0.8$ & $6.5\,\pm\,1.0$ & $10.9\,\pm\,0.9$ & $2.8\,\pm\,0.7$ & $7.4\,\pm\,0.6$ & $8.4\,\pm\,1.1$ & $3.9\,\pm\,1.0$ & $8.2\,\pm\,0.8$ & $10.2\,\pm\,0.7$ \\
alphabetically\_last\_3 & $3.4\,\pm\,0.5$ & $10.2\,\pm\,0.9$ & $20.8\,\pm\,0.6$ & $2.4\,\pm\,0.5$ & $7.7\,\pm\,0.2$ & $15.8\,\pm\,1.0$ & $1.9\,\pm\,0.6$ & $8.0\,\pm\,0.9$ & $13.3\,\pm\,0.8$ & $3.9\,\pm\,0.5$ & $10.9\,\pm\,0.5$ & $15.3\,\pm\,1.3$ \\
alphabetically\_last\_5 & $2.5\,\pm\,0.7$ & $6.0\,\pm\,1.3$ & $9.9\,\pm\,0.8$ & $1.6\,\pm\,0.3$ & $5.5\,\pm\,0.4$ & $8.5\,\pm\,0.3$ & $1.8\,\pm\,0.4$ & $6.1\,\pm\,0.8$ & $7.8\,\pm\,0.5$ & $2.5\,\pm\,0.8$ & $6.7\,\pm\,0.6$ & $9.3\,\pm\,0.4$ \\
capitalize & $8.0\,\pm\,1.0$ & $20.7\,\pm\,1.4$ & $54.8\,\pm\,2.1$ & $3.3\,\pm\,0.3$ & $13.2\,\pm\,1.3$ & $33.4\,\pm\,1.3$ & $3.7\,\pm\,0.1$ & $13.5\,\pm\,1.3$ & $39.2\,\pm\,2.7$ & $6.0\,\pm\,1.6$ & $14.7\,\pm\,1.3$ & $33.6\,\pm\,2.1$ \\
capitalize\_first\_letter & $10.1\,\pm\,1.2$ & $12.5\,\pm\,1.8$ & $28.6\,\pm\,1.1$ & $5.3\,\pm\,1.0$ & $13.4\,\pm\,1.7$ & $13.7\,\pm\,1.2$ & $5.2\,\pm\,0.3$ & $11.2\,\pm\,1.2$ & $17.4\,\pm\,0.8$ & $8.9\,\pm\,0.7$ & $10.0\,\pm\,1.1$ & $20.4\,\pm\,1.2$ \\
capitalize\_last\_letter & $4.8\,\pm\,1.3$ & $9.6\,\pm\,1.0$ & $8.3\,\pm\,1.2$ & $8.6\,\pm\,1.1$ & $7.5\,\pm\,1.8$ & $8.6\,\pm\,0.9$ & $9.1\,\pm\,1.4$ & $9.3\,\pm\,1.2$ & $7.3\,\pm\,1.5$ & $5.5\,\pm\,0.7$ & $5.8\,\pm\,1.4$ & $6.7\,\pm\,0.7$ \\
choose\_first\_of\_3 & $10.3\,\pm\,2.3$ & $25.5\,\pm\,1.8$ & $69.4\,\pm\,1.8$ & $6.1\,\pm\,0.9$ & $19.0\,\pm\,1.7$ & $52.4\,\pm\,2.0$ & $4.1\,\pm\,0.7$ & $19.1\,\pm\,1.8$ & $46.0\,\pm\,2.0$ & $11.3\,\pm\,0.7$ & $35.2\,\pm\,1.7$ & $54.1\,\pm\,1.7$ \\
choose\_first\_of\_5 & $8.7\,\pm\,1.3$ & $19.7\,\pm\,1.9$ & $55.5\,\pm\,1.3$ & $4.9\,\pm\,0.7$ & $15.3\,\pm\,0.9$ & $42.6\,\pm\,3.0$ & $3.6\,\pm\,0.7$ & $16.2\,\pm\,1.5$ & $32.7\,\pm\,1.6$ & $7.9\,\pm\,1.0$ & $28.2\,\pm\,2.4$ & $42.1\,\pm\,1.4$ \\
choose\_last\_of\_3 & $2.0\,\pm\,0.4$ & $3.2\,\pm\,0.2$ & $4.4\,\pm\,1.0$ & $1.4\,\pm\,0.4$ & $2.8\,\pm\,0.3$ & $5.4\,\pm\,0.9$ & $1.4\,\pm\,0.5$ & $3.0\,\pm\,0.5$ & $5.0\,\pm\,0.5$ & $1.7\,\pm\,0.3$ & $3.0\,\pm\,0.5$ & $5.2\,\pm\,0.3$ \\
choose\_last\_of\_5 & $1.5\,\pm\,0.3$ & $2.9\,\pm\,0.6$ & $3.9\,\pm\,1.1$ & $1.7\,\pm\,0.5$ & $2.4\,\pm\,0.4$ & $4.6\,\pm\,0.6$ & $1.1\,\pm\,0.4$ & $2.9\,\pm\,0.5$ & $5.6\,\pm\,0.5$ & $1.5\,\pm\,0.3$ & $2.6\,\pm\,0.4$ & $5.1\,\pm\,0.4$ \\
choose\_middle\_of\_3 & $1.7\,\pm\,0.6$ & $3.3\,\pm\,0.5$ & $4.2\,\pm\,0.5$ & $2.1\,\pm\,0.8$ & $2.2\,\pm\,0.3$ & $6.0\,\pm\,1.0$ & $1.3\,\pm\,0.5$ & $3.1\,\pm\,0.2$ & $5.0\,\pm\,0.3$ & $1.7\,\pm\,0.7$ & $3.4\,\pm\,0.7$ & $4.9\,\pm\,0.7$ \\
choose\_middle\_of\_5 & $1.6\,\pm\,0.3$ & $3.0\,\pm\,0.6$ & $2.7\,\pm\,0.4$ & $1.6\,\pm\,0.2$ & $2.1\,\pm\,0.3$ & $3.1\,\pm\,0.6$ & $1.7\,\pm\,0.4$ & $2.9\,\pm\,0.7$ & $3.9\,\pm\,0.7$ & $1.7\,\pm\,0.4$ & $2.3\,\pm\,0.6$ & $4.7\,\pm\,0.6$ \\
lowercase\_first\_letter & $5.5\,\pm\,0.8$ & $8.4\,\pm\,0.9$ & $28.5\,\pm\,2.2$ & $4.1\,\pm\,0.9$ & $6.1\,\pm\,1.2$ & $20.1\,\pm\,1.6$ & $4.7\,\pm\,0.7$ & $6.5\,\pm\,0.6$ & $20.6\,\pm\,1.1$ & $2.2\,\pm\,0.6$ & $8.8\,\pm\,0.7$ & $13.3\,\pm\,0.8$ \\
lowercase\_last\_letter & $11.1\,\pm\,0.8$ & $7.7\,\pm\,0.7$ & $10.5\,\pm\,1.1$ & $3.6\,\pm\,0.7$ & $8.0\,\pm\,0.7$ & $9.6\,\pm\,1.0$ & $7.4\,\pm\,1.2$ & $8.9\,\pm\,1.0$ & $13.3\,\pm\,0.5$ & $7.8\,\pm\,1.1$ & $10.9\,\pm\,1.5$ & $8.9\,\pm\,1.2$ \\
next\_capital\_letter & $4.9\,\pm\,1.1$ & $4.2\,\pm\,0.9$ & $2.4\,\pm\,0.8$ & $3.9\,\pm\,0.3$ & $4.2\,\pm\,1.0$ & $3.6\,\pm\,0.7$ & $4.7\,\pm\,0.8$ & $4.1\,\pm\,0.7$ & $3.1\,\pm\,0.9$ & $4.8\,\pm\,0.9$ & $3.5\,\pm\,0.9$ & $4.2\,\pm\,1.1$ \\
next\_item & $3.4\,\pm\,2.2$ & $8.6\,\pm\,2.5$ & $16.8\,\pm\,2.1$ & $2.9\,\pm\,1.9$ & $5.7\,\pm\,1.3$ & $11.8\,\pm\,1.2$ & $1.3\,\pm\,0.9$ & $7.8\,\pm\,1.1$ & $11.8\,\pm\,2.5$ & $6.3\,\pm\,0.8$ & $9.0\,\pm\,1.6$ & $9.4\,\pm\,1.7$ \\
prev\_item & $2.5\,\pm\,0.8$ & $7.7\,\pm\,1.6$ & $15.1\,\pm\,2.3$ & $2.0\,\pm\,0.8$ & $5.9\,\pm\,0.7$ & $10.3\,\pm\,1.0$ & $1.4\,\pm\,0.5$ & $6.6\,\pm\,2.5$ & $11.4\,\pm\,2.5$ & $5.7\,\pm\,1.3$ & $8.4\,\pm\,1.2$ & $7.7\,\pm\,2.2$ \\
word\_length & $9.9\,\pm\,1.0$ & $13.3\,\pm\,0.9$ & $12.8\,\pm\,1.1$ & $13.4\,\pm\,0.8$ & $14.0\,\pm\,1.0$ & $13.3\,\pm\,0.8$ & $12.0\,\pm\,1.3$ & $14.3\,\pm\,0.6$ & $13.6\,\pm\,1.6$ & $10.8\,\pm\,1.1$ & $14.7\,\pm\,1.8$ & $12.6\,\pm\,0.7$ \\
\rowcolor{gray!15}
ICL composite (macro) $\uparrow$ & $5.3\,\pm\,0.9$ & $9.6\,\pm\,1.1$ & $20.0\,\pm\,1.3$ & $4.1\,\pm\,0.7$ & $7.8\,\pm\,0.9$ & $15.2\,\pm\,1.1$ & $3.8\,\pm\,0.7$ & $8.4\,\pm\,1.0$ & $14.7\,\pm\,1.2$ & $5.2\,\pm\,0.8$ & $10.4\,\pm\,1.1$ & $14.9\,\pm\,1.1$ \\

\bottomrule

\end{tabular}}
\end{table*}

\subsection{In-Context Learning Robustness}\label{app:icl_robustness}

\subsubsection{Sensitivity to Number of Shots}
\begin{figure*}[t]
  \centering

  \begin{subfigure}[t]{\textwidth}
    \centering
    \includegraphics[width=\textwidth]{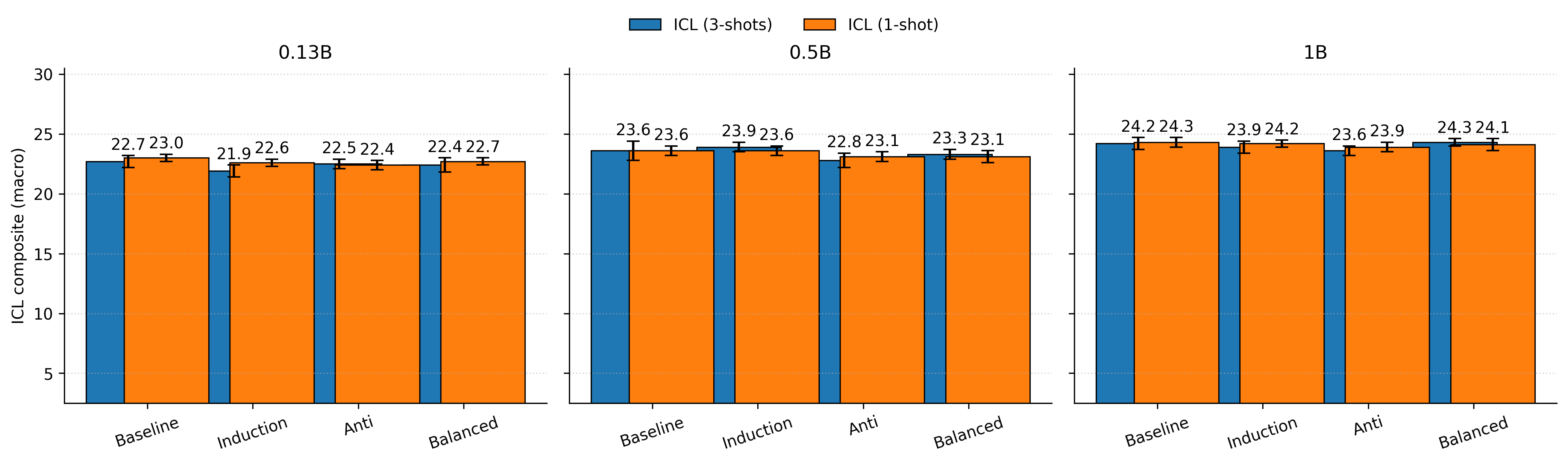}
    \caption{}
    \label{fig:lm-icl-shots}
  \end{subfigure}

  \vspace{0.6em}

  \begin{subfigure}[t]{\textwidth}
    \centering
    \includegraphics[width=\textwidth]{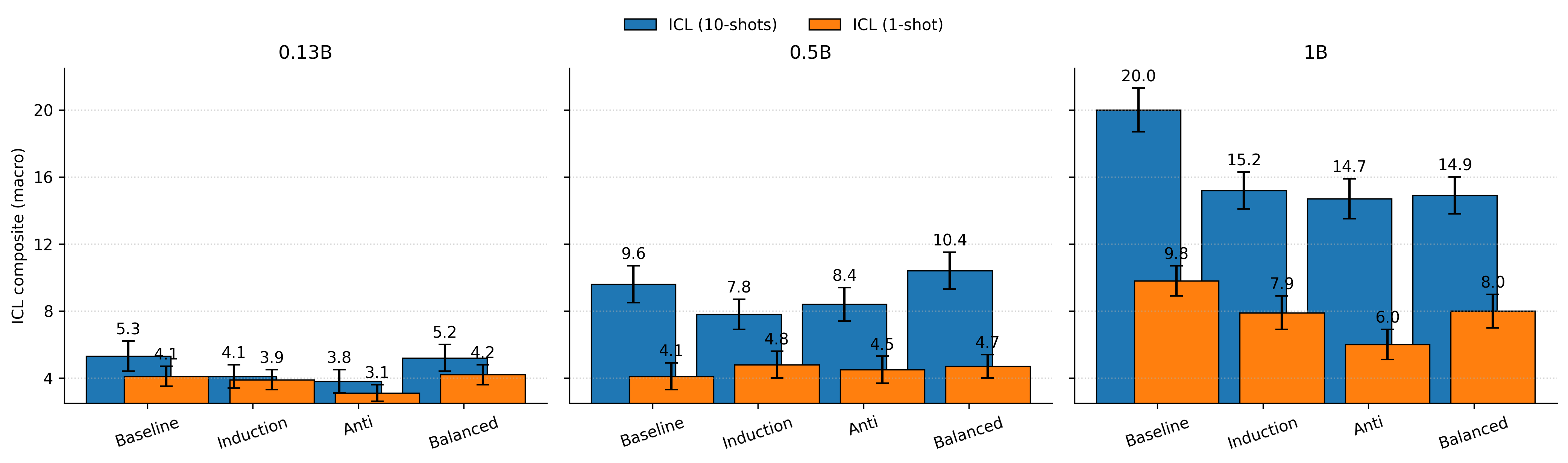}
    \caption{}
    \label{fig:todd-icl-shots}
  \end{subfigure}

  \caption{Sensitivity of ICL composite (macro) to the number of shots across two evaluation families: (a) standard LM benchmarks (3-shot vs. 1-shot); (b) Function-probe suite of \cite{todd2024functionvectorslargelanguage} (10-shot vs. 1-shot). Each panel groups models by size (0.13B, 0.5B, 1B), colors the bars by regime (Baseline, Induction, Anti, Balanced), and shows $\pm 1$ s.d. error bars.}
  \label{fig:icl-1shot-combined}
\end{figure*}

We assess how the ICL results in §\ref{sec:icl_performance} (with details in Appendix~\ref{app:icl_performance}) vary with the number of in\mbox{-}context demonstrations. Concretely, we change the evaluation from the main\mbox{-}text setting (3\mbox{-}shot for standard LM benchmarks and 10\mbox{-}shot for function\mbox{-}probe tasks) to a unified 1\mbox{-}shot setting for both families. Summaries appear in Figure~\ref{fig:lm-icl-shots} (standard LM) and Figure~\ref{fig:todd-icl-shots} (function probes); per\mbox{-}task scores are in Table~\ref{tab:icl_probes_shots_sensitivty}.

For the standard LM benchmarks, moving to 1\mbox{-}shot produces negligible changes across all model sizes (0.13B, 0.5B, 1B). In contrast, the function\mbox{-}style probes degrade notably at 0.5B and 1B when reduced to 1\mbox{-}shot, while the 0.13B model shows only a small drop. This scale\mbox{-}dependent sensitivity aligns with prior observations that larger models more reliably use the demonstration label$\to$token mapping (and thus benefit from more shots), whereas smaller models often gain primarily from format/structure and topical priming~\citep{wei2023largerlanguagemodelsincontext,min2022rethinkingroledemonstrationsmakes,zhao2021calibrateuseimprovingfewshot}. Consistently, our label\mbox{-}permutation stress test shows minimal impact at 0.13B but clear degradation at 0.5B/1B, indicating that bigger models lean more on the (now corrupted) mapping signal.

Across shot conditions, the \textsc{Baseline} vs.\ \textsc{Bi\mbox{-}Induct} ordering remains stable: reducing shots changes the absolute level but not the ranking among curricula.

\begin{table*}[t]
\centering
\small
\setlength{\tabcolsep}{3.5pt}
\renewcommand{\arraystretch}{1.15}
\caption{Results across model scales (0.13B, 0.5B, 1B) on \textsc{The Pile} at iso\mbox{-}FLOPs. Copy snippets use span $L{=}20$. Evaluation is few\mbox{-}shot: 1\mbox{-}shot for both standard LM benchmarks, and function\mbox{-}style probes. We report per\mbox{-}task accuracy (or EM where standard), averaged over three seeds, and the ICL composite (macro\mbox{-}average across tasks). Higher is better.}
\label{tab:icl_probes_shots_sensitivty}
\resizebox{\textwidth}{!}{
\begin{tabular}{l ccc ccc ccc ccc}
\toprule
& \multicolumn{3}{c}{\textbf{Baseline}} & \multicolumn{3}{c}{\textbf{Induction}} & \multicolumn{3}{c}{\textbf{Anti-induction}} & \multicolumn{3}{c}{\textbf{Balanced}} \\
\cmidrule(lr){2-4}\cmidrule(lr){5-7}\cmidrule(lr){8-10}\cmidrule(lr){11-13}
 & 0.13B & 0.5B & 1B & 0.13B & 0.5B & 1B & 0.13B & 0.5B & 1B & 0.13B & 0.5B & 1B \\
 \midrule
MMLU             & 25.8 $\pm$ 0.0 & 25.2 $\pm$ 0.3 & 26.8 $\pm$ 0.0 & 25.8 $\pm$ 0.0 & 24.4 $\pm$ 0.3 & 26.4 $\pm$ 0.0 & 25.8 $\pm$ 0.0 & 24.7 $\pm$ 0.3 & 26.8 $\pm$ 0.0 & 25.9 $\pm$ 0.0 & 24.3 $\pm$ 0.2 & 27.3 $\pm$ 0.0 \\
Winogrande                 & 49.8 $\pm$ 0.8 & 50.3 $\pm$ 0.5 & 49.6 $\pm$ 0.4 & 48.9 $\pm$ 0.3 & 50.2 $\pm$ 0.3 & 50.6 $\pm$ 0.4 & 50.9 $\pm$ 1.2 & 50.4 $\pm$ 0.8 & 49.3 $\pm$ 0.9 & 50.8 $\pm$ 0.3 & 52.0 $\pm$ 0.4 & 51.0 $\pm$ 0.0 \\
CommonSenseQA              & 20.9 $\pm$ 0.9 & 20.6 $\pm$ 1.0 & 20.5 $\pm$ 0.5 & 20.8 $\pm$ 0.8 & 20.6 $\pm$ 0.9 & 20.6 $\pm$ 0.6 & 20.9 $\pm$ 0.8 & 20.6 $\pm$ 1.0 & 20.7 $\pm$ 0.1 & 21.0 $\pm$ 0.8 & 20.7 $\pm$ 0.9 & 20.6 $\pm$ 0.6 \\
PIQA                 & 56.8 $\pm$ 0.4 & 58.7 $\pm$ 0.8 & 59.9 $\pm$ 0.4 & 55.4 $\pm$ 0.2 & 58.8 $\pm$ 0.2 & 58.3 $\pm$ 0.3 & 55.9 $\pm$ 0.1 & 57.4 $\pm$ 0.5 & 59.0 $\pm$ 1.1 & 55.4 $\pm$ 0.9 & 57.3 $\pm$ 0.3 & 58.1 $\pm$ 0.2 \\
HellaSwag           & 26.4 $\pm$ 0.1 & 27.0 $\pm$ 0.1 & 27.8 $\pm$ 0.1 & 26.2 $\pm$ 0.1 & 26.7 $\pm$ 0.0 & 27.3 $\pm$ 0.1 & 26.3 $\pm$ 0.1 & 26.7 $\pm$ 0.2 & 27.3 $\pm$ 0.1 & 26.2 $\pm$ 0.1 & 26.6 $\pm$ 0.1 & 27.2 $\pm$ 0.2 \\
TriviaQA-Wiki             &  0.1 $\pm$ 0.0 & 0.1 $\pm$ 0.0 &  0.3 $\pm$ 0.0 &  0.1 $\pm$ 0.0 & 0.1 $\pm$ 0.0 &  0.1 $\pm$ 0.0 &  0.1 $\pm$ 0.0 & 0.1 $\pm$ 0.0 &  0.1 $\pm$ 0.0 &  0.1 $\pm$ 0.0 & 0.1 $\pm$ 0.0 &  0.2 $\pm$ 0.0 \\
BBH (CoT)                   &  0.0 $\pm$ 0.0 &  0.6 $\pm$ 0.0 &  3.4 $\pm$ 0.0 &  0.1 $\pm$ 0.02 &  2.0 $\pm$ 0.0 &  4.3 $\pm$ 0.0 &  0.4 $\pm$ 0.00 &  0.1 $\pm$ 0.0 &  0.6 $\pm$ 0.1 &  0.4 $\pm$ 0.00 &  1.8 $\pm$ 1.1 &  2.8 $\pm$ 0.0 \\
OpenBookQA          & 14.3 $\pm$ 0.5 & 15.6 $\pm$ 0.7 & 15.8 $\pm$ 0.5 & 14.3 $\pm$ 0.2 & 14.5 $\pm$ 0.8 & 15.5 $\pm$ 0.5 & 13.0 $\pm$ 0.0 & 14.7 $\pm$ 0.2 & 16.7 $\pm$ 0.7 & 13.3 $\pm$ 0.1 & 13.9 $\pm$ 0.1 & 15.6 $\pm$ 1.2 \\
ARC-Challenge        & 18.4 $\pm$ 0.2 & 19.0 $\pm$ 0.4 & 18.1 $\pm$ 0.1 & 18.7 $\pm$ 0.4 & 18.2 $\pm$ 0.1 & 18.3 $\pm$ 0.2 & 17.7 $\pm$ 0.3 & 17.8 $\pm$ 0.2 & 18.4 $\pm$ 0.5 & 17.5 $\pm$ 0.4 & 17.8 $\pm$ 0.4 & 17.7 $\pm$ 0.7 \\
GPQA         & 24.8 $\pm$ 0.6 & 25.0 $\pm$ 1.0 & 26.2 $\pm$ 2.5 & 25.0 $\pm$ 0.8 & 24.4 $\pm$ 1.3 & 25.2 $\pm$ 0.6 & 25.1 $\pm$ 1.1 & 25.9 $\pm$ 1.2 & 25.7 $\pm$ 1.2 & 25.5 $\pm$ 0.3 & 24.2 $\pm$ 2.1 & 26.0 $\pm$ 2.0 \\ 
GSM-8K                &  1.3 $\pm$ 0.1 &  1.9 $\pm$ 0.3 &  1.3 $\pm$ 0.1 &  1.5 $\pm$ 0.6 &  2.2 $\pm$ 0.3 &  1.5 $\pm$ 0.2 &  1.3 $\pm$ 0.5 &  1.8 $\pm$ 0.2 &  1.6 $\pm$ 0.3 &  1.1 $\pm$ 0.1 &  2.0 $\pm$ 0.3 &  1.5 $\pm$ 0.2 \\

MathQA                     & 20.7 $\pm$ 0.3 & 20.4 $\pm$ 0.2 & 20.7 $\pm$ 0.3 & 20.3 $\pm$ 0.5 & 20.2 $\pm$ 0.3 & 21.1 $\pm$ 0.3 & 20.1 $\pm$ 0.1 & 20.3 $\pm$ 0.1 & 20.8 $\pm$ 0.7 & 20.6 $\pm$ 0.4 & 20.2 $\pm$ 0.3 & 21.0 $\pm$ 1.0 \\
BoolQ                      & 53.2 $\pm$ 0.8 & 53.4 $\pm$ 0.8 & 54.2 $\pm$ 0.3 & 53.2 $\pm$ 0.8 & 57.6 $\pm$ 1.1 & 56.0 $\pm$ 0.2 & 50.8 $\pm$ 0.8 & 53.0 $\pm$ 0.7 & 54.1 $\pm$ 0.1 & 53.1 $\pm$ 0.8 & 53.1 $\pm$ 0.7 & 55.2 $\pm$ 0.4 \\
LAMBADA              &  9.5 $\pm$ 0.1 & 11.9 $\pm$ 0.1 & 15.1 $\pm$ 0.1 &  6.5 $\pm$ 0.1 &  9.9 $\pm$ 0.2 & 13.4 $\pm$ 0.2 &  5.9 $\pm$ 0.2 &  9.6 $\pm$ 0.3 & 13.5 $\pm$ 0.2 &  7.0 $\pm$ 0.1 &  9.2 $\pm$ 0.1 & 13.8 $\pm$ 0.5 \\
\rowcolor{gray!15}
ICL composite (macro)     & 23.0 $\pm$ 0.3 & 23.6 $\pm$ 0.4 & 24.3 $\pm$ 0.4
& 22.6 $\pm$ 0.3 & 23.6 $\pm$ 0.4 & 24.2 $\pm$ 0.3
& 22.4 $\pm$ 0.4 & 23.1 $\pm$ 0.4 & 23.9 $\pm$ 0.4
& 22.7 $\pm$ 0.3 & 23.1 $\pm$ 0.5 & 24.1 $\pm$ 0.5 \\
\midrule
alphabetically\_first\_3 & $2.6\,\pm\,0.6$ & $3.0\,\pm\,0.7$ & $13.2\,\pm\,1.3$ & $3.3\,\pm\,0.7$ & $4.8\,\pm\,0.6$ & $8.9\,\pm\,0.8$ & $1.9\,\pm\,0.4$ & $3.5\,\pm\,0.8$ & $5.8\,\pm\,1.0$ & $3.6\,\pm\,0.7$ & $4.4\,\pm\,0.4$ & $8.5\,\pm\,0.8$ \\ alphabetically\_first\_5 & $2.7\,\pm\,0.4$ & $3.3\,\pm\,0.5$ & $8.9\,\pm\,0.5$ & $3.3\,\pm\,0.1$ & $4.3\,\pm\,0.2$ & $7.5\,\pm\,0.4$ & $2.0\,\pm\,0.3$ & $3.7\,\pm\,0.8$ & $4.4\,\pm\,0.8$ & $2.2\,\pm\,0.4$ & $3.1\,\pm\,0.6$ & $6.8\,\pm\,1.1$ \\ alphabetically\_last\_3 & $2.1\,\pm\,0.5$ & $2.9\,\pm\,0.5$ & $13.3\,\pm\,1.2$ & $3.1\,\pm\,0.4$ & $4.3\,\pm\,0.9$ & $9.2\,\pm\,1.0$ & $1.9\,\pm\,0.4$ & $2.9\,\pm\,0.3$ & $6.8\,\pm\,0.6$ & $4.5\,\pm\,0.3$ & $3.8\,\pm\,0.5$ & $10.6\,\pm\,0.9$ \\ alphabetically\_last\_5 & $1.6\,\pm\,0.5$ & $2.8\,\pm\,0.4$ & $8.5\,\pm\,0.9$ & $1.9\,\pm\,0.6$ & $3.2\,\pm\,0.4$ & $6.0\,\pm\,0.3$ & $1.4\,\pm\,0.4$ & $2.3\,\pm\,0.5$ & $4.5\,\pm\,0.4$ & $2.2\,\pm\,0.4$ & $3.0\,\pm\,0.8$ & $6.8\,\pm\,0.6$ \\ capitalize & $2.9\,\pm\,0.6$ & $1.5\,\pm\,0.4$ & $4.4\,\pm\,0.7$ & $1.8\,\pm\,0.3$ & $5.1\,\pm\,0.8$ & $2.6\,\pm\,0.8$ & $1.8\,\pm\,0.2$ & $2.5\,\pm\,0.5$ & $4.9\,\pm\,0.6$ & $2.7\,\pm\,0.4$ & $3.1\,\pm\,0.5$ & $3.7\,\pm\,1.1$ \\ capitalize\_first\_letter & $4.1\,\pm\,1.1$ & $3.0\,\pm\,0.9$ & $3.4\,\pm\,0.7$ & $3.5\,\pm\,1.0$ & $5.6\,\pm\,1.4$ & $4.0\,\pm\,1.2$ & $3.5\,\pm\,1.0$ & $4.7\,\pm\,1.2$ & $4.0\,\pm\,0.9$ & $4.4\,\pm\,1.0$ & $4.9\,\pm\,0.8$ & $3.5\,\pm\,1.1$ \\ capitalize\_last\_letter & $9.0\,\pm\,0.7$ & $7.4\,\pm\,0.4$ & $5.6\,\pm\,0.9$ & $9.1\,\pm\,0.8$ & $6.0\,\pm\,0.7$ & $8.4\,\pm\,0.4$ & $9.2\,\pm\,0.8$ & $8.7\,\pm\,0.9$ & $8.4\,\pm\,0.8$ & $8.7\,\pm\,0.7$ & $8.8\,\pm\,0.8$ & $8.8\,\pm\,0.5$ \\ choose\_first\_of\_3 & $5.6\,\pm\,0.5$ & $6.7\,\pm\,1.0$ & $37.0\,\pm\,1.2$ & $5.5\,\pm\,0.8$ & $7.4\,\pm\,1.3$ & $25.1\,\pm\,2.1$ & $1.5\,\pm\,0.4$ & $6.0\,\pm\,1.1$ & $14.6\,\pm\,1.0$ & $8.9\,\pm\,1.0$ & $7.2\,\pm\,0.8$ & $26.1\,\pm\,1.8$ \\ choose\_first\_of\_5 & $5.9\,\pm\,0.4$ & $6.9\,\pm\,0.8$ & $32.0\,\pm\,1.2$ & $3.9\,\pm\,0.8$ & $5.6\,\pm\,0.6$ & $24.9\,\pm\,1.9$ & $1.1\,\pm\,0.3$ & $6.1\,\pm\,0.8$ & $12.5\,\pm\,1.9$ & $5.3\,\pm\,0.3$ & $4.8\,\pm\,0.6$ & $23.8\,\pm\,1.4$ \\ choose\_last\_of\_3 & $0.8\,\pm\,0.3$ & $1.1\,\pm\,0.3$ & $3.7\,\pm\,0.6$ & $1.3\,\pm\,0.4$ & $1.9\,\pm\,0.3$ & $2.4\,\pm\,0.5$ & $1.2\,\pm\,0.4$ & $1.3\,\pm\,0.7$ & $2.0\,\pm\,0.4$ & $0.7\,\pm\,0.2$ & $1.9\,\pm\,0.7$ & $3.7\,\pm\,0.3$ \\ choose\_last\_of\_5 & $0.9\,\pm\,0.3$ & $1.1\,\pm\,0.4$ & $2.4\,\pm\,0.4$ & $1.4\,\pm\,0.3$ & $1.5\,\pm\,0.3$ & $2.0\,\pm\,0.4$ & $0.9\,\pm\,0.3$ & $1.3\,\pm\,0.4$ & $1.9\,\pm\,0.6$ & $0.6\,\pm\,0.2$ & $1.3\,\pm\,0.4$ & $3.1\,\pm\,0.6$ \\ choose\_middle\_of\_3 & $0.7\,\pm\,0.3$ & $1.1\,\pm\,0.4$ & $3.5\,\pm\,0.6$ & $0.9\,\pm\,0.1$ & $1.6\,\pm\,0.4$ & $2.5\,\pm\,0.6$ & $0.7\,\pm\,0.2$ & $1.1\,\pm\,0.1$ & $1.9\,\pm\,0.5$ & $0.6\,\pm\,0.1$ & $1.1\,\pm\,0.2$ & $3.6\,\pm\,0.4$ \\ choose\_middle\_of\_5 & $0.9\,\pm\,0.4$ & $1.4\,\pm\,0.5$ & $2.3\,\pm\,0.5$ & $1.2\,\pm\,0.3$ & $1.4\,\pm\,0.2$ & $2.2\,\pm\,0.5$ & $1.2\,\pm\,0.3$ & $1.3\,\pm\,0.5$ & $1.6\,\pm\,0.4$ & $0.8\,\pm\,0.4$ & $1.5\,\pm\,0.7$ & $3.0\,\pm\,0.6$ \\ lowercase\_first\_letter & $4.9\,\pm\,0.7$ & $3.7\,\pm\,0.7$ & $4.0\,\pm\,0.3$ & $4.0\,\pm\,0.7$ & $3.9\,\pm\,0.6$ & $4.6\,\pm\,0.8$ & $2.6\,\pm\,0.5$ & $4.7\,\pm\,0.7$ & $4.7\,\pm\,0.8$ & $2.3\,\pm\,0.5$ & $4.7\,\pm\,0.7$ & $4.4\,\pm\,0.8$ \\ lowercase\_last\_letter & $10.6\,\pm\,1.2$ & $9.3\,\pm\,1.2$ & $9.5\,\pm\,1.2$ & $6.6\,\pm\,0.9$ & $7.7\,\pm\,1.3$ & $10.6\,\pm\,1.2$ & $7.6\,\pm\,0.5$ & $10.6\,\pm\,1.2$ & $10.6\,\pm\,1.2$ & $8.7\,\pm\,1.3$ & $10.6\,\pm\,1.2$ & $10.3\,\pm\,1.2$ \\ next\_capital\_letter & $4.5\,\pm\,0.3$ & $3.8\,\pm\,0.6$ & $3.6\,\pm\,1.0$ & $4.6\,\pm\,0.5$ & $3.9\,\pm\,0.4$ & $4.5\,\pm\,0.4$ & $4.4\,\pm\,0.4$ & $4.7\,\pm\,0.1$ & $4.1\,\pm\,0.5$ & $4.1\,\pm\,0.4$ & $4.5\,\pm\,0.4$ & $4.6\,\pm\,0.5$ \\ next\_item & $2.5\,\pm\,0.4$ & $2.0\,\pm\,1.7$ & $9.7\,\pm\,1.0$ & $3.9\,\pm\,0.8$ & $5.1\,\pm\,1.4$ & $5.1\,\pm\,1.2$ & $1.5\,\pm\,1.1$ & $2.8\,\pm\,1.1$ & $3.5\,\pm\,0.8$ & $4.4\,\pm\,2.1$ & $4.2\,\pm\,1.1$ & $3.4\,\pm\,1.5$ \\ prev\_item & $2.5\,\pm\,1.2$ & $2.3\,\pm\,1.7$ & $9.2\,\pm\,1.6$ & $3.0\,\pm\,1.1$ & $5.1\,\pm\,1.6$ & $5.7\,\pm\,2.0$ & $1.8\,\pm\,0.7$ & $3.3\,\pm\,1.9$ & $3.4\,\pm\,1.9$ & $4.6\,\pm\,0.7$ & $3.0\,\pm\,1.5$ & $3.8\,\pm\,1.3$ \\ word\_length & $13.8\,\pm\,1.9$ & $13.7\,\pm\,1.7$ & $12.7\,\pm\,2.2$ & $12.7\,\pm\,1.5$ & $13.4\,\pm\,2.1$ & $13.6\,\pm\,1.5$ & $12.5\,\pm\,1.6$ & $13.9\,\pm\,1.7$ & $13.9\,\pm\,1.4$ & $10.6\,\pm\,1.0$ & $13.9\,\pm\,1.7$ & $13.9\,\pm\,1.9$ \\ 
\rowcolor{gray!15} 
ICL composite (macro) $\uparrow$ & $4.1\,\pm\,0.6$ & $4.1\,\pm\,0.8$ & $9.8\,\pm\,0.9$ & $3.9\,\pm\,0.6$ & $4.8\,\pm\,0.8$ & $7.9\,\pm\,1.0$ & $3.1\,\pm\,0.5$ & $4.5\,\pm\,0.8$ & $6.0\,\pm\,0.9$ & $4.2\,\pm\,0.6$ & $4.7\,\pm\,0.7$ & $8.0\,\pm\,1.0$ \\ 
\bottomrule
\end{tabular}}
\end{table*}

\subsubsection{Function\mbox{-}Probe Task Stress Tests}
Because stress tests were explicitly considered during the development of the \cite{todd2024functionvectorslargelanguage} function\mbox{-}probe suite, we evaluate two that directly target ICL robustness: (i) \emph{label permutation} within the in\mbox{-}context demonstration shots, and (ii) \emph{decision\mbox{-}rule sensitivity}, contrasting the commonly reported HITS@3 with our primary metric, HITS@1. These tests probe robustness to spurious label-token mappings and to the choice of evaluation rule, respectively.

\begin{figure}[t]
  \centering
  \includegraphics[width=\linewidth]{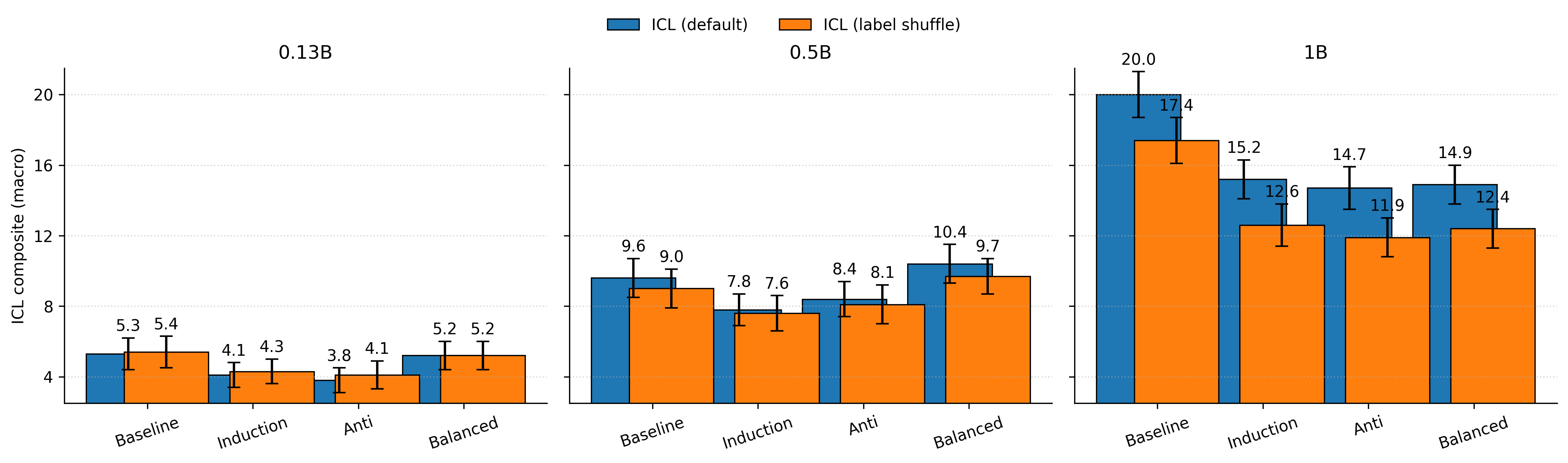}
  \caption{Function-probe suite of \cite{todd2024functionvectorslargelanguage} - ICL Composite (macro). Three panels (0.13B, 0.5B, 1B). For each model, bars compare \emph{ICL (default - no label shuffle)} vs \emph{ICL (label shuffle)} across four regimes (Baseline, Induction, Anti, Balanced). Error bars show $\pm$1 s.d.}
  \label{fig:todd-icl-shuffle}
\end{figure}

\paragraph{Label\mbox{-}Permutation Stress Test:}
\label{sec:label_perm_test}
We stress\mbox{-}test in\mbox{-}context usage on the \citet{todd2024functionvectorslargelanguage} probes by randomly permuting the targets within the $10$ demonstration shots (inputs unchanged) and evaluating on the true task distribution. If a model relies on the demonstrations, accuracy should drop; if it leans on parametric priors, it should be less affected. As shown in Figure~\ref{fig:todd-icl-shuffle} and Table~\ref{tab:icl_probes_shuffle}, the relative ordering between Baseline and Bi\mbox{-}Induct variants is largely preserved. At \textbf{0.13B}, permutation produces no degradation for either, suggesting heavier reliance on parametric knowledge. At \textbf{0.5B} and \textbf{1B}, all curricula degrade, indicating increased sensitivity to the in\mbox{-}context mapping. Overall, Bi\mbox{-}Induct mirrors Baseline at each scale: robust at 0.13B and increasingly demonstration\mbox{-}sensitive as scale grows.

\textit{\textbf{Why permutation hurts 0.5B-1B but not 0.13B?}}
At 0.13B, the robustness to label permutation suggests it benefits from demonstrations via format/topical priming and answer-frequency priors, but does not reliably exploit the label$\to$token mapping. In contrast, 0.5B-1B models more strongly use the in-context mapping; permuting labels therefore contradicts a cue they have learned to trust, producing clear drops. This is consistent with reports that (i) labels in demos can be non-essential for smaller/less capable settings-format and priors often dominate \citep{min2022rethinkingroledemonstrationsmakes, zhao2021calibrateuseimprovingfewshot}, and (ii) the ability to override priors and follow contradictory, flipped labels emerges with scale \citep{wei2023largerlanguagemodelsincontext}.

\begin{table*}[t]
\centering
\small
\setlength{\tabcolsep}{3.5pt}
\renewcommand{\arraystretch}{1.15}
\caption{Function-probe suite of \cite{todd2024functionvectorslargelanguage} under label-permutation stress (\cite{todd2024functionvectorslargelanguage} suite): HITS@1 accuracy on 10-shot prompts with demonstration labels randomly permuted; reported as mean\(\pm\)std across three seeds for 0.13B, 0.5B, and 1B, comparing Baseline, Induction, Anti, and Balanced curricula.}

\label{tab:icl_probes_shuffle}
\resizebox{\textwidth}{!}{
\begin{tabular}{l ccc ccc ccc ccc}
\toprule
& \multicolumn{3}{c}{\textbf{Baseline}} & \multicolumn{3}{c}{\textbf{Induction}} & \multicolumn{3}{c}{\textbf{Anti-induction}} & \multicolumn{3}{c}{\textbf{Balanced}} \\
\cmidrule(lr){2-4}\cmidrule(lr){5-7}\cmidrule(lr){8-10}\cmidrule(lr){11-13}
 & 0.13B & 0.5B & 1B & 0.13B & 0.5B & 1B & 0.13B & 0.5B & 1B & 0.13B & 0.5B & 1B \\
\midrule
alphabetically\_first\_3 & $4.4\,\pm\,0.4$ & $9.2\,\pm\,0.7$ & $16.6\,\pm\,0.6$ & $4.4\,\pm\,0.3$ & $7.0\,\pm\,0.7$ & $12.3\,\pm\,1.4$ & $3.6\,\pm\,0.5$ & $8.2\,\pm\,0.8$ & $11.0\,\pm\,1.3$ & $4.2\,\pm\,0.7$ & $10.7\,\pm\,1.4$ & $12.5\,\pm\,0.8$ \\
alphabetically\_first\_5 & $4.1\,\pm\,0.7$ & $7.1\,\pm\,1.1$ & $11.1\,\pm\,0.9$ & $4.2\,\pm\,0.6$ & $5.9\,\pm\,0.4$ & $9.4\,\pm\,0.5$ & $2.6\,\pm\,0.9$ & $7.0\,\pm\,0.9$ & $7.0\,\pm\,0.4$ & $3.5\,\pm\,0.4$ & $7.8\,\pm\,1.0$ & $9.0\,\pm\,0.5$ \\
alphabetically\_last\_3 & $3.2\,\pm\,0.6$ & $9.7\,\pm\,0.8$ & $16.1\,\pm\,1.6$ & $2.6\,\pm\,0.6$ & $7.3\,\pm\,0.9$ & $12.3\,\pm\,0.6$ & $2.7\,\pm\,1.0$ & $7.4\,\pm\,1.3$ & $10.4\,\pm\,0.9$ & $4.8\,\pm\,0.6$ & $10.7\,\pm\,1.4$ & $13.0\,\pm\,0.6$ \\
alphabetically\_last\_5 & $2.6\,\pm\,0.3$ & $5.8\,\pm\,0.9$ & $9.5\,\pm\,0.2$ & $1.6\,\pm\,0.3$ & $5.3\,\pm\,0.3$ & $7.8\,\pm\,0.7$ & $1.6\,\pm\,0.5$ & $5.5\,\pm\,0.6$ & $6.4\,\pm\,0.8$ & $2.0\,\pm\,0.2$ & $6.7\,\pm\,0.5$ & $8.5\,\pm\,0.9$ \\
capitalize & $8.7\,\pm\,0.8$ & $18.2\,\pm\,1.9$ & $49.3\,\pm\,2.9$ & $3.6\,\pm\,1.0$ & $13.3\,\pm\,1.8$ & $29.9\,\pm\,0.8$ & $4.3\,\pm\,0.7$ & $13.5\,\pm\,1.6$ & $34.3\,\pm\,1.4$ & $6.8\,\pm\,0.6$ & $14.5\,\pm\,0.9$ & $31.0\,\pm\,1.9$ \\
capitalize\_first\_letter & $10.4\,\pm\,2.1$ & $13.2\,\pm\,1.4$ & $29.0\,\pm\,1.4$ & $6.4\,\pm\,0.4$ & $13.3\,\pm\,0.8$ & $15.2\,\pm\,1.4$ & $6.4\,\pm\,0.9$ & $12.2\,\pm\,1.9$ & $17.5\,\pm\,1.3$ & $9.2\,\pm\,1.5$ & $10.8\,\pm\,2.2$ & $21.2\,\pm\,0.7$ \\
capitalize\_last\_letter & $4.7\,\pm\,1.0$ & $8.2\,\pm\,1.0$ & $7.7\,\pm\,0.7$ & $7.9\,\pm\,1.2$ & $6.7\,\pm\,0.8$ & $7.4\,\pm\,1.3$ & $8.9\,\pm\,1.7$ & $8.6\,\pm\,0.9$ & $5.9\,\pm\,1.4$ & $5.2\,\pm\,0.7$ & $5.3\,\pm\,0.9$ & $6.5\,\pm\,1.4$ \\
choose\_first\_of\_3 & $11.2\,\pm\,1.2$ & $21.9\,\pm\,2.1$ & $54.9\,\pm\,2.6$ & $6.9\,\pm\,1.2$ & $19.1\,\pm\,2.5$ & $35.0\,\pm\,2.5$ & $5.2\,\pm\,0.4$ & $19.0\,\pm\,2.0$ & $28.9\,\pm\,0.8$ & $13.4\,\pm\,1.0$ & $32.4\,\pm\,1.1$ & $34.5\,\pm\,1.6$ \\
choose\_first\_of\_5 & $9.3\,\pm\,1.5$ & $17.3\,\pm\,1.6$ & $42.0\,\pm\,2.3$ & $5.6\,\pm\,0.9$ & $15.3\,\pm\,1.8$ & $28.9\,\pm\,2.1$ & $3.6\,\pm\,0.4$ & $16.8\,\pm\,1.9$ & $20.3\,\pm\,1.3$ & $8.5\,\pm\,1.7$ & $23.7\,\pm\,1.9$ & $26.4\,\pm\,1.9$ \\
choose\_last\_of\_3 & $1.6\,\pm\,0.4$ & $2.7\,\pm\,0.6$ & $4.3\,\pm\,0.9$ & $2.1\,\pm\,0.6$ & $2.6\,\pm\,0.4$ & $4.9\,\pm\,0.5$ & $1.3\,\pm\,0.4$ & $2.8\,\pm\,0.3$ & $4.9\,\pm\,1.2$ & $1.3\,\pm\,0.3$ & $3.3\,\pm\,0.7$ & $5.4\,\pm\,0.6$ \\
choose\_last\_of\_5 & $1.7\,\pm\,0.4$ & $2.2\,\pm\,0.6$ & $3.9\,\pm\,0.8$ & $1.9\,\pm\,0.4$ & $2.1\,\pm\,0.3$ & $4.5\,\pm\,0.6$ & $1.1\,\pm\,0.4$ & $2.3\,\pm\,0.6$ & $5.5\,\pm\,0.7$ & $1.2\,\pm\,0.5$ & $2.3\,\pm\,0.8$ & $4.7\,\pm\,0.3$ \\
choose\_middle\_of\_3 & $1.9\,\pm\,0.3$ & $3.5\,\pm\,0.5$ & $4.5\,\pm\,0.9$ & $2.3\,\pm\,0.6$ & $2.8\,\pm\,0.7$ & $5.6\,\pm\,1.1$ & $1.4\,\pm\,0.5$ & $3.0\,\pm\,0.4$ & $4.6\,\pm\,0.3$ & $1.5\,\pm\,0.2$ & $3.1\,\pm\,0.3$ & $5.2\,\pm\,0.8$ \\
choose\_middle\_of\_5 & $1.8\,\pm\,0.7$ & $3.6\,\pm\,0.8$ & $2.9\,\pm\,0.5$ & $1.9\,\pm\,0.3$ & $2.2\,\pm\,0.6$ & $3.3\,\pm\,0.4$ & $1.5\,\pm\,0.6$ & $2.8\,\pm\,0.4$ & $3.9\,\pm\,0.5$ & $1.7\,\pm\,0.2$ & $2.4\,\pm\,0.4$ & $4.6\,\pm\,0.7$ \\
lowercase\_first\_letter & $6.7\,\pm\,0.8$ & $9.3\,\pm\,0.6$ & $27.8\,\pm\,1.6$ & $4.9\,\pm\,0.7$ & $6.5\,\pm\,0.4$ & $19.0\,\pm\,1.1$ & $5.4\,\pm\,1.1$ & $7.5\,\pm\,1.1$ & $18.4\,\pm\,1.9$ & $2.5\,\pm\,0.6$ & $10.4\,\pm\,0.9$ & $13.9\,\pm\,1.4$ \\
lowercase\_last\_letter & $10.1\,\pm\,1.0$ & $7.5\,\pm\,1.2$ & $8.8\,\pm\,1.9$ & $3.5\,\pm\,1.2$ & $8.0\,\pm\,1.6$ & $8.2\,\pm\,1.2$ & $7.1\,\pm\,0.9$ & $8.4\,\pm\,1.7$ & $12.8\,\pm\,1.8$ & $7.1\,\pm\,0.7$ & $9.5\,\pm\,1.3$ & $7.8\,\pm\,1.5$ \\
next\_capital\_letter & $4.3\,\pm\,0.9$ & $4.3\,\pm\,0.5$ & $2.8\,\pm\,0.4$ & $4.2\,\pm\,0.6$ & $3.5\,\pm\,0.6$ & $3.6\,\pm\,0.6$ & $4.7\,\pm\,0.6$ & $4.2\,\pm\,0.7$ & $3.0\,\pm\,0.6$ & $4.3\,\pm\,0.5$ & $3.3\,\pm\,0.6$ & $3.0\,\pm\,0.1$ \\
next\_item & $3.3\,\pm\,1.0$ & $7.3\,\pm\,1.3$ & $15.3\,\pm\,1.8$ & $2.7\,\pm\,0.8$ & $4.7\,\pm\,0.8$ & $10.8\,\pm\,2.2$ & $1.8\,\pm\,1.0$ & $4.9\,\pm\,1.2$ & $9.7\,\pm\,2.7$ & $5.3\,\pm\,1.5$ & $6.6\,\pm\,1.4$ & $7.2\,\pm\,1.0$ \\
prev\_item & $3.2\,\pm\,1.7$ & $6.5\,\pm\,2.3$ & $13.5\,\pm\,1.6$ & $2.5\,\pm\,1.4$ & $5.6\,\pm\,1.4$ & $8.6\,\pm\,1.8$ & $2.3\,\pm\,1.5$ & $4.9\,\pm\,1.1$ & $9.2\,\pm\,1.1$ & $4.8\,\pm\,1.7$ & $6.5\,\pm\,1.4$ & $7.8\,\pm\,1.0$ \\
word\_length & $9.7\,\pm\,0.8$ & $12.6\,\pm\,1.3$ & $11.4\,\pm\,1.8$ & $12.8\,\pm\,1.0$ & $12.9\,\pm\,1.4$ & $12.4\,\pm\,1.1$ & $12.1\,\pm\,1.1$ & $14.5\,\pm\,1.6$ & $13.2\,\pm\,0.2$ & $11.1\,\pm\,0.9$ & $14.3\,\pm\,0.8$ & $12.6\,\pm\,1.0$ \\
\rowcolor{gray!15}
ICL composite (macro) $\uparrow$ & $5.4\,\pm\,0.9$ & $9.0\,\pm\,1.1$ & $17.4\,\pm\,1.3$ & $4.3\,\pm\,0.7$ & $7.6\,\pm\,1.0$ & $12.6\,\pm\,1.2$ & $4.1\,\pm\,0.8$ & $8.1\,\pm\,1.1$ & $11.9\,\pm\,1.1$ & $5.2\,\pm\,0.8$ & $9.7\,\pm\,1.0$ & $12.4\,\pm\,1.0$ \\
\bottomrule
\end{tabular}}
\end{table*}

\begin{figure}[t]
  \centering
  \includegraphics[width=\linewidth]{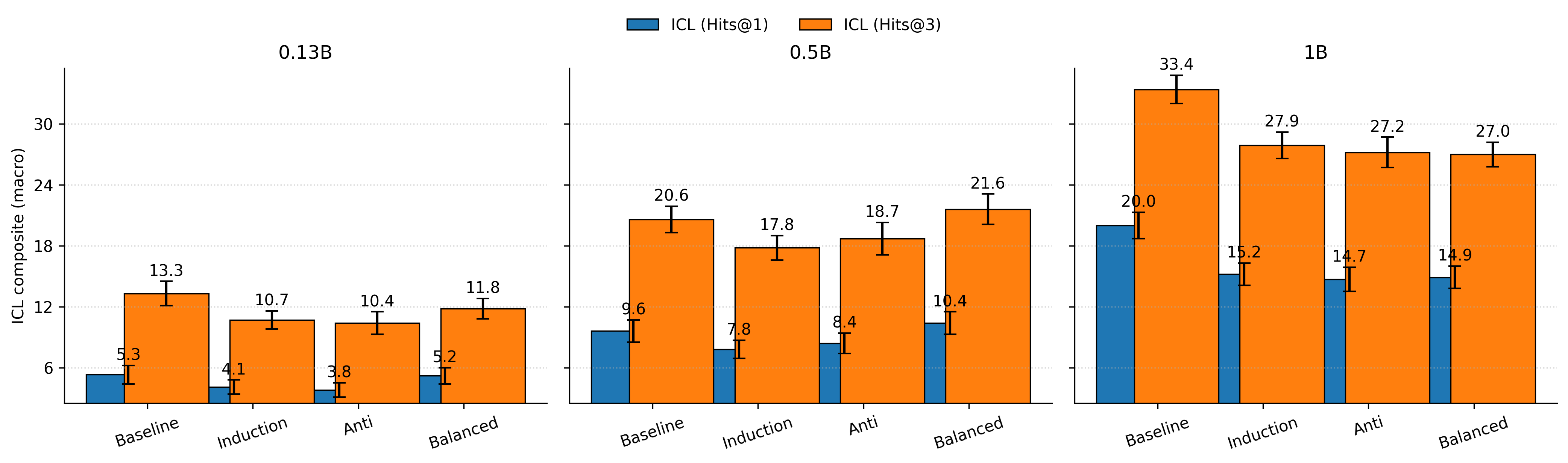}
  \caption{Function-probe suite of \cite{todd2024functionvectorslargelanguage} - ICL Composite (macro). Three panels (0.13B, 0.5B, 1B). For each model, bars compare \emph{ICL (HITS@1)} vs \emph{ICL (HITS@3)} across four regimes (Baseline, Induction, Anti, Balanced). Error bars show $\pm$1 s.d.}
  \label{fig:todd-icl-hits3}
\end{figure}

\paragraph{Decision\mbox{-}Rule Sensitivity - HITS@$k$ ($k\!\in\!\{1,3\}$):}
Following common practice for function-probe tasks in the \cite{todd2024functionvectorslargelanguage} suite-which reports HITS@3 (top-3 token accuracy)-we compare our primary decision rule (HITS@1) with HITS@3. ICL performance is summarized in Figure~\ref{fig:todd-icl-hits3}, and per-task HITS@3 accuracies are listed in Table~\ref{tab:icl_hits3_models_runs}. While HITS@3 increases absolute scores across the board, the relative ordering and gaps between variants remain effectively unchanged.

\begin{table*}[t]
\centering
\small
\setlength{\tabcolsep}{4pt}
\renewcommand{\arraystretch}{1.15}
\caption{Function-probe suite of \cite{todd2024functionvectorslargelanguage} under decision-rule sensitivity: HITS@3 accuracy on 10-shot prompts, reported as mean\(\pm\)std across three seeds, for 0.13B, 0.5B, and 1B, comparing Baseline, Induction, Anti, and Balanced curricula.}

\label{tab:icl_hits3_models_runs}
\resizebox{\textwidth}{!}{
\begin{tabular}{lcccccccccccc}
\toprule
& \multicolumn{3}{c}{\textbf{Baseline}} & \multicolumn{3}{c}{\textbf{Induction}} & \multicolumn{3}{c}{\textbf{Anti-induction}} & \multicolumn{3}{c}{\textbf{Balanced}} \\
\cmidrule(lr){2-4}\cmidrule(lr){5-7}\cmidrule(lr){8-10}\cmidrule(lr){11-13}
 & 0.13B & 0.5B & 1B & 0.13B & 0.5B & 1B & 0.13B & 0.5B & 1B & 0.13B & 0.5B & 1B \\ 
\midrule
alphabetically\_first\_3 & $8.8\,\pm\,0.8$ & $15.3\,\pm\,0.5$ & $31.0\,\pm\,0.5$ & $8.6\,\pm\,1.3$ & $12.7\,\pm\,0.3$ & $23.0\,\pm\,0.8$ & $7.1\,\pm\,0.3$ & $13.7\,\pm\,0.8$ & $22.6\,\pm\,1.4$ & $9.3\,\pm\,1.2$ & $18.6\,\pm\,0.6$ & $24.1\,\pm\,0.8$ \\
alphabetically\_first\_5 & $8.6\,\pm\,1.3$ & $13.0\,\pm\,0.5$ & $20.7\,\pm\,1.5$ & $8.3\,\pm\,0.9$ & $12.1\,\pm\,1.0$ & $16.4\,\pm\,0.6$ & $7.4\,\pm\,0.7$ & $13.6\,\pm\,1.3$ & $15.5\,\pm\,1.3$ & $7.8\,\pm\,1.3$ & $13.3\,\pm\,0.6$ & $15.9\,\pm\,0.8$ \\
alphabetically\_last\_3 & $7.3\,\pm\,0.9$ & $17.4\,\pm\,0.7$ & $30.3\,\pm\,1.0$ & $6.3\,\pm\,0.7$ & $14.3\,\pm\,0.5$ & $24.5\,\pm\,1.1$ & $5.7\,\pm\,0.7$ & $14.6\,\pm\,1.5$ & $22.3\,\pm\,1.5$ & $8.9\,\pm\,0.3$ & $19.1\,\pm\,1.1$ & $24.0\,\pm\,0.9$ \\
alphabetically\_last\_5 & $5.8\,\pm\,0.8$ & $12.6\,\pm\,1.7$ & $18.6\,\pm\,1.1$ & $4.9\,\pm\,0.8$ & $11.4\,\pm\,0.8$ & $14.6\,\pm\,0.7$ & $4.5\,\pm\,0.5$ & $10.9\,\pm\,1.4$ & $14.0\,\pm\,1.0$ & $5.6\,\pm\,0.8$ & $12.9\,\pm\,0.9$ & $15.6\,\pm\,1.0$ \\
capitalize & $17.8\,\pm\,1.1$ & $39.7\,\pm\,1.6$ & $73.8\,\pm\,1.4$ & $10.6\,\pm\,0.4$ & $28.5\,\pm\,0.9$ & $57.7\,\pm\,1.3$ & $11.5\,\pm\,1.2$ & $28.9\,\pm\,1.3$ & $60.6\,\pm\,0.8$ & $14.0\,\pm\,2.0$ & $32.2\,\pm\,1.4$ & $56.0\,\pm\,0.8$ \\
capitalize\_first\_letter & $25.6\,\pm\,1.6$ & $34.7\,\pm\,1.8$ & $55.2\,\pm\,2.0$ & $17.1\,\pm\,1.1$ & $30.7\,\pm\,1.7$ & $40.1\,\pm\,1.6$ & $19.1\,\pm\,1.5$ & $29.9\,\pm\,1.7$ & $43.5\,\pm\,2.1$ & $22.3\,\pm\,1.7$ & $29.6\,\pm\,1.3$ & $42.0\,\pm\,2.1$ \\
capitalize\_last\_letter & $14.1\,\pm\,1.7$ & $25.6\,\pm\,1.7$ & $20.4\,\pm\,2.8$ & $20.0\,\pm\,1.1$ & $18.8\,\pm\,1.9$ & $23.0\,\pm\,2.4$ & $20.8\,\pm\,2.1$ & $23.2\,\pm\,2.8$ & $17.2\,\pm\,1.8$ & $15.6\,\pm\,0.8$ & $15.9\,\pm\,1.7$ & $18.6\,\pm\,2.7$ \\
choose\_first\_of\_3 & $18.3\,\pm\,1.9$ & $38.6\,\pm\,1.9$ & $83.6\,\pm\,0.6$ & $12.9\,\pm\,1.0$ & $27.8\,\pm\,1.7$ & $66.6\,\pm\,1.6$ & $9.8\,\pm\,1.1$ & $28.3\,\pm\,1.2$ & $60.7\,\pm\,0.9$ & $19.5\,\pm\,1.1$ & $50.3\,\pm\,2.1$ & $67.7\,\pm\,1.0$ \\
choose\_first\_of\_5 & $17.1\,\pm\,1.7$ & $30.6\,\pm\,1.9$ & $74.1\,\pm\,1.1$ & $11.3\,\pm\,0.7$ & $24.4\,\pm\,1.6$ & $56.2\,\pm\,1.3$ & $8.0\,\pm\,1.3$ & $26.3\,\pm\,1.7$ & $48.3\,\pm\,1.5$ & $15.4\,\pm\,1.2$ & $41.5\,\pm\,1.8$ & $56.7\,\pm\,1.8$ \\
choose\_last\_of\_3 & $5.1\,\pm\,0.4$ & $8.0\,\pm\,0.6$ & $11.0\,\pm\,1.0$ & $4.5\,\pm\,0.9$ & $8.7\,\pm\,0.7$ & $11.5\,\pm\,1.3$ & $3.9\,\pm\,0.5$ & $7.9\,\pm\,0.6$ & $11.6\,\pm\,0.6$ & $4.0\,\pm\,0.6$ & $8.7\,\pm\,1.2$ & $12.4\,\pm\,0.4$ \\
choose\_last\_of\_5 & $4.0\,\pm\,0.6$ & $7.1\,\pm\,0.7$ & $9.3\,\pm\,1.1$ & $4.9\,\pm\,0.6$ & $7.0\,\pm\,0.8$ & $10.3\,\pm\,0.7$ & $3.5\,\pm\,0.6$ & $7.6\,\pm\,0.8$ & $10.1\,\pm\,0.6$ & $4.1\,\pm\,0.7$ & $6.6\,\pm\,0.4$ & $10.6\,\pm\,1.0$ \\
choose\_middle\_of\_3 & $4.8\,\pm\,1.1$ & $8.7\,\pm\,0.7$ & $12.5\,\pm\,1.0$ & $5.5\,\pm\,0.4$ & $7.7\,\pm\,0.7$ & $13.1\,\pm\,0.7$ & $4.0\,\pm\,0.2$ & $7.8\,\pm\,0.4$ & $10.3\,\pm\,0.7$ & $4.4\,\pm\,0.6$ & $8.1\,\pm\,1.1$ & $11.4\,\pm\,0.7$ \\
choose\_middle\_of\_5 & $4.6\,\pm\,0.7$ & $7.6\,\pm\,0.5$ & $7.7\,\pm\,0.8$ & $4.5\,\pm\,0.2$ & $6.2\,\pm\,0.8$ & $7.4\,\pm\,0.9$ & $4.5\,\pm\,0.9$ & $7.4\,\pm\,1.1$ & $8.4\,\pm\,0.7$ & $4.2\,\pm\,0.8$ & $6.5\,\pm\,0.9$ & $9.3\,\pm\,1.0$ \\
lowercase\_first\_letter & $19.4\,\pm\,2.0$ & $28.7\,\pm\,0.9$ & $58.6\,\pm\,2.3$ & $11.4\,\pm\,1.0$ & $22.1\,\pm\,1.4$ & $49.5\,\pm\,1.9$ & $12.4\,\pm\,1.7$ & $22.8\,\pm\,3.2$ & $45.1\,\pm\,2.2$ & $5.5\,\pm\,0.3$ & $30.5\,\pm\,1.6$ & $38.7\,\pm\,0.5$ \\
lowercase\_last\_letter & $28.0\,\pm\,1.4$ & $19.4\,\pm\,1.1$ & $23.5\,\pm\,1.2$ & $8.7\,\pm\,1.2$ & $23.1\,\pm\,0.6$ & $26.2\,\pm\,1.3$ & $20.2\,\pm\,1.1$ & $24.0\,\pm\,0.9$ & $29.7\,\pm\,0.9$ & $13.2\,\pm\,1.2$ & $26.8\,\pm\,2.0$ & $22.8\,\pm\,1.8$ \\
next\_capital\_letter & $15.0\,\pm\,0.5$ & $13.2\,\pm\,1.2$ & $12.5\,\pm\,0.7$ & $13.2\,\pm\,0.9$ & $12.4\,\pm\,1.8$ & $11.7\,\pm\,1.4$ & $16.0\,\pm\,1.0$ & $13.7\,\pm\,1.4$ & $10.9\,\pm\,2.0$ & $12.5\,\pm\,0.8$ & $11.1\,\pm\,1.4$ & $13.3\,\pm\,1.5$ \\
next\_item & $8.5\,\pm\,2.2$ & $18.7\,\pm\,3.9$ & $29.5\,\pm\,3.2$ & $9.0\,\pm\,1.9$ & $15.8\,\pm\,1.9$ & $23.4\,\pm\,0.8$ & $5.8\,\pm\,2.0$ & $18.2\,\pm\,3.4$ & $25.4\,\pm\,1.9$ & $13.5\,\pm\,1.2$ & $20.5\,\pm\,3.0$ & $20.3\,\pm\,2.1$ \\
prev\_item & $8.4\,\pm\,0.8$ & $15.1\,\pm\,1.4$ & $24.7\,\pm\,2.4$ & $7.6\,\pm\,1.2$ & $14.9\,\pm\,2.1$ & $17.3\,\pm\,1.3$ & $5.3\,\pm\,1.8$ & $17.0\,\pm\,2.9$ & $20.1\,\pm\,3.7$ & $13.8\,\pm\,0.8$ & $17.1\,\pm\,2.1$ & $17.6\,\pm\,0.8$ \\
word\_length & $31.0\,\pm\,2.2$ & $36.5\,\pm\,2.1$ & $37.6\,\pm\,0.6$ & $33.4\,\pm\,1.4$ & $40.2\,\pm\,2.2$ & $38.4\,\pm\,2.2$ & $28.5\,\pm\,1.1$ & $39.9\,\pm\,1.9$ & $40.0\,\pm\,3.2$ & $31.5\,\pm\,1.0$ & $41.1\,\pm\,2.4$ & $36.6\,\pm\,1.1$ \\
\rowcolor{gray!15}
ICL composite (macro) $\uparrow$ & $13.3\,\pm\,1.2$ & $20.6\,\pm\,1.3$ & $33.4\,\pm\,1.4$ & $10.7\,\pm\,0.9$ & $17.8\,\pm\,1.2$ & $27.9\,\pm\,1.3$ & $10.4\,\pm\,1.1$ & $18.7\,\pm\,1.6$ & $27.2\,\pm\,1.5$ & $11.8\,\pm\,1.0$ & $21.6\,\pm\,1.5$ & $27.0\,\pm\,1.2$ \\
\bottomrule
\end{tabular}}
\end{table*}

\section{Induction Head ablation}\label{app:head_ablation}
We quantify how much the in-context learning (ICL) composite depends on the model’s most induction-like attention heads by ablating them at evaluation time and comparing the drop to ablating an equal number of random heads.

\paragraph{Selecting induction heads:}
For each model, we compute a per-head \emph{copy score} exactly as in Section~\ref{sec:copy_scores} and select the top 2\% per layer for ablation.

\paragraph{Ablation mechanism (value-stream zeroing):}
At inference, for a chosen set of heads $\mathcal{S}_\ell$ in layer $\ell$, we zero their value-stream contribution before the output projection:
\[
\tilde{Z}^{(\ell)} \;=\; \big[\, QK\!V^{(1)}\; \|\; \dots \; \|\; \underbrace{0}_{h\in\mathcal{S}_\ell}\; \|\; \dots \; \|\; QK\!V^{(H)} \,\big], 
\qquad
\text{attn\_out}^{(\ell)} \;=\; \tilde{Z}^{(\ell)} W_O^{(\ell)}.
\]
Queries/keys/softmax are unchanged; only the selected heads’ post-attention vectors are set to zero. This follows common practice in mechanistic interpretability and avoids softmax renormalization artifacts. We compare two conditions:
\begin{itemize}\itemsep0.2em
\item \textbf{Induct-head ablation:} zero the per-layer top-2\% induction heads defined above.
\item \textbf{Random-head ablation:} zero the same count of uniformly random heads per layer.
\end{itemize}

We evaluate the same prompts, shots, and metrics as in the main text (Section~\ref{sec:icl_performance}): the macro-averaged ICL composite aggregates task scores (e.g., \textsc{HITS@}1 unless otherwise specified). For each model-curriculum pair we report (i) the clean score, and (ii) the percent change under each ablation relative to its own clean run:
\[
\Delta_{\text{induct}} = 100 \times \frac{\text{ICL}_{\text{induct-abl}} - \text{ICL}_{\text{clean}}}{\text{ICL}_{\text{clean}}}, 
\qquad
\Delta_{\text{rand}} = 100 \times \frac{\text{ICL}_{\text{rand-abl}} - \text{ICL}_{\text{clean}}}{\text{ICL}_{\text{clean}}}.
\]

\begin{figure}[t]
  \centering
  \includegraphics[width=\linewidth]{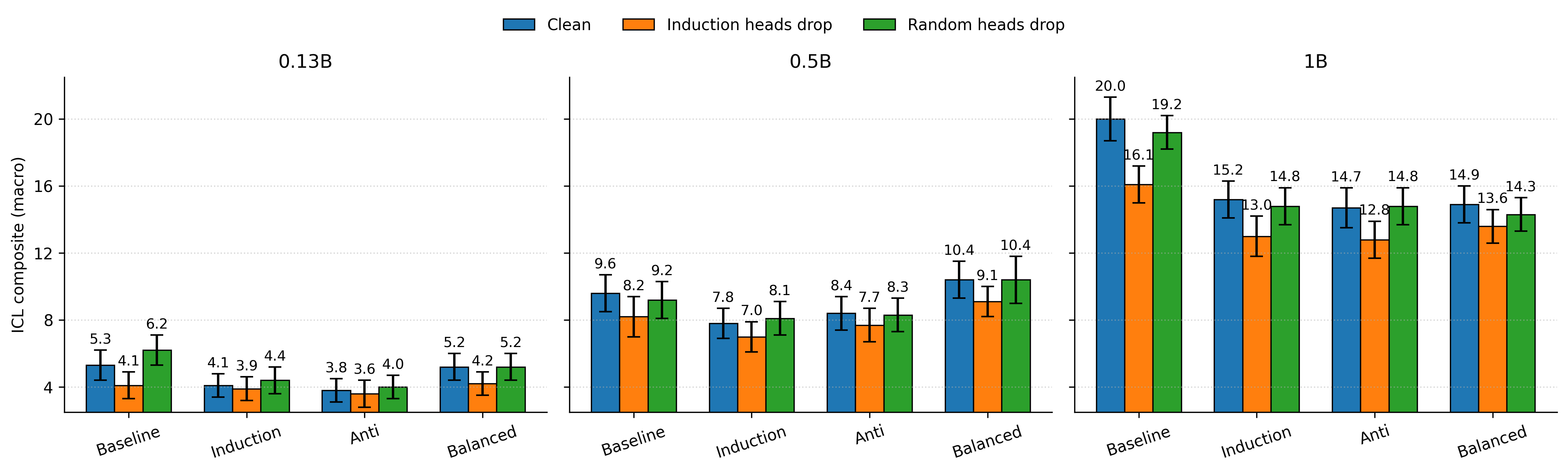}
  \caption{\textbf{Function-probe suite of \cite{todd2024functionvectorslargelanguage} - ICL composite under clean, induct-head ablation, and random-head ablation.} Three panels (0.13B, 0.5B, 1B),
  across four regimes (Baseline, Induction, Anti, Balanced). Error bars show $\pm$1 s.d.}
  \label{fig:app-icl-composite-head_abaltion}
\end{figure}

Figure~\ref{fig:app-icl-composite-head_abaltion} shows the ICL composite for clean vs.\ ablations across scales and curricula; per-task deltas are in Table~\ref{tab:head-ablate-all}.

\begin{table*}[t]
\centering
\small
\setlength{\tabcolsep}{8pt}
\renewcommand{\arraystretch}{1.35}
\caption{Function-probe suite of \cite{todd2024functionvectorslargelanguage}: HITS@1 accuracy on 10-shots prompts, reported as mean\(\pm\)std across three seeds, for 0.13B, 0.5B, and 1B, comparing  Clean run, top-2\% induction heads drop (\(\downarrow\)\textbf{Induct Hd\,2\%}), random heads drop (\(\downarrow\)\textbf{Rand}) across  Baseline, Induction, Anti, and Balanced curricula.}
\label{tab:head-ablate-all}
\adjustbox{max width=\textwidth, max height=\textheight}{%
\begin{tabular}{lcccccccccccccccccccccccccccccccccccc}
\toprule
 & \multicolumn{9}{c}{Baseline} & \multicolumn{9}{c}{Induction} & \multicolumn{9}{c}{Anti-Induction} & \multicolumn{9}{c}{Balanced} \\
\cmidrule(lr){2-10}\cmidrule(lr){11-19}\cmidrule(lr){20-28}\cmidrule(lr){29-37}
 & \multicolumn{3}{c}{0.13B} & \multicolumn{3}{c}{0.5B} & \multicolumn{3}{c}{1B} & \multicolumn{3}{c}{0.13B} & \multicolumn{3}{c}{0.5B} & \multicolumn{3}{c}{1B} & \multicolumn{3}{c}{0.13B} & \multicolumn{3}{c}{0.5B} & \multicolumn{3}{c}{1B} & \multicolumn{3}{c}{0.13B} & \multicolumn{3}{c}{0.5B} & \multicolumn{3}{c}{1B} \\
\cmidrule(lr){2-4}\cmidrule(lr){5-7}\cmidrule(lr){8-10}\cmidrule(lr){11-13}\cmidrule(lr){14-16}\cmidrule(lr){17-19}\cmidrule(lr){20-22}\cmidrule(lr){23-25}\cmidrule(lr){26-28}\cmidrule(lr){29-31}\cmidrule(lr){32-34}\cmidrule(lr){35-37}
 & \textbf{Clean} 
& \(\downarrow\)\textbf{Induct Hd\,2\%} 
& \(\downarrow\)\textbf{Rand} 
& \textbf{Clean} 
& \(\downarrow\)\textbf{Induct Hd\,2\%} 
& \(\downarrow\)\textbf{Rand} 
& \textbf{Clean} 
& \(\downarrow\)\textbf{Induct Hd\,2\%} 
& \(\downarrow\)\textbf{Rand} 
& \textbf{Clean} 
& \(\downarrow\)\textbf{Induct Hd\,2\%} 
& \(\downarrow\)\textbf{Rand} 
& \textbf{Clean} 
& \(\downarrow\)\textbf{Induct Hd\,2\%} 
& \(\downarrow\)\textbf{Rand} 
& \textbf{Clean} 
& \(\downarrow\)\textbf{Induct Hd\,2\%} 
& \(\downarrow\)\textbf{Rand} 
& \textbf{Clean} 
& \(\downarrow\)\textbf{Induct Hd\,2\%} 
& \(\downarrow\)\textbf{Rand} 
& \textbf{Clean} 
& \(\downarrow\)\textbf{Induct Hd\,2\%} 
& \(\downarrow\)\textbf{Rand}
& \textbf{Clean} 
& \(\downarrow\)\textbf{Induct Hd\,2\%} 
& \(\downarrow\)\textbf{Rand} 
& \textbf{Clean} 
& \(\downarrow\)\textbf{Induct Hd\,2\%} 
& \(\downarrow\)\textbf{Rand} 
& \textbf{Clean} 
& \(\downarrow\)\textbf{Induct Hd\,2\%} 
& \(\downarrow\)\textbf{Rand} 
& \textbf{Clean} 
& \(\downarrow\)\textbf{Induct Hd\,2\%} 
& \(\downarrow\)\textbf{Rand}
\\
\midrule
alphabetically\_first\_3 & 4.86 $\pm$ 0.62 & 3.74 $\pm$ 0.66 & 5.69 $\pm$ 0.37 & 9.40 $\pm$ 0.57 & 8.86 $\pm$ 0.56 & 8.83 $\pm$ 0.58 & 19.71 $\pm$ 0.90 & 15.00 $\pm$ 0.86 & 19.51 $\pm$ 0.96 & 4.63 $\pm$ 0.57 & 4.43 $\pm$ 0.93 & 4.14 $\pm$ 0.53 & 7.34 $\pm$ 0.33 & 5.89 $\pm$ 0.57 & 7.03 $\pm$ 0.48 & 15.11 $\pm$ 0.89 & 12.94 $\pm$ 0.81 & 15.09 $\pm$ 0.91 & 3.34 $\pm$ 0.55 & 2.83 $\pm$ 0.50 & 3.57 $\pm$ 0.45 & 8.46 $\pm$ 0.88 & 7.40 $\pm$ 0.26 & 8.00 $\pm$ 0.51 & 14.43 $\pm$ 1.47 & 11.34 $\pm$ 1.13 & 14.49 $\pm$ 1.33 & 4.83 $\pm$ 0.62 & 3.17 $\pm$ 0.37 & 4.71 $\pm$ 0.80 & 11.54 $\pm$ 0.60 & 9.54 $\pm$ 0.47 & 11.26 $\pm$ 1.04 & 15.31 $\pm$ 0.96 & 13.66 $\pm$ 0.58 & 14.83 $\pm$ 0.68 \\
alphabetically\_first\_5 & 3.97 $\pm$ 0.57 & 3.43 $\pm$ 0.52 & 5.40 $\pm$ 0.94 & 6.71 $\pm$ 0.83 & 7.74 $\pm$ 0.66 & 6.69 $\pm$ 1.29 & 12.37 $\pm$ 1.41 & 11.03 $\pm$ 0.93 & 12.31 $\pm$ 1.14 & 4.23 $\pm$ 0.79 & 4.89 $\pm$ 0.42 & 4.57 $\pm$ 0.44 & 6.51 $\pm$ 1.03 & 6.46 $\pm$ 0.74 & 6.94 $\pm$ 0.80 & 10.89 $\pm$ 0.87 & 9.09 $\pm$ 0.65 & 9.63 $\pm$ 1.21 & 2.83 $\pm$ 0.68 & 2.57 $\pm$ 0.34 & 2.71 $\pm$ 0.30 & 7.37 $\pm$ 0.63 & 7.69 $\pm$ 0.23 & 7.80 $\pm$ 1.03 & 8.43 $\pm$ 1.15 & 7.97 $\pm$ 0.76 & 9.11 $\pm$ 0.80 & 3.94 $\pm$ 0.97 & 3.14 $\pm$ 0.83 & 4.26 $\pm$ 0.93 & 8.23 $\pm$ 0.78 & 6.80 $\pm$ 0.67 & 8.11 $\pm$ 1.16 & 10.20 $\pm$ 0.70 & 9.11 $\pm$ 0.69 & 10.26 $\pm$ 0.72 \\
alphabetically\_last\_3 & 3.37 $\pm$ 0.55 & 2.74 $\pm$ 0.34 & 4.69 $\pm$ 0.71 & 10.23 $\pm$ 0.88 & 8.40 $\pm$ 0.55 & 9.17 $\pm$ 0.77 & 20.80 $\pm$ 0.64 & 16.31 $\pm$ 1.23 & 18.80 $\pm$ 1.01 & 2.43 $\pm$ 0.54 & 2.31 $\pm$ 0.40 & 2.54 $\pm$ 0.55 & 7.69 $\pm$ 0.16 & 7.29 $\pm$ 0.77 & 7.83 $\pm$ 0.56 & 15.77 $\pm$ 0.96 & 12.31 $\pm$ 0.64 & 15.94 $\pm$ 0.97 & 1.94 $\pm$ 0.60 & 2.20 $\pm$ 0.30 & 2.60 $\pm$ 0.40 & 7.97 $\pm$ 0.87 & 7.71 $\pm$ 0.61 & 7.63 $\pm$ 0.58 & 13.26 $\pm$ 0.77 & 10.40 $\pm$ 0.75 & 13.80 $\pm$ 0.46 & 3.94 $\pm$ 0.47 & 2.00 $\pm$ 0.53 & 4.74 $\pm$ 0.43 & 10.89 $\pm$ 0.50 & 9.71 $\pm$ 0.66 & 11.49 $\pm$ 0.16 & 15.34 $\pm$ 1.35 & 13.91 $\pm$ 0.46 & 15.20 $\pm$ 0.78 \\
alphabetically\_last\_5 & 2.51 $\pm$ 0.75 & 2.09 $\pm$ 0.37 & 3.80 $\pm$ 0.78 & 6.03 $\pm$ 1.30 & 6.11 $\pm$ 1.36 & 6.06 $\pm$ 0.34 & 9.89 $\pm$ 0.82 & 9.20 $\pm$ 0.87 & 9.83 $\pm$ 0.76 & 1.60 $\pm$ 0.27 & 1.31 $\pm$ 0.48 & 1.89 $\pm$ 0.12 & 5.49 $\pm$ 0.39 & 5.40 $\pm$ 0.99 & 6.06 $\pm$ 0.61 & 8.51 $\pm$ 0.33 & 7.51 $\pm$ 1.21 & 8.97 $\pm$ 0.62 & 1.77 $\pm$ 0.36 & 1.49 $\pm$ 0.58 & 2.09 $\pm$ 0.59 & 6.06 $\pm$ 0.79 & 5.89 $\pm$ 0.91 & 6.94 $\pm$ 0.59 & 7.80 $\pm$ 0.48 & 7.14 $\pm$ 0.83 & 8.29 $\pm$ 0.42 & 2.51 $\pm$ 0.77 & 1.34 $\pm$ 0.22 & 2.14 $\pm$ 0.35 & 6.71 $\pm$ 0.55 & 6.54 $\pm$ 0.64 & 7.09 $\pm$ 0.41 & 9.31 $\pm$ 0.43 & 9.37 $\pm$ 0.63 & 9.63 $\pm$ 0.50 \\
capitalize & 8.00 $\pm$ 1.03 & 7.02 $\pm$ 0.62 & 9.93 $\pm$ 0.83 & 20.70 $\pm$ 1.40 & 13.19 $\pm$ 1.17 & 15.09 $\pm$ 1.61 & 54.81 $\pm$ 2.05 & 36.74 $\pm$ 2.61 & 52.42 $\pm$ 2.48 & 3.33 $\pm$ 0.28 & 3.75 $\pm$ 1.09 & 4.56 $\pm$ 0.25 & 13.19 $\pm$ 1.27 & 12.00 $\pm$ 1.28 & 14.84 $\pm$ 1.24 & 33.37 $\pm$ 1.31 & 25.93 $\pm$ 0.72 & 31.30 $\pm$ 1.15 & 3.72 $\pm$ 0.15 & 5.33 $\pm$ 1.26 & 4.67 $\pm$ 0.75 & 13.47 $\pm$ 1.33 & 11.16 $\pm$ 1.34 & 13.30 $\pm$ 1.62 & 39.23 $\pm$ 2.71 & 34.18 $\pm$ 1.46 & 37.40 $\pm$ 0.83 & 6.00 $\pm$ 1.63 & 5.65 $\pm$ 1.06 & 7.09 $\pm$ 1.12 & 14.67 $\pm$ 1.30 & 14.49 $\pm$ 1.15 & 12.77 $\pm$ 1.99 & 33.65 $\pm$ 2.14 & 27.93 $\pm$ 1.35 & 31.82 $\pm$ 1.33 \\
capitalize\_first\_letter & 10.07 $\pm$ 1.18 & 7.89 $\pm$ 0.98 & 13.30 $\pm$ 1.15 & 12.53 $\pm$ 1.77 & 11.05 $\pm$ 1.70 & 10.88 $\pm$ 1.36 & 28.56 $\pm$ 1.08 & 22.70 $\pm$ 0.52 & 24.74 $\pm$ 0.92 & 5.33 $\pm$ 1.00 & 5.40 $\pm$ 1.11 & 5.26 $\pm$ 0.62 & 13.40 $\pm$ 1.65 & 11.40 $\pm$ 1.16 & 14.14 $\pm$ 1.27 & 13.68 $\pm$ 1.16 & 11.58 $\pm$ 1.41 & 14.28 $\pm$ 1.23 & 5.19 $\pm$ 0.32 & 6.11 $\pm$ 2.06 & 5.58 $\pm$ 0.45 & 11.16 $\pm$ 1.21 & 10.56 $\pm$ 1.75 & 10.67 $\pm$ 1.33 & 17.44 $\pm$ 0.84 & 17.65 $\pm$ 2.57 & 20.98 $\pm$ 1.51 & 8.88 $\pm$ 0.66 & 7.75 $\pm$ 1.39 & 9.44 $\pm$ 0.96 & 10.00 $\pm$ 1.07 & 11.65 $\pm$ 1.76 & 10.98 $\pm$ 2.45 & 20.39 $\pm$ 1.17 & 18.84 $\pm$ 1.32 & 19.26 $\pm$ 1.54 \\
capitalize\_last\_letter & 4.84 $\pm$ 1.27 & 3.65 $\pm$ 0.94 & 5.19 $\pm$ 0.70 & 9.65 $\pm$ 1.05 & 8.98 $\pm$ 0.92 & 9.40 $\pm$ 1.43 & 8.28 $\pm$ 1.21 & 6.84 $\pm$ 1.27 & 7.61 $\pm$ 1.14 & 8.63 $\pm$ 1.12 & 10.88 $\pm$ 0.66 & 6.81 $\pm$ 0.70 & 7.51 $\pm$ 1.75 & 7.09 $\pm$ 0.88 & 5.93 $\pm$ 0.95 & 8.63 $\pm$ 0.85 & 9.33 $\pm$ 1.17 & 7.30 $\pm$ 0.90 & 9.09 $\pm$ 1.39 & 6.74 $\pm$ 1.56 & 7.93 $\pm$ 1.13 & 9.30 $\pm$ 1.16 & 8.39 $\pm$ 0.89 & 8.04 $\pm$ 1.22 & 7.26 $\pm$ 1.49 & 6.67 $\pm$ 0.76 & 6.32 $\pm$ 1.40 & 5.51 $\pm$ 0.67 & 5.65 $\pm$ 0.57 & 4.04 $\pm$ 0.92 & 5.75 $\pm$ 1.38 & 7.37 $\pm$ 0.67 & 5.33 $\pm$ 1.26 & 6.67 $\pm$ 0.66 & 5.68 $\pm$ 1.68 & 6.11 $\pm$ 0.65 \\
choose\_first\_of\_3 & 10.34 $\pm$ 2.29 & 6.86 $\pm$ 0.74 & 11.86 $\pm$ 0.90 & 25.46 $\pm$ 1.76 & 20.11 $\pm$ 1.77 & 23.29 $\pm$ 1.34 & 69.37 $\pm$ 1.77 & 49.20 $\pm$ 2.03 & 68.80 $\pm$ 0.61 & 6.14 $\pm$ 0.89 & 4.80 $\pm$ 0.85 & 6.11 $\pm$ 0.67 & 19.03 $\pm$ 1.72 & 15.14 $\pm$ 1.55 & 20.06 $\pm$ 2.31 & 52.37 $\pm$ 2.00 & 41.03 $\pm$ 2.98 & 52.89 $\pm$ 1.19 & 4.11 $\pm$ 0.72 & 3.60 $\pm$ 0.21 & 3.83 $\pm$ 0.36 & 19.06 $\pm$ 1.75 & 16.69 $\pm$ 1.56 & 19.60 $\pm$ 1.64 & 46.03 $\pm$ 2.04 & 34.43 $\pm$ 1.29 & 46.31 $\pm$ 1.30 & 11.29 $\pm$ 0.71 & 4.37 $\pm$ 0.46 & 11.71 $\pm$ 0.69 & 35.20 $\pm$ 1.70 & 23.49 $\pm$ 1.25 & 35.51 $\pm$ 2.16 & 54.09 $\pm$ 1.70 & 46.29 $\pm$ 1.12 & 50.03 $\pm$ 1.59 \\
choose\_first\_of\_5 & 8.66 $\pm$ 1.30 & 5.31 $\pm$ 1.01 & 10.43 $\pm$ 1.15 & 19.66 $\pm$ 1.94 & 17.63 $\pm$ 2.75 & 18.14 $\pm$ 2.56 & 55.51 $\pm$ 1.26 & 39.20 $\pm$ 2.13 & 53.71 $\pm$ 0.83 & 4.86 $\pm$ 0.73 & 4.23 $\pm$ 0.39 & 5.26 $\pm$ 0.58 & 15.31 $\pm$ 0.91 & 11.74 $\pm$ 1.82 & 16.31 $\pm$ 1.40 & 42.63 $\pm$ 3.05 & 33.89 $\pm$ 3.08 & 43.63 $\pm$ 1.43 & 3.63 $\pm$ 0.67 & 2.69 $\pm$ 0.40 & 3.17 $\pm$ 0.38 & 16.20 $\pm$ 1.49 & 14.43 $\pm$ 1.66 & 17.49 $\pm$ 1.29 & 32.69 $\pm$ 1.64 & 23.97 $\pm$ 0.53 & 33.71 $\pm$ 1.21 & 7.89 $\pm$ 1.01 & 3.43 $\pm$ 0.34 & 7.43 $\pm$ 0.77 & 28.20 $\pm$ 2.41 & 19.31 $\pm$ 0.53 & 27.74 $\pm$ 2.46 & 42.14 $\pm$ 1.37 & 34.14 $\pm$ 1.15 & 39.91 $\pm$ 0.82 \\
choose\_last\_of\_3 & 2.03 $\pm$ 0.42 & 1.86 $\pm$ 0.29 & 2.14 $\pm$ 0.40 & 3.17 $\pm$ 0.16 & 3.17 $\pm$ 1.29 & 2.86 $\pm$ 0.44 & 4.40 $\pm$ 0.98 & 5.11 $\pm$ 1.18 & 4.23 $\pm$ 0.88 & 1.43 $\pm$ 0.44 & 1.63 $\pm$ 0.34 & 1.94 $\pm$ 0.28 & 2.83 $\pm$ 0.26 & 2.54 $\pm$ 1.13 & 2.66 $\pm$ 0.52 & 5.40 $\pm$ 0.92 & 5.77 $\pm$ 1.38 & 5.63 $\pm$ 0.96 & 1.40 $\pm$ 0.51 & 0.94 $\pm$ 0.19 & 0.94 $\pm$ 0.13 & 2.97 $\pm$ 0.52 & 3.34 $\pm$ 0.50 & 3.46 $\pm$ 1.11 & 5.00 $\pm$ 0.51 & 5.11 $\pm$ 0.86 & 5.03 $\pm$ 1.18 & 1.69 $\pm$ 0.34 & 1.20 $\pm$ 0.16 & 1.37 $\pm$ 0.44 & 3.00 $\pm$ 0.53 & 4.00 $\pm$ 0.94 & 3.34 $\pm$ 0.70 & 5.20 $\pm$ 0.34 & 5.86 $\pm$ 0.91 & 5.69 $\pm$ 0.69 \\
choose\_last\_of\_5 & 1.51 $\pm$ 0.33 & 1.57 $\pm$ 0.58 & 2.00 $\pm$ 0.53 & 2.86 $\pm$ 0.61 & 2.46 $\pm$ 0.60 & 2.31 $\pm$ 0.38 & 3.89 $\pm$ 1.14 & 4.06 $\pm$ 0.65 & 3.63 $\pm$ 0.31 & 1.74 $\pm$ 0.47 & 1.63 $\pm$ 0.55 & 2.26 $\pm$ 0.44 & 2.37 $\pm$ 0.41 & 1.94 $\pm$ 0.16 & 2.69 $\pm$ 0.63 & 4.57 $\pm$ 0.58 & 4.86 $\pm$ 0.61 & 4.03 $\pm$ 0.62 & 1.09 $\pm$ 0.37 & 0.97 $\pm$ 0.33 & 1.31 $\pm$ 0.31 & 2.86 $\pm$ 0.52 & 2.66 $\pm$ 0.26 & 2.51 $\pm$ 0.47 & 5.63 $\pm$ 0.47 & 5.06 $\pm$ 0.52 & 4.86 $\pm$ 0.92 & 1.46 $\pm$ 0.33 & 1.14 $\pm$ 0.30 & 1.40 $\pm$ 0.42 & 2.63 $\pm$ 0.36 & 3.09 $\pm$ 0.47 & 2.89 $\pm$ 0.62 & 5.06 $\pm$ 0.36 & 4.97 $\pm$ 0.42 & 5.80 $\pm$ 0.44 \\
choose\_middle\_of\_3 & 1.69 $\pm$ 0.64 & 1.49 $\pm$ 0.16 & 1.83 $\pm$ 0.60 & 3.34 $\pm$ 0.55 & 3.31 $\pm$ 0.48 & 3.06 $\pm$ 0.47 & 4.23 $\pm$ 0.48 & 4.94 $\pm$ 0.52 & 4.37 $\pm$ 0.59 & 2.11 $\pm$ 0.80 & 1.86 $\pm$ 0.53 & 1.89 $\pm$ 0.51 & 2.23 $\pm$ 0.26 & 2.09 $\pm$ 0.41 & 2.37 $\pm$ 0.30 & 6.00 $\pm$ 1.03 & 5.51 $\pm$ 0.36 & 5.51 $\pm$ 0.64 & 1.34 $\pm$ 0.52 & 1.31 $\pm$ 0.70 & 1.43 $\pm$ 0.23 & 3.09 $\pm$ 0.24 & 2.63 $\pm$ 0.52 & 2.94 $\pm$ 0.52 & 4.97 $\pm$ 0.27 & 4.97 $\pm$ 0.31 & 4.83 $\pm$ 0.93 & 1.71 $\pm$ 0.66 & 1.20 $\pm$ 0.30 & 1.46 $\pm$ 0.23 & 3.40 $\pm$ 0.69 & 3.66 $\pm$ 0.58 & 3.11 $\pm$ 1.00 & 4.94 $\pm$ 0.70 & 5.20 $\pm$ 0.85 & 5.34 $\pm$ 0.63 \\
choose\_middle\_of\_5 & 1.63 $\pm$ 0.30 & 1.63 $\pm$ 0.46 & 2.00 $\pm$ 0.73 & 2.97 $\pm$ 0.64 & 3.29 $\pm$ 0.48 & 2.69 $\pm$ 0.52 & 2.71 $\pm$ 0.36 & 3.23 $\pm$ 0.56 & 2.89 $\pm$ 0.34 & 1.60 $\pm$ 0.23 & 1.63 $\pm$ 0.49 & 1.60 $\pm$ 0.53 & 2.09 $\pm$ 0.28 & 1.66 $\pm$ 0.46 & 1.86 $\pm$ 0.48 & 3.09 $\pm$ 0.62 & 3.66 $\pm$ 0.52 & 3.03 $\pm$ 0.63 & 1.66 $\pm$ 0.36 & 1.71 $\pm$ 0.44 & 1.57 $\pm$ 0.39 & 2.86 $\pm$ 0.69 & 2.86 $\pm$ 0.30 & 1.94 $\pm$ 0.52 & 3.89 $\pm$ 0.73 & 3.71 $\pm$ 0.27 & 3.34 $\pm$ 0.89 & 1.74 $\pm$ 0.44 & 1.40 $\pm$ 0.47 & 1.69 $\pm$ 0.29 & 2.34 $\pm$ 0.63 & 3.51 $\pm$ 0.22 & 2.71 $\pm$ 1.04 & 4.66 $\pm$ 0.63 & 4.54 $\pm$ 0.86 & 4.37 $\pm$ 0.49 \\
lowercase\_first\_letter & 5.51 $\pm$ 0.77 & 4.74 $\pm$ 0.99 & 7.72 $\pm$ 1.05 & 8.42 $\pm$ 0.86 & 7.26 $\pm$ 1.02 & 7.72 $\pm$ 1.11 & 28.53 $\pm$ 2.16 & 24.63 $\pm$ 1.69 & 26.60 $\pm$ 1.94 & 4.14 $\pm$ 0.92 & 2.63 $\pm$ 0.30 & 6.00 $\pm$ 1.31 & 6.07 $\pm$ 1.23 & 5.61 $\pm$ 0.72 & 7.26 $\pm$ 0.75 & 20.11 $\pm$ 1.56 & 17.05 $\pm$ 1.46 & 18.21 $\pm$ 1.22 & 4.67 $\pm$ 0.74 & 3.61 $\pm$ 0.97 & 4.98 $\pm$ 1.01 & 6.53 $\pm$ 0.59 & 6.56 $\pm$ 1.17 & 7.58 $\pm$ 0.60 & 20.56 $\pm$ 1.11 & 18.74 $\pm$ 1.52 & 18.98 $\pm$ 1.86 & 2.25 $\pm$ 0.57 & 4.14 $\pm$ 0.90 & 2.39 $\pm$ 0.69 & 8.84 $\pm$ 0.73 & 5.75 $\pm$ 0.44 & 9.44 $\pm$ 1.81 & 13.30 $\pm$ 0.84 & 13.61 $\pm$ 1.09 & 13.40 $\pm$ 1.56 \\
lowercase\_last\_letter & 11.09 $\pm$ 0.85 & 7.51 $\pm$ 1.80 & 8.49 $\pm$ 1.63 & 7.72 $\pm$ 0.73 & 7.37 $\pm$ 0.98 & 7.54 $\pm$ 1.19 & 10.46 $\pm$ 1.09 & 10.32 $\pm$ 1.00 & 8.81 $\pm$ 1.20 & 3.61 $\pm$ 0.67 & 2.84 $\pm$ 0.34 & 3.65 $\pm$ 1.16 & 8.04 $\pm$ 0.73 & 7.86 $\pm$ 1.48 & 8.18 $\pm$ 1.72 & 9.61 $\pm$ 0.96 & 9.26 $\pm$ 0.44 & 7.68 $\pm$ 1.43 & 7.37 $\pm$ 1.15 & 5.47 $\pm$ 0.64 & 8.18 $\pm$ 1.45 & 8.88 $\pm$ 0.97 & 9.26 $\pm$ 0.73 & 8.04 $\pm$ 1.58 & 13.26 $\pm$ 0.47 & 12.46 $\pm$ 1.75 & 12.67 $\pm$ 1.57 & 7.82 $\pm$ 1.10 & 9.79 $\pm$ 1.18 & 7.19 $\pm$ 0.71 & 10.88 $\pm$ 1.53 & 8.84 $\pm$ 1.68 & 8.67 $\pm$ 1.72 & 8.91 $\pm$ 1.18 & 9.09 $\pm$ 2.06 & 6.56 $\pm$ 1.77 \\
next\_capital\_letter & 4.88 $\pm$ 1.09 & 3.19 $\pm$ 0.98 & 6.42 $\pm$ 0.59 & 4.21 $\pm$ 0.94 & 3.44 $\pm$ 1.56 & 4.32 $\pm$ 0.20 & 2.42 $\pm$ 0.81 & 1.86 $\pm$ 0.24 & 2.74 $\pm$ 0.40 & 3.86 $\pm$ 0.33 & 4.28 $\pm$ 1.08 & 3.82 $\pm$ 0.78 & 4.18 $\pm$ 0.96 & 4.04 $\pm$ 0.83 & 3.44 $\pm$ 0.64 & 3.58 $\pm$ 0.72 & 3.47 $\pm$ 0.68 & 3.09 $\pm$ 0.36 & 4.67 $\pm$ 0.79 & 5.05 $\pm$ 1.24 & 5.09 $\pm$ 0.92 & 4.14 $\pm$ 0.71 & 3.65 $\pm$ 0.57 & 4.04 $\pm$ 0.55 & 3.05 $\pm$ 0.91 & 3.19 $\pm$ 0.60 & 2.98 $\pm$ 0.79 & 4.81 $\pm$ 0.91 & 3.96 $\pm$ 0.69 & 4.88 $\pm$ 0.88 & 3.47 $\pm$ 0.89 & 3.33 $\pm$ 0.62 & 4.07 $\pm$ 0.81 & 4.18 $\pm$ 1.15 & 2.56 $\pm$ 0.56 & 3.37 $\pm$ 0.74 \\
next\_item & 3.42 $\pm$ 2.17 & 2.78 $\pm$ 2.03 & 5.19 $\pm$ 1.22 & 8.61 $\pm$ 2.48 & 5.95 $\pm$ 2.26 & 13.16 $\pm$ 1.51 & 16.84 $\pm$ 2.13 & 17.09 $\pm$ 1.10 & 16.08 $\pm$ 0.57 & 2.91 $\pm$ 1.93 & 2.66 $\pm$ 1.58 & 4.30 $\pm$ 2.07 & 5.70 $\pm$ 1.34 & 6.46 $\pm$ 1.44 & 7.72 $\pm$ 1.76 & 11.77 $\pm$ 1.23 & 11.14 $\pm$ 2.03 & 12.91 $\pm$ 2.13 & 1.27 $\pm$ 0.90 & 2.15 $\pm$ 0.72 & 2.41 $\pm$ 1.37 & 7.85 $\pm$ 1.06 & 6.71 $\pm$ 2.71 & 6.71 $\pm$ 0.72 & 11.77 $\pm$ 2.48 & 12.41 $\pm$ 2.31 & 13.04 $\pm$ 1.15 & 6.33 $\pm$ 0.78 & 5.44 $\pm$ 1.82 & 7.85 $\pm$ 1.46 & 8.99 $\pm$ 1.64 & 8.73 $\pm$ 1.51 & 10.38 $\pm$ 2.85 & 9.37 $\pm$ 1.70 & 10.89 $\pm$ 1.87 & 9.87 $\pm$ 0.96 \\
prev\_item & 2.53 $\pm$ 0.78 & 2.03 $\pm$ 1.44 & 3.67 $\pm$ 1.51 & 7.72 $\pm$ 1.64 & 5.32 $\pm$ 1.15 & 11.39 $\pm$ 1.85 & 15.06 $\pm$ 2.30 & 15.82 $\pm$ 1.10 & 14.68 $\pm$ 1.13 & 2.03 $\pm$ 0.83 & 2.78 $\pm$ 1.31 & 3.04 $\pm$ 1.04 & 5.95 $\pm$ 0.72 & 6.71 $\pm$ 0.96 & 6.20 $\pm$ 1.70 & 10.25 $\pm$ 1.04 & 9.62 $\pm$ 2.16 & 9.49 $\pm$ 1.34 & 1.39 $\pm$ 0.53 & 2.28 $\pm$ 1.65 & 1.77 $\pm$ 0.28 & 6.58 $\pm$ 2.52 & 5.70 $\pm$ 1.85 & 7.09 $\pm$ 0.83 & 11.39 $\pm$ 2.45 & 12.03 $\pm$ 1.00 & 10.89 $\pm$ 1.44 & 5.70 $\pm$ 1.27 & 5.19 $\pm$ 1.13 & 7.97 $\pm$ 1.65 & 8.35 $\pm$ 1.22 & 9.49 $\pm$ 1.73 & 8.61 $\pm$ 2.39 & 7.72 $\pm$ 2.21 & 10.51 $\pm$ 0.57 & 7.97 $\pm$ 0.96 \\
word\_length & 9.89 $\pm$ 0.99 & 9.19 $\pm$ 1.06 & 8.74 $\pm$ 1.05 & 13.33 $\pm$ 0.94 & 11.96 $\pm$ 0.67 & 12.70 $\pm$ 1.76 & 12.77 $\pm$ 1.11 & 13.12 $\pm$ 0.68 & 12.07 $\pm$ 1.87 & 13.40 $\pm$ 0.83 & 10.21 $\pm$ 0.98 & 13.86 $\pm$ 1.67 & 14.00 $\pm$ 0.97 & 11.79 $\pm$ 0.57 & 12.49 $\pm$ 1.31 & 13.33 $\pm$ 0.84 & 12.18 $\pm$ 1.10 & 13.26 $\pm$ 1.19 & 12.00 $\pm$ 1.29 & 11.05 $\pm$ 1.20 & 11.40 $\pm$ 1.59 & 14.32 $\pm$ 0.58 & 13.33 $\pm$ 0.75 & 14.18 $\pm$ 1.68 & 13.58 $\pm$ 1.60 & 12.49 $\pm$ 1.83 & 13.37 $\pm$ 1.69 & 10.81 $\pm$ 1.09 & 10.14 $\pm$ 1.33 & 7.89 $\pm$ 0.76 & 14.74 $\pm$ 1.81 & 12.95 $\pm$ 1.16 & 13.51 $\pm$ 1.22 & 12.60 $\pm$ 0.72 & 12.42 $\pm$ 1.49 & 12.91 $\pm$ 1.60 \\
\midrule
\textbf{ICL Composite (macro)} & 5.3 $\pm$ 0.9 & 4.1 $\pm$ 0.8 & 6.2 $\pm$ 0.9 & 9.6 $\pm$ 1.1 & 8.2 $\pm$ 1.2 & 9.2 $\pm$ 1.1 & 20.0 $\pm$ 1.3 & 16.1 $\pm$ 1.1 & 19.2 $\pm$ 1.0 & 4.1 $\pm$ 0.7 & 3.9 $\pm$ 0.7 & 4.4 $\pm$ 0.8 & 7.8 $\pm$ 0.9 & 7.0 $\pm$ 0.9 & 8.1 $\pm$ 1.0 & 15.2 $\pm$ 1.1 & 13.0 $\pm$ 1.2 & 14.8 $\pm$ 1.1 & 3.8 $\pm$ 0.7 & 3.6 $\pm$ 0.8 & 4.0 $\pm$ 0.7 & 8.4 $\pm$ 1.0 & 7.7 $\pm$ 1.0 & 8.3 $\pm$ 1.0 & 14.7 $\pm$ 1.2 & 12.8 $\pm$ 1.1 & 14.8 $\pm$ 1.1 & 5.2 $\pm$ 0.8 & 4.2 $\pm$ 0.7 & 5.2 $\pm$ 0.8 & 10.4 $\pm$ 1.1 & 9.1 $\pm$ 0.9 & 10.4 $\pm$ 1.4 & 14.9 $\pm$ 1.1 & 13.6 $\pm$ 1.0 & 14.3 $\pm$ 1.0 \\

\bottomrule
\end{tabular}%
}
\end{table*}

\end{document}